\documentclass[10pt,twocolumn,letterpaper,pagenumbers]{article}

\usepackage{iccv}      

\usepackage{booktabs}
\usepackage{multirow}
\usepackage[T1]{fontenc}

\usepackage{adjustbox}
\usepackage{tikz}
\usepackage{forest}
\usetikzlibrary{arrows.meta, shapes, positioning, shadows, trees}

\definecolor{iccvblue}{rgb}{0.21,0.49,0.74}
\usepackage[pagebackref,breaklinks,colorlinks,allcolors=iccvblue]{hyperref}

\def\paperID{13262} 
\def\confName{ICCV}
\def\confYear{2025}

\title{Adversarial Robustness of Discriminative Self-Supervised Learning in Vision}

\author{Ömer Veysel Çağatan \and Ömer Faruk Tal \and M. Emre Gürsoy\\
  Department of Computer Engineering, Koç University\\
  {\tt\small ocagatan19@ku.edu.tr, otal19@ku.edu.tr, emregursoy@ku.edu.tr}
}

\begin{document}
\maketitle
\begin{abstract}
Self-supervised learning (SSL) has advanced significantly in visual representation learning, yet comprehensive evaluations of its adversarial robustness remain limited. In this study, we evaluate the adversarial robustness of seven discriminative self-supervised models and one supervised model across diverse tasks, including ImageNet classification, transfer learning, segmentation, and detection. Our findings suggest that discriminative SSL models generally exhibit better robustness to adversarial attacks compared to their supervised counterpart on ImageNet, with this advantage extending to transfer learning when using linear evaluation. However, when fine-tuning is applied, the robustness gap between SSL and supervised models narrows considerably. Similarly, this robustness advantage diminishes in segmentation and detection tasks. We also investigate how various factors might influence adversarial robustness, including architectural choices, training duration, data augmentations, and batch sizes. Our analysis contributes to the ongoing exploration of adversarial robustness in visual self-supervised representation systems.
\end{abstract}

\section{Introduction}

Self-supervised learning (SSL)~\cite{Balestriero2023ACO}, particularly discriminative approaches, has emerged as a foundational method for training models with remarkable capabilities in areas such as language~\cite{touvron2023llamaopenefficientfoundation}, vision~\cite{oquab2024dinov2learningrobustvisual}, and decision-making~\cite{kim2024openvlaopensourcevisionlanguageactionmodel}. As these models become increasingly widespread and integrated into various applications, ensuring their reliability and safety has become a critical concern~\cite{bommasani2022opportunitiesrisksfoundationmodels,Bengio_2024}. 

One particular challenge is the surprising vulnerability of deep learning models to adversarial examples, where slight input alterations can significantly impact model performance~\cite{szegedy2013intriguingpropertiesofneuralnetworks,goodfellow2014explainingandharnessingadversarialexamples}. This phenomenon has sparked significant debate, seeking to understand and mitigate these vulnerabilities~\cite{fawzi2016robustnessclassifiersadversarialrandom,tanay2016boundarytiltingpersepectivephenomenon,shafahi2020adversarialexamplesinevitable,schmidt2018adversariallyrobustgeneralizationrequires,wang2022convergencerobustnessadversarialtraining,Wang2020Improving,wu2020adversarialweightperturbationhelps,bai2022improvingadversarialrobustnesschannelwise}. One prominent theory~\cite{ilyas2019adversarialexamplesbugsfeatures} suggests that adversarial examples arise from the model's sensitivity to non-robust features in the input data. According to this view, both robust (stable) and non-robust (vulnerable) features contribute to classification, with adversarial attacks manipulating the latter to cause misclassification. However, this theory, developed primarily in the context of supervised learning, faces challenges when extended to other self-supervised paradigms. ~\cite{li2024adversarialexamplesrealfeatures} indicates that non-robust features are less effective in SSL methods such as contrastive learning~\cite{Chen2020ASF}, masked image modeling~\cite{He2021MaskedAA}, or diffusion models~\cite{ho2020denoisingdiffusionprobabilisticmodels}. This discrepancy suggests that non-robust features may lack the transferability across learning paradigms that robust or natural features possess. Thus, it becomes essential to systematically evaluate and compare how different SSL approaches respond to adversarial attacks, particularly given the theoretical evidence suggesting their feature representations may differ fundamentally from supervised models.

These theoretical insights into how adversarial examples affect different learning paradigms highlight several critical gaps in our understanding of SSL's adversarial robustness. Notwithstanding the progress made in understanding the adversarial robustness of SSL, particularly contrastive learning, which we extensively discuss in Section \ref{sec:rel}, several key questions remain unresolved. First, with the wide variety of self-supervised representations available, employing different pretext tasks and data augmentations, which approaches demonstrate the greatest adversarial robustness? This remains unclear since most methods don't provide any results on adversarial robustness unless it is a specific focus of the proposed approach. Secondly, robustness is typically assessed by the model's accuracy on the pretraining dataset. Still, its adversarial impact on transfer learning or downstream tasks like detection and segmentation has not been thoroughly investigated~\cite{kowalczuk2024benchmarkingrobustselfsupervisedlearning}.

The choice of model architecture also raises questions about robustness. Standard vision SSL pretraining typically utilizes a ResNet~\cite{he2015deepresiduallearningimage} as the backbone, but more recently, larger and more powerful models~\cite{chen2021mocov3,Caron2021EmergingPI,oquab2024dinov2learningrobustvisual} have been developed using vision transformers~\cite{dosovitskiy2021imageworth16x16words}. This leads to the question: Which architecture demonstrates greater robustness under the same SSL objective and with comparable parameter sizes?

Another factor to consider is the training duration. State-of-the-art SSL models are trained for longer durations compared to their supervised counterparts. Several studies indicate that this extended training consistently enhances performance, raising the question of whether this might compromise the models' adversarial robustness.

While previous work has examined aspects of adversarial robustness in SSL, our study provides the first comprehensive cross-model comparison across multiple tasks, architectures, and training regimes. We assess seven different SSL models (Barlow Twins~\cite{zbontar2021barlowtwinsselfsupervisedlearning}, BYOL~\cite{Grill2020BootstrapYO}, DINO~\cite{Caron2021EmergingPI}, MoCoV3~\cite{chen2021mocov3}, SimCLR~\cite{Chen2020ASF}, SwAV~\cite{caron2020unsupervised}, and VICReg~\cite{bardes2022vicregvarianceinvariancecovarianceregularizationselfsupervised}) alongside a supervised model against various adversarial attacks on ImageNet~\cite{russakovsky2015imagenetlargescalevisual} and nine other image-recognition datasets. We also evaluate their robustness in segmentation and detection tasks. Our investigation addresses the following key questions:

\begin{enumerate}
    
    \item \textbf{How does the adversarial robustness of various SSL models compare to that of supervised models on ImageNet?} \\
    SSL models consistently demonstrate greater adversarial robustness than supervised models on ImageNet. Non-contrastive methods show particular resilience against IAA~\cite{chakraborty2018adversarialattacksdefencessurvey} attacks, while all SSL approaches exhibit strong resistance to UAP~\cite{chaubey2020universaladversarialperturbations}, with MoCoV3 demonstrating the strongest overall performance.
        
    \item \textbf{Does SSL robustness transfer to downstream tasks like transfer learning, segmentation, and detection?} \\
    The robustness advantages transfer effectively to classification tasks via linear probing and fine-tuning. However, in segmentation and detection, all models exhibit similar vulnerability regardless of pretraining methodology, suggesting task-specific architectural components may override backbone robustness properties.

    \item \textbf{How does model architecture influence adversarial robustness under the same SSL objective?} \\
    Architecture impact is highly objective dependent. MoCoV3 shows reduced robustness with Vision Transformers, whereas DINO demonstrates improved performance with ViT compared to ResNet, challenging the notion that architectural effects are uniform across SSL paradigms. Additionally, we have evaluated DINOv2~\cite{oquab2024dinov2learningrobustvisual} and MAE~\cite{He2021MaskedAA} that do not demonstrate a consistently high or low level of robustness.

    \item \textbf{Does extending training duration compromise adversarial robustness in SSL models?} \\
    Extended training either maintains or slightly enhances adversarial performance. For UAP attacks, performance improves meaningfully after more epochs in both SwAV and MoCoV3, indicating that longer training periods do not create a performance-robustness tradeoff.
\end{enumerate}

\section{Related Work}\label{sec:rel}

\textbf{Self Supervised Learning}
Self-supervised learning(SSL) seeks to extract meaningful and general representations from unlabeled data by leveraging pretext tasks. These tasks can vary, such as predicting the next word~\cite{Radford2018ImprovingLU} or neighboring words~\cite{devlin2019bertpretrainingdeepbidirectional} in a text, reconstructing masked sections of an image~\cite{He2021MaskedAA}, or ensuring that two different perspectives of the same image result in similar visual representations~\cite{Chen2020ASF}.

Avoiding collapse is a key challenge in SSL for computer vision, and various methods can be classified based on how they address this issue. Contrastive approaches like SimCLR~\cite{Chen2020ASF} and MoCo~\cite{he2019moco,chen2020mocov2,chen2021mocov3} use an objective that pushes apart representations of different inputs (negative samples) while bringing together those of the same input (positive samples). The performance and scalability of these methods heavily depend on the number and selection of negative samples. In another category, distillation methods such as BYOL~\cite{Grill2020BootstrapYO}, SimSiam~\cite{Chen2020ExploringSS}, and DINO~\cite{Caron2021EmergingPI}, prevent collapse by introducing asymmetry between different encoder branches and employing algorithmic adjustments. Additional SSL techniques, including DeepCluster~\cite{caron2019deepclusteringunsupervisedlearning}, SeLa~\cite{asano2020selflabellingsimultaneousclusteringrepresentation}, and SwAV~\cite{caron2020unsupervised}, enforce a clustering structure in the feature space to avoid constant representations. Meanwhile, methods like Barlow Twins~\cite{zbontar2021barlowtwinsselfsupervisedlearning}, Whitening MSE (W-MSE)~\cite{ermolov2021whiteningselfsupervisedrepresentationlearning}, VICReg~\cite{bardes2022vicregvarianceinvariancecovarianceregularizationselfsupervised}, CorInfoMax~\cite{ozsoy2022selfsupervisedlearninginformationmaximization} prevent collapse by using feature decorrelation.

\begin{figure*}[t]
    \centering
    \small
    \includegraphics[width=0.8\textwidth,height=0.18\textheight]{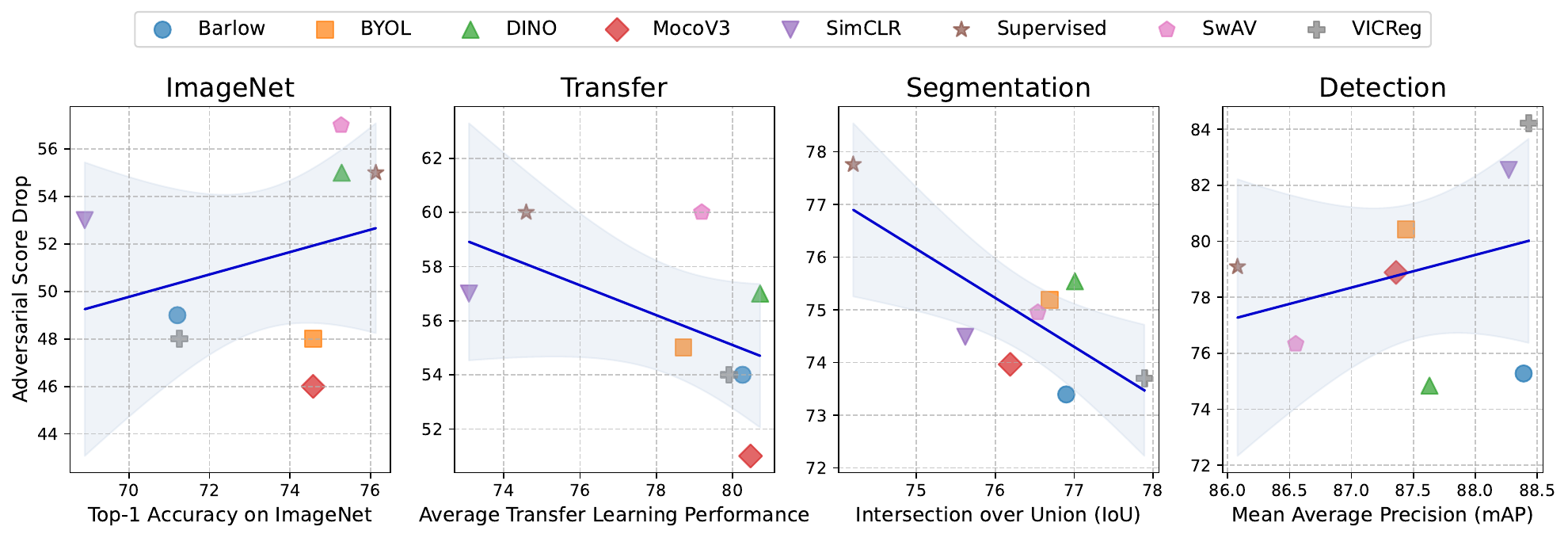}
    \caption{Performance scores for tasks such as ImageNet classification, transfer learning (with linear probing), segmentation, and detection (both with frozen backbones) are shown with the percentage drop in adversarial robustness. The shaded regions indicate the 95\% confidence interval around the regression line. Note the consistent pattern of higher robustness (lower percentage drop) among SSL models compared to supervised approaches in classification tasks.}
    \label{fig:adversarial-robustness}
\end{figure*}

\begin{figure*}[t]
    \centering
    \small
    \includegraphics[width=0.7\textwidth,height=0.18\textheight]{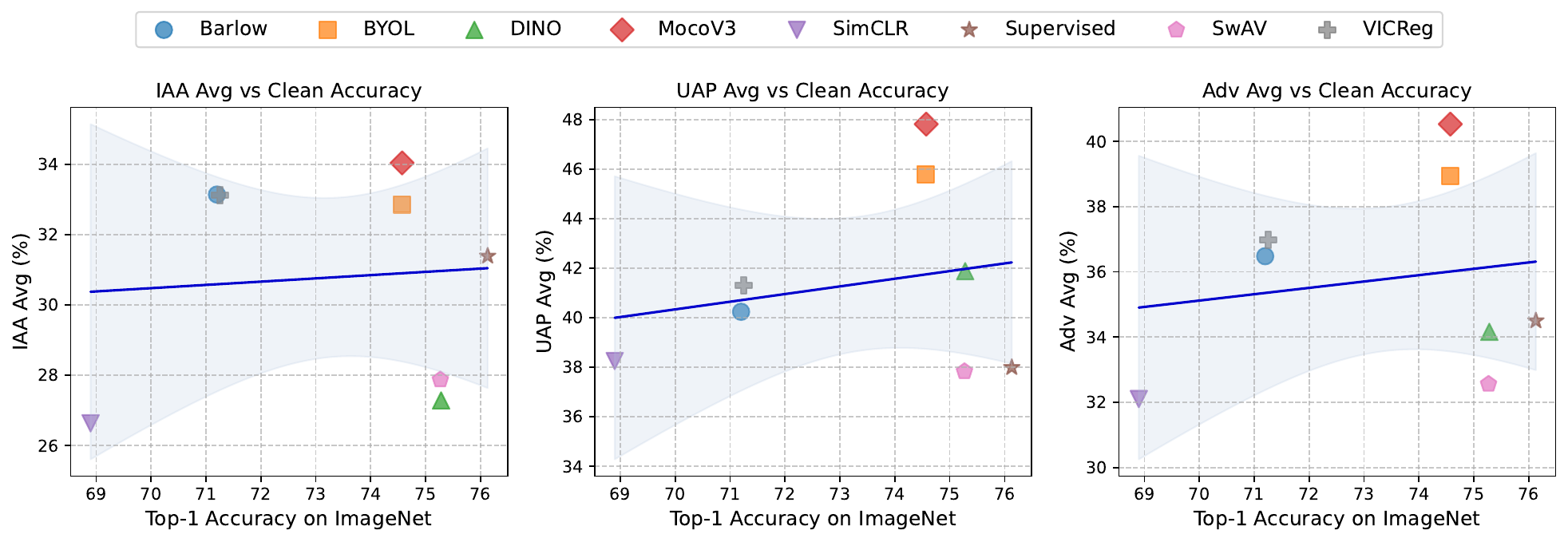}
    \caption{Averaged scores of SSL models on ImageNet across various attack types, including Instance Adversarial Attacks (IAA) and Universal Adversarial Perturbations (UAP). ~\textit{Adv Avg} refers to the average score across all attacks combined. The shaded regions indicate the 95\% confidence interval around the regression line.}
    \label{fig:iaa}
\end{figure*}

\textbf{Adversarial Self-Supervised Learning}
While self-supervised learning (SSL) has outperformed supervised training~\cite{Chen2020ASF}, numerous studies highlight that contrastive learning remains susceptible to adversarial attacks when transferring the learned features to downstream classification tasks~\cite{ho2020contrastivelearningadversarialexamples,kim2020adversarialselfsupervisedcontrastivelearning}. 
To improve the robustness of contrastive learning, adversarial training has been adapted to self-supervised settings. In the absence of labels, adversarial examples are generated by maximizing the contrastive loss with respect to all input samples. Several prior works, such as ACL~\cite{jiang2020robustpretrainingadversarialcontrastive}, RoCL~\cite{kim2020adversarialselfsupervisedcontrastivelearning}, and CLAE~\cite{ho2020contrastivelearningadversarialexamples}, adopt this approach. Additionally, ACL incorporates the dual-BN technique~\cite{xie2020adversarialexamplesimproveimage} to further enhance performance. DeACL~\cite{zhang2022decoupledadversarialcontrastivelearning} introduces a two-stage approach, distilling a standard pretrained encoder through adversarial training. ~\citet{nguyen2022taskagnosticrobustrepresentationlearning} establishes an upper bound on the adversarial loss of a prediction model, which is based on the learned representations, for any downstream task. This upper bound is determined using the model’s loss on clean data and a robustness regularization term, which helps make the prediction model more resistant to adversarial attacks.
~\cite{gupta2022contrastiveselfsupervisedlearningleads} demonstrates that adversarial sensitivity stems from the uniform distribution of data representations on a unit hypersphere in the representation space. The presence of false negative pairs during training contributes to this effect, increasing the model's vulnerability to input perturbations. 

Although self-supervised adversarial training has made progress, it still does not match the performance of supervised methods.~\citet{luo2023rethinkingeffectdataaugmentation} suggests that this shortfall is due to data augmentation and proposes a dynamic data augmentation scheduler to achieve comparable results to supervised training. ~\citet{xu2023efficientadversarialcontrastivelearning}  efficiently applies ACL on the ImageNet~\cite{russakovsky2015imagenetlargescalevisual} to obtain a robust representation using robustness-aware core set selection.

\textbf{Robustness of Self-Supervised Learning}

~\cite{hendrycks2019usingselfsupervisedlearningimprove} found that incorporating an extra self-supervised task in a multi-task framework can enhance the adversarial robustness of supervised models. In a similar vein, ~\citet{carmon2022unlabeleddataimprovesadversarial} discovered that using additional unlabeled data also strengthens the model's adversarial resilience. Furthermore, ~\citet{chen2020adversarialrobustnessselfsupervisedpretraining} created robust variants of pretext-based SSL tasks, showing that their integration with robust fine-tuning leads to a notable increase in robustness compared to standard adversarial training.
~\citet{chhipa2023selfsupervisedrepresentationlearningmethods} demonstrates a clear relationship between the performance of learned representations within SSL paradigms and the severity of distribution shifts and corruptions, and highlights the critical impact of distribution shifts and image corruptions on the performance and resilience of SSL methods. Similarly, ~\citet{zhong2022selfsupervisedlearningrobustsupervised} conduct robustness tests to assess the behavioral differences between contrastive and supervised learning under changes in downstream or pre-training data distributions, while also exploring the effects of data augmentation and feature space characteristics.~\citet{kowalczuk2024benchmarkingrobustselfsupervisedlearning} conducts a comprehensive empirical evaluation of the adversarial robustness of self-supervised vision encoders across multiple downstream tasks, revealing the need for broader enhancements in encoder robustness.~\citet{goldblum2023battlebackboneslargescalecomparison} benchmarks diverse pretrained models across multiple computer vision tasks, finding that supervised convolutional neural networks still outperform newer architectures on most metrics, while revealing self-supervised learning backbones show competitive potential when compared under equivalent conditions.

Unlike prior studies that primarily focus on individual SSL methods, specific tasks, or limited adversarial scenarios, our work provides a comprehensive, unified benchmark across multiple SSL paradigms, architectures, and tasks—including classification, transfer learning, segmentation, and detection—under a diverse set of adversarial attacks, offering a broader and deeper understanding of adversarial robustness in SSL.

\section{Experimental Setup}

\subsection{SSL Models}
While numerous SSL approaches have been proposed~\cite{ozbulak2023knowselfsupervisedlearningsurvey}, we focus on a subset of widely used models due to computational constraints: Barlow Twins~\cite{zbontar2021barlowtwinsselfsupervisedlearning}, BYOL~\cite{Grill2020BootstrapYO}, DINO~\cite{Caron2021EmergingPI}, MoCoV3~\cite{chen2021mocov3}, SimCLR~\cite{Chen2020ASF}, SwAV~\cite{caron2020unsupervised}, and VICReg~\cite{bardes2022vicregvarianceinvariancecovarianceregularizationselfsupervised}. We primarily use ResNet50~\cite{he2015deepresiduallearningimage} backbones, as most SSL checkpoints are released in this format, with DINO and MoCoV3 also offering ViT~\cite{dosovitskiy2021imageworth16x16words} variants. For BYOL, DINO, MoCoV3, SimCLR, and SwAV, complete model checkpoints were available. In contrast, Barlow Twins and VICReg only provided backbone weights, requiring linear evaluation via official code, which led to a slight drop in performance. For comparison, we also include a supervised ResNet50 baseline from PyTorch~\cite{paszke2019pytorchimperativestylehighperformance}.

\subsection{ImageNet and Transfer Learning}

We use the benchmark suite introduced in the transfer learning study~\cite{huh2016makesimagenetgoodtransfer}, which encompasses the target datasets like FGVC Aircraft~\cite{maji2013finegrainedvisualclassificationaircraft}, Caltech-101~\cite{1384978}, Stanford Cars~\cite{6755945}, CIFAR 10~\cite{Krizhevsky2009LearningML}, CIFAR 100~\cite{Krizhevsky2009LearningML}, DTD~\cite{cimpoi2013describingtextureswild}, Oxford 102 Flowers~\cite{cimpoi2013describingtextureswild}, and Food-101~\cite{10.1007/978-3-319-10599-4_29}. We follow ~\citet{ericsson2021selfsupervisedmodelstransfer} for both linear evaluation and fine-tuning of these datasets. We prioritized linear evaluation in our analysis as the backbone remains frozen during this process, allowing for a more equitable comparison of objectives within this setup. We apply the same adversarial techniques to ImageNet and transfer learning: Instance Adversarial Attacks (IAA) and Universal Adversarial Perturbations (UAP). In brief, instance-based methods generate unique perturbations for each image, while UAP creates a single perturbation that applies across the entire dataset. Comprehensive details and categorizations of all attack methods—such as white-box, black-box, gradient-free, among others—are provided in Section 1 of the Supp Mat..

\subsection{Segmentation}
For segmentation, we use the Pascal VOC 2012 dataset~\cite{pascal-voc-2012} and CityScapes~\cite{cordts2016cityscapesdatasetsemanticurban} dataset, training a DeepLabV3+ model~\cite{chen2018encoderdecoderatrousseparableconvolution}. To conduct the attacks, we follow the setup from ~\citet{rony2022proximal}, utilizing Alma~\cite{rony2022proximal}, Asma~\cite{rony2022proximal}, DAG~\cite{xie2017adversarialexamplessemanticsegmentation}, DDN~\cite{rony2022proximal}, FGSM~\cite{goodfellow2014explainingandharnessingadversarialexamples}, FMN~\cite{pintor2021fastminimumnormadversarialattacks}, and PGD~\cite{madry2017towardsdeeplearningmodelsresitanttoadversarialattacks}. While our primary metric is the mean Intersection Over Union (mIOU), we also report the Attack Pixel Success Rate (APSR) introduced by ~\cite{rony2022proximal}. Although our main focus is on using a frozen backbone, we also perform training following the standard procedure. Our CityScapes results include only the frozen backbone approach, while our Pascal VOC results include both frozen and unfrozen backbone configurations. 
\subsection{Detection}
For object detection, we utilized the INRIA Person~\cite{1467360} and CoCo~\cite{lin2015microsoftcococommonobjects} datasets, and trained a Faster R-CNN~\cite{ren2016fasterrcnnrealtimeobject}. To perform adversarial attacks, we followed the setup described by \citep{huang2023tseatransferbasedselfensembleattackonobjectdetection}, employing the Transfer-based Self-Ensemble Attack (T-SEA). The T-SEA attack can be deployed using various methods and optimizers. In our experiments, we employed BIM~\cite{huang2023tseatransferbasedselfensembleattackonobjectdetection}, MIM~\cite{dong2018boostingadversarialattacksmomentum}, PGD~\cite{madry2017towardsdeeplearningmodelsresitanttoadversarialattacks}, and Optim~\cite{huang2023tseatransferbasedselfensembleattackonobjectdetection} methods. Additionally, we explored simpler methods that rely on common optimizers, such as Adam~\cite{kingma2017adammethodstochasticoptimization}, SGD, and Nesterov~\cite{Nesterov1983AMF}. Throughout our evaluation, we report the mean average precision (mAP) scores as the primary performance metric. While our primary focus was on employing a frozen backbone, we also conducted training experiments following the standard training procedures for comparative analysis. Our COCO results include only the frozen backbone approach.

\begin{table}[t]
\caption{Performance of various models on ImageNet, Transfer Learning, Segmentation (Pascal VOC), and Detection (INRIA Person) tasks with frozen backbones, showing original (Orig.) and adversarial (Adv.) scores with performance drops in red.}
\label{tab:ssl-robustness}
\centering
\resizebox{\columnwidth}{!}{%
\begin{tabular}{l cccc cccc}
\toprule
\multirow{2}{*}{Model} & \multicolumn{2}{c}{ImageNet} & \multicolumn{2}{c}{Transfer Learning} & \multicolumn{2}{c}{Segmentation} & \multicolumn{2}{c}{Detection} \\
\cmidrule(r){2-3} \cmidrule(r){4-5} \cmidrule(r){6-7} \cmidrule(r){8-9}
& Orig. & Adv. & Orig. & Adv. & Orig. & Adv. & Orig. & Adv. \\
\midrule
Barlow Twins &71.2 & 36.5 \textcolor{red}{\scriptsize{$\downarrow$49\%}} &80.3 &37.8 \textcolor{red}{\scriptsize{$\downarrow$54\%}} &76.9 &20.5 \textcolor{red}{\scriptsize{$\downarrow$73\%}} &88.4 & 21.9 \textcolor{red}{\scriptsize{$\downarrow$75\%}}\\
BYOL &74.6 &38.9 \textcolor{red}{\scriptsize{$\downarrow$48\%}} &78.7 &36.6 \textcolor{red}{\scriptsize{$\downarrow$55\%}} &76.7 &19.0 \textcolor{red}{\scriptsize{$\downarrow$75\%}} & 87.4 &17.3 \textcolor{red}{\scriptsize{$\downarrow$80\%}} \\
DINO  &75.3 & 34.2 \textcolor{red}{\scriptsize{$\downarrow$55\%}}&80.7 & 35.6 \textcolor{red}{\scriptsize{$\downarrow$57\%}}& 77.0&18.9 \textcolor{red}{\scriptsize{$\downarrow$76\%}} & 87.6&22.0 \textcolor{red}{\scriptsize{$\downarrow$75\%}} \\
MoCoV3 &74.6 & 40.5 \textcolor{red}{\scriptsize{$\downarrow$46\%}}& 80.5& 40.3 \textcolor{red}{\scriptsize{$\downarrow$51\%}} &76.2 &19.9 \textcolor{red}{\scriptsize{$\downarrow$74\%}} &87.3 &18.5 \textcolor{red}{\scriptsize{$\downarrow$79\%}} \\
SimCLR & 68.9& 32.1\textcolor{red}{\scriptsize{$\downarrow$53\%}}&73.1 &33.0 \textcolor{red}{\scriptsize{$\downarrow$57\%}} &75.6 &19.3 \textcolor{red}{\scriptsize{$\downarrow$74\%}} &87.7 &13.1 \textcolor{red}{\scriptsize{$\downarrow$85\%}} \\
Supervised &76.1 &34.5 \textcolor{red}{\scriptsize{$\downarrow$55\%}} &74.6 &31.2 \textcolor{red}{\scriptsize{$\downarrow$60\%}} &74.2 &16.5 \textcolor{red}{\scriptsize{$\downarrow$78\%}} & 86.1&18.0 \textcolor{red}{\scriptsize{$\downarrow$79\%}} \\
SwAV & 75.3& 32.6 \textcolor{red}{\scriptsize{$\downarrow$57\%}}&79.2 &32.4 \textcolor{red}{\scriptsize{$\downarrow$60\%}} & 76.5&19.2 \textcolor{red}{\scriptsize{$\downarrow$75\%}} &86.6 &20.5 \textcolor{red}{\scriptsize{$\downarrow$76\%}} \\
VICReg  &71.3 &37.0 \textcolor{red}{\scriptsize{$\downarrow$48\%}} &79.9 &37.7 \textcolor{red}{\scriptsize{$\downarrow$54\%}} &77.9 &20.5 \textcolor{red}{\scriptsize{$\downarrow$74\%}} &88.4 &14.0 \textcolor{red}{\scriptsize{$\downarrow$84\%}} \\
\bottomrule
\end{tabular}%
}
\end{table}

\section{Results and Discussion}
In this section, we present our experimental findings on ImageNet, transfer learning, and detection, and discuss each in turn. Figure \ref{fig:adversarial-robustness} and table \ref{tab:ssl-robustness} summarize the performance of various SSL models compared to supervised learning across our main evaluation tasks in the frozen backbone setup. While we address the results individually, the full detailed results are provided in Section 4 of the Supp. Mat..
\subsection{ImageNet}

\textbf{SSL vs Supervised}.
Most robustness studies on contrastive learning~\cite{ho2020contrastivelearningadversarialexamples,kim2020adversarialselfsupervisedcontrastivelearning,jiang2020robustpretrainingadversarialcontrastive,xie2020adversarialexamplesimproveimage,zhang2022decoupledadversarialcontrastivelearning,nguyen2022taskagnosticrobustrepresentationlearning} focus on small datasets like CIFAR10~\cite{Krizhevsky2009LearningML} and primarily evaluate robustness using adversarial attacks such as FGSM~\cite{goodfellow2014explainingandharnessingadversarialexamples} and PGD~\cite{madry2017towardsdeeplearningmodelsresitanttoadversarialattacks}. While computational constraints explain the reluctance to scale to larger datasets like ImageNet~\cite{russakovsky2015imagenetlargescalevisual}, many evaluations inadequately incorporate Universal Adversarial Perturbations (UAP). Our findings, as shown in Figure~\ref {fig:iaa} contradict previous research by~\citet{gupta2022contrastiveselfsupervisedlearningleads}. Under IAA, MoCoV3 demonstrates the strongest robustness (54\% drop), while SimCLR shows the weakest performance (61\% drop). For UAP attacks, MoCoV3 again leads (38\% drop), while the supervised model and SwAV both show 50\% drops. Notably, our results challenge the conclusion about contrastive vs. non-contrastive methods. MoCoV3, a contrastive model, consistently demonstrates the highest adversarial robustness with ResNet architecture, while DINO, which was claimed to perform better due to its non-contrastive nature, shows the weakest IAA robustness in our ResNet evaluation (64\% drop). Our data reveals that Barlow, BYOL, MoCoV3, and VICReg all demonstrate comparable resilience against IAA (around 54\% drop), contradicting the simple categorization of robustness based on contrastive versus non-contrastive approaches. Our comprehensive evaluation using diverse attack methods demonstrates that the relationship between self-supervised learning approaches and adversarial robustness is more complex than previously suggested. The full results are in Section 4.1 of the Supp. Mat..

\textbf{What makes MoCoV3 robust?}
Although MoCoV3 and SimCLR both utilize the InfoNCE~\cite{Sohn2016ImprovedDM,oord2019representationlearningcontrastivepredictive} objective, there is a stark contrast in their adversarial robustness. To understand this disparity, we evaluate MoCoV1~\cite{he2019moco}, MoCoV2~\cite{chen2020mocov2}, and MoCoV3.

\textbf{A brief MoCo History}. \textit{MoCoV1 introduced a dynamic dictionary with a queue and momentum-updated encoder to maintain consistent negative samples. MoCoV2 enhanced this with a multi-layer projection head and stronger data augmentation. MoCoV3 further evolved by eliminating the memory bank and incorporating a prediction head similar to BYOL and SimSiam~\cite{Chen2020ExploringSS}.}

Our analysis reveals a clear progression in robustness across MoCo versions. While MoCoV1 demonstrates limited resilience with a 71\% drop in overall adversarial accuracy, MoCoV2 shows significant improvement (52\% drop), and MoCoV3 achieves the strongest performance (46\% drop). The most dramatic enhancement occurs in UAP resistance, where MoCoV1's performance drops by 77\%, compared to MoCoV2's 39\% and MoCoV3's 36\%. The substantial improvement from MoCoV1 to MoCoV2 (7 \% in clean accuracy, 27 percentage points in UAP robustness) primarily stems from the non-linear projector, with data augmentation providing marginal benefits. While~\cite{ibrahim2024occamsrazorselfsupervised} suggests non-linear projectors aren't always essential, our results indicate they significantly enhance both performance and adversarial robustness. MoCoV3's superior performance over MoCoV2 (an additional 7\% in clean accuracy and 7 percentage points in UAP robustness) can be attributed to its prediction head and larger batch size. Unlike the transition from V1 to V2, MoCoV3 shows substantial improvements in both IAA (34\% vs 25\%) and UAP (48\% vs 41\%), highlighting the prediction head's critical role in enhancing overall robustness. Our findings suggest that momentum, a common feature in robust models like MoCoV3 and BYOL, significantly contributes to adversarial resilience, while MoCoV2's performance more closely resembles that of SimCLR, which lacks this feature. The full results are in Section 4.6 of the supplementary material.

\textbf{Augmentations vs Algorithms}.

~\citet{morningstar2024augmentationsvsalgorithmsworks} challenges the notion that SSL progress is primarily driven by algorithmic advancements, suggesting instead that augmentation diversity, along with data and model scale, play more critical roles. Their analysis argues that many algorithmic improvements, such as prediction networks or new loss functions, had minimal impact on downstream task performance compared to stronger augmentation techniques.

Our comprehensive analysis of the MoCo family evolution provides a more nuanced perspective on this debate. The substantial progression in adversarial robustness from MoCoV1 to MoCoV2 and further to MoCoV3 suggests that architectural innovations like non-linear projectors and prediction heads significantly impact robustness. MoCoV2's dramatic improvement over MoCoV1, particularly in UAP performance, indicates that the multi-layer projection head provides substantial benefits beyond mere augmentation changes. Similarly, MoCoV3's further enhancements in both IAA and UAP performance relative to MoCoV2 highlight the crucial role of the prediction head in overall robustness.

While our comparison lacks perfectly controlled baselines for augmentations across different objectives (with slight variations in augmentation between MoCoV2 and V3), this limitation stems from the unavailability of public checkpoints rather than our experimental design. Despite this constraint, the consistent improvements in adversarial performance strongly suggest that algorithmic innovations significantly contribute to adversarial robustness. Importantly, our findings demonstrate that higher clean accuracy doesn't automatically translate to improved robustness on ImageNet.

Although the checkpoints from ~\cite{morningstar2024augmentationsvsalgorithmsworks} are not publicly available, we conducted an ablation study on augmentation types and batch sizes using BYOL which is the only model in our SSL pool that includes configurations with varied augmentations and batch sizes. For this analysis, we consider four distinct models: BYOL-NC (without color distortions), BYOL-CC (with only color and crop augmentations), and the standard BYOL models with batch sizes of 128 (BYOL-128) and 512 (BYOL-512).

We observe that batch size and augmentation choices have varying effects on the adversarial robustness of BYOL variants. For IAA robustness, both BYOL-NC and BYOL-CC show similar performance (60\% drop), while BYOL-128, BYOL-512, and standard BYOL demonstrate slightly better resilience (57\%, 56\%, and 56\% drops, respectively). These modest differences suggest that batch size and augmentation have a limited impact on instance-level attack robustness.

For UAP attacks, we see more substantial variations: BYOL-128 (41\% drop) and BYOL-512 (43\% drop) perform comparably, while BYOL-NC and BYOL-CC show notably weaker performance (46\% and 48\% drops). Standard BYOL demonstrates a 39\% drop, which is marginally better than the batch size variants but substantially better than the limited-augmentation models. The 7-9 percentage point difference between standard BYOL and the limited-augmentation variants suggests that comprehensive augmentation strategies may contribute to improved UAP robustness.

Interestingly, we note that BYOL-CC performs slightly worse than BYOL-NC despite having more augmentations, though this difference is too small to draw meaningful conclusions. Similarly, the differences between BYOL-128, BYOL-512, and standard BYOL in UAP robustness are relatively minor and should be interpreted cautiously. Our results indicate that the relationship between augmentation and adversarial robustness is complex and that meaningful improvements likely require more than just incremental changes to batch size or augmentation strategies. While our data hints at potential benefits from comprehensive augmentation for UAP robustness, the effects are modest and require further investigation with more controlled experiments. The full results are in Section 4.7 of the supplementary material.

\begin{table*}[t]
\caption{Component Contribution Analysis Across Datasets. Metrics include: Original Accuracy (Orig), Adversarial Accuracy (Adv) with relative performance drop percentage, Direction Ratio (D) and Magnitude Ratio (M). Ratio interpretation: Head dominant (<0.67), Balanced (0.67-1.5), Backbone dominant (>1.5).}

\label{tab:component-analysis}
\centering
\footnotesize
\setlength{\tabcolsep}{1.5pt}
\begin{tabular}{l|cccc|cccc|cccc|cccc|cccc}
\toprule
& \multicolumn{4}{c|}{\textbf{CIFAR-10}} & \multicolumn{4}{c|}{\textbf{CIFAR-100}} & \multicolumn{4}{c|}{\textbf{ImageNet}} & \multicolumn{4}{c|}{\textbf{CIFAR-10 (FT)}} & \multicolumn{4}{c}{\textbf{CIFAR-100 (FT)}} \\
\textbf{Model} & Orig & Adv $\downarrow$ & D & M & Orig & Adv $\downarrow$ & D & M & Orig & Adv $\downarrow$ & D & M & Orig & Adv $\downarrow$ & D & M & Orig & Adv $\downarrow$ & D & M \\
\midrule
Barlow    & 92.3 & 33.0 \textcolor{red}{$\downarrow$64\%} & 2.7 & 3.1 & 77.9 & 20.5 \textcolor{red}{$\downarrow$74\%} & 1.4 & 5.0 & 71.2 & 42.4 \textcolor{red}{$\downarrow$40\%} & 0.9 & 9.4 & 97.1 & 66.9 \textcolor{red}{$\downarrow$31\%} & 1.5 & 2.4 & 84.6 & 45.0 \textcolor{red}{$\downarrow$57\%} & 0.8 & 2.4 \\
BYOL      & 93.0 & 31.0 \textcolor{red}{$\downarrow$67\%} & 1.6 & 4.2 & 78.2 & 19.01 \textcolor{red}{$\downarrow$76\%} & 1.07 & 6.0 & 74.6 & 39.4 \textcolor{red}{$\downarrow$47\%} & 0.9 & 12.5 & 96.9 & 67.0 \textcolor{red}{$\downarrow$31\%} & 1.2 & 1.4 & 83.9 & 61.2 \textcolor{red}{$\downarrow$27\%} & 0.7 & 2.7 \\
DINO    & 93.9 & 27.6 \textcolor{red}{$\downarrow$71\%} & 3.2 & 3.3 & 79.7 & 16.0 \textcolor{red}{$\downarrow$80\%} & 1.6 & 6.2 & 75.3 & 24.7 \textcolor{red}{$\downarrow$67\%} & 1.2 & 14.1 & 96.9 & 77.8 \textcolor{red}{$\downarrow$20\%} & 1.9 & 2.9 & 84.7 & 52.9 \textcolor{red}{$\downarrow$38\%} & 1.79 & 4.6 \\
MoCoV3 & 94.7 & 33.0\textcolor{red}{$\downarrow$65\%} & 2.5 & 1.7 & 80.2 & 19.3 \textcolor{red}{$\downarrow$76\%} & 1.1 & 4.5 & 74.6 & 42.7 \textcolor{red}{$\downarrow$43\%} & 0.7 & 10.9 & 96.9 & 72.0 \textcolor{red}{$\downarrow$26\%} & 1.1 & 1.3 & 84.5 & 62.6 \textcolor{red}{$\downarrow$26\%} & 0.7 & 2.6 \\
SimCLR   & 91.0 & 37.9 \textcolor{red}{$\downarrow$58\%} & 1.5 & 1.9 & 73.0 & 19.5 \textcolor{red}{$\downarrow$73\%} & 1.3 & 3.5 & 68.9 & 24.3 \textcolor{red}{$\downarrow$65\%} & 1.0 & 7.7 & 97.2 & 67.3 \textcolor{red}{$\downarrow$31\%} & 1.0 & 1.9 & 84.4 & 43.0 \textcolor{red}{$\downarrow$50\%} & 0.6 & 2.0\\
Supervised      & 91.4 & 42.8 \textcolor{red}{$\downarrow$53\%} & 1.8 & 0.7 & 73.9 & 24.5 \textcolor{red}{$\downarrow$67\%} & 1.3 & 1.5 & 76.1 & 38.8\textcolor{red}{$\downarrow$49\%} & 1.4 & 3.0 & 96.2 & 62.3 \textcolor{red}{$\downarrow$35\%} & 2.0 & 0.5 & 82.6 & 59.3 \textcolor{red}{$\downarrow$28\%} & 1.3 & 1.1 \\
SwAV      & 93.9 & 19.4 \textcolor{red}{$\downarrow$79\%} & 3.1 & 3.0 & 79.4 & 11.1 \textcolor{red}{$\downarrow$86\%} & 1.8 & 6.4 & 75.3 & 24.7 \textcolor{red}{$\downarrow$67\%} & 1.2 & 17.4 & 96.8 & 79.9 \textcolor{red}{$\downarrow$17\%} & 1.7 & 2.7 & 84.4 & 54.6 \textcolor{red}{$\downarrow$35\%} & 1.0 & 3.0 \\
VICReg    & 92.8 & 33.0 \textcolor{red}{$\downarrow$64\%} & 2.7 & 3.1 & 77.8 & 22.3 \textcolor{red}{$\downarrow$71\%} & 1.3 & 4.3 & 71.3 & 42.4 \textcolor{red}{$\downarrow$40\%} & 0.9 & 9.3 & 97.1 & 68.6 \textcolor{red}{$\downarrow$29\%} & 1.4 & 2.4 & 84.3 & 43.4 \textcolor{red}{$\downarrow$48\%} & 0.8 & 2.4 \\
\bottomrule
\end{tabular}
\end{table*}

\textbf{ResNet vs ViT in Adversarial Robustness}.
While ViTs are generally seen as more robust than CNNs~\cite{naseer2020selfsupervisedapproachforadversarialrobustness}, \citet{pinto2022impartialcnnvstransformer,bai2021transformersrobustcnns} demonstrate that with the right training methods, CNNs~\cite{726791} can achieve comparable robustness. Despite ViT's success~\cite{dehghani2023scalingvisiontransformers22,dosovitskiy2021imageworth16x16words,chen2021mocov3,Caron2021EmergingPI,oquab2024dinov2learningrobustvisual}, most SSL methods still use ResNet for validation. We examine MoCoV3 and DINO, as they are the only models that include ViT training, with our analysis covering both standard ViTs comparable to ResNet50 and larger ViT-B variants with approximately 4x more parameters.

Our results reveal notable architectural differences. DINO performs significantly better with ViT architectures than ResNet, with DINO-ViT-B achieving 45.19\% adversarial average (42\% drop from clean accuracy) compared to DINO-ResNet's 55\% drop. In contrast, MoCo struggles with transformer architectures, with MoCo-ViT showing poor performance at a 61\% drop, significantly worse than MoCoV3-ResNet at a 46\% drop. Even MoCoV3-ViT-B 51\% drop underperforms its ResNet counterpart despite having 4x more parameters.

These differences are especially evident in UAP results, where DINO-ViT-B shows greater resilience (38\% drop) than DINO-ResNet (44\% drop), while MoCo-ViT performs poorly (26.31\%, 64\% drop) compared to MoCoV3-ResNet (36\% drop). These findings suggest that the interaction between self-supervised approaches and model architectures significantly impacts adversarial robustness, with DINO benefiting from the ViT architecture while MoCo struggles with transformer models. The full results are in Section 4.6 of the Supp. Mat..

Lastly, we have evaluted  DINOv2's~\cite{oquab2024dinov2learningrobustvisual} small and base and MAE's~\cite{He2021MaskedAA} base versions. We discuss them in the section 2 in Supp. Mat..

\textbf{Impact of Training Duration.}
SSL models tend to demonstrate better performance as training epochs increase~\cite{Chen2020ASF,chen2021mocov3,caron2020unsupervised}. However, due to computational constraints, many models are reported with different numbers of epochs. This prompts the question of whether longer training durations enhance or reduce adversarial robustness. As ViT models do not have checkpoints at various epochs, we focus on ResNet-based SSL models, specifically SwAV and MoCoV3, which offer multiple checkpoints throughout the training process. We find that both SwAV and MoCo show very marginal improvement of about 1\% on IAA across various epochs, which is minimal compared to the rise in original accuracy. In contrast, both methods exhibit a modest increase in UAP performance after surpassing 100 epochs, with SwAV improving from 33\% to 40\% and MoCoV3 from 41\% to 48\%. Overall, our results suggest that despite differences in reported checkpoints, robustness generally remains largely stable during training for IAA, with slightly more noticeable but still modest gains for UAP. This indicates that training duration has a limited impact on adversarial robustness, even as clean accuracy continues to improve with extended training. The full results are in Section 4.5 of the Supp. Mat.

\subsection{Transfer Learning}

In this section, we analyze how adversarial robustness transfers from ImageNet pre-training to downstream tasks. We examine whether vulnerability patterns established during pre-training persist when models are evaluated through linear probing or fine-tuning across various ResNet-based self-supervised and supervised learning approaches. This analysis quantifies robustness transfer relationships through both Spearman rank and Pearson correlations.

\textbf{LINEAR}.
Our correlation analysis shows strong relationships between ImageNet and linear evaluation vulnerability. For performance drops, both Spearman and Pearson correlations are high across all attack types (Spearman—Adv: 0.93, UAP: 0.83, IAA: 0.79; Pearson—Adv: 0.94, UAP: 0.94, IAA: 0.88). These consistently high correlations suggest that robustness characteristics established during pre-training largely persist through linear probing.
The performance drop analysis reveals method-specific robustness characteristics. MoCoV3 consistently exhibits smaller performance drops across both ImageNet (IAA: 54\%, UAP: 36\%) and linear evaluation (IAA: 56\%, UAP: 43\%), indicating its contrastive learning approach with momentum encoders develops features with inherently better transferable robustness properties. In contrast, supervised pre-training and clustering-based approaches like SwAV experience more severe drops in both settings, suggesting these methods may create more brittle representations.

The high correlation in adversarial vulnerability between ImageNet and linear evaluation indicates that linear probing provides a reliable assessment of a pre-trained model's downstream robustness characteristics without requiring extensive adaptation. This finding has practical implications for model selection, suggesting that robustness evaluations on ImageNet can effectively predict linear transfer performance. The full results are in Section 4.8 of the Supp. Mat..

\textbf{FINETUNE}.
Fine-tuning reveals distinct transfer patterns by attack type. For IAA, the correlation between ImageNet and fine-tuned vulnerability nearly vanishes for performance drops (Spearman: 0.12, Pearson: 0.36), indicating that fine-tuning substantially reshapes defense against instance-specific attacks. Conversely, UAP vulnerability correlation remains high (Spearman: 0.88, Pearson: 0.93), suggesting that susceptibility to universal perturbations is more persistently encoded in the network regardless of parameter adaptation.
This attack-dependent correlation disparity reveals that universal perturbation vulnerabilities, which exploit systematic weaknesses across the feature space, are deeply encoded in the network's architecture and learning approach, while instance-specific vulnerabilities are more malleable through fine-tuning.
The absolute magnitude of performance drops diminishes considerably after fine-tuning across all methods (IAA: 49\% vs. 60\% in linear, UAP: 43\% vs. 49\% in linear), highlighting fine-tuning's effectiveness in mitigating vulnerability. MoCoV3 continues to demonstrate superior robustness with the smallest drops (IAA: 45\%, UAP: 38\%), while BYOL shows dramatic improvement from ImageNet to fine-tuning for IAA attacks (from 56\% to 47\%).
These findings suggest different mechanisms for robustness transfer: UAP vulnerability appears tied to fundamental architectural and algorithmic properties that persist across transfer paradigms, while instance-specific attack vulnerability depends more on the fine-tuning process than initial representation properties. This distinction may be valuable for practical applications, suggesting that pre-training method selection strongly impacts universal attack robustness, while defense against instance-specific attacks can be substantially improved through appropriate fine-tuning strategies. The full results are in Section 4.9 of the Supp. Mat..

\textbf{Attributing Adversarial Vulnerability.}
Having established the correlations between ImageNet and transfer learning robustness patterns, we now seek to understand the underlying mechanisms causing these disparities across models and training paradigms. Specifically, we investigate whether adversarial vulnerability stems primarily from the backbone feature extractor or the classification head and how this attribution changes between linear probing and fine-tuning.

Our analysis investigates the relationship between adversarial robustness and component-specific vulnerability under $\text{FGSM}_1$ attack (detail in Supp. Mat.), focusing on two metrics: the Direction Ratio $D = \frac{\text{mean}(1-\cos(\mathbf{l}, \mathbf{l}_{adv}))}{\text{mean}(1-\cos(\mathbf{f}, \mathbf{f}_{adv}))}$, which compares directional shifts in logits ($\mathbf{l}$) versus backbone features ($\mathbf{f}$), and the Magnitude Ratio $M = \frac{\text{mean}(\|\mathbf{l} - \mathbf{l}_{adv}\|_2)}{\text{mean}(\|\mathbf{f} - \mathbf{f}_{adv}\|_2)}$, quantifying relative sensitivity to perturbation magnitudes. We fit linear regression models using these ratios to predict adversarial accuracy drop across models.

For probed models, Direction Ratio strongly correlates with vulnerability ($R^2 = 0.56$--$0.85$), suggesting that directional instability in the head dominates FGSM robustness. SwAV, with $D=3.1$ on CIFAR-10, suffers a 79\% accuracy drop, while SimCLR ($D=1.5$) drops 58\%. This may arise from the linear head's limited capacity to compensate for adversarial noise, amplifying directional shifts in logit space.

After fine-tuning, this correlation weakens significantly ($R^2 = 0.05$--$0.52$): SwAV achieves only a 17\% drop despite retaining $D=1.7$, implying that fine-tuning enables the head to stabilize predictions even under directional feature shifts. This pattern aligns with our transfer learning analysis, where fine-tuning disrupted the correlation between ImageNet and downstream instance-specific attack robustness. Regression figures are in Section 2 of the Supp. Mat.

As in Table~\ref{tab:component-analysis}, self-supervised models exhibit higher pre-fine-tuning Magnitude Ratios (e.g., SwAV $M=17.4$ on ImageNet vs. supervised $M=3.0$), potentially due to their reliance on globally normalized feature spaces, where FGSM perturbations propagate more aggressively to logits. Fine-tuning reduces $M$ across models (SwAV $M=2.7$ post-FT), aligning with improved robustness, possibly by suppressing logit magnitude distortions.

These patterns are specific to FGSM, as its single-step gradient reliance may emphasize head-layer instability, whereas iterative attacks could exploit backbone vulnerabilities differently. While our analysis tentatively links directional logit shifts ($D$) and magnitude sensitivity ($M$) to robustness, dataset-dependent variations—such as weaker correlations on CIFAR-100 FT—highlight the need for broader evaluation across threat models to generalize these insights.

We also measured inter-class and intra-class distances for both probed and fine-tuned CIFAR-10 representations; however, these metrics did not yield meaningful correlations that explain the observed differences in adversarial robustness. The detailed regression figures, inter/intra-class distance measurements, and t-SNE~\cite{JMLR:v9:vandermaaten08a} visualizations of feature spaces are provided in Section 3 of the Supp. Mat.

\subsection{Segmentation}
Unlike in classification, we observe no strong correlation between ImageNet robustness and segmentation performance across both PASCAL VOC and CityScapes datasets. Self-supervised approaches demonstrate competitive performance in clean conditions, with VICReg leading on PASCAL VOC (77.89 mIOU with frozen backbone) and SwAV on CityScapes (66.48 mIOU). All models suffer catastrophic performance degradation (74-84\%) under adversarial attacks regardless of training methodology. This uniform degradation pattern suggests attacks primarily target segmentation modules rather than backbones.

Our experiments with frozen versus unfrozen backbones reveal that frozen backbones generally achieve both higher clean performance and slightly better adversarial robustness on PASCAL VOC. The supervised model's exception to this pattern stems from using the standard MMSegmentation model for the unfrozen case due to reproduction challenges. These findings indicate that while SSL models produce competitive segmentation performance in clean conditions, they offer minimal advantage in terms of adversarial robustness for segmentation tasks, unlike their significant impact on object recognition. This highlights the need for robustness techniques specifically designed for segmentation architectures rather than focusing solely on backbone improvements. The full results are in Section 4.2 of the Supp. Mat.

\subsection{Detection}
Detection results show distinct patterns from recognition and segmentation tasks. On INRIA Person with unfrozen backbones, SwAV demonstrates the highest robustness (72\% decrease), while the Supervised model shows the poorest performance (89\% decrease). With frozen backbones, Barlow Twins and DINO lead in robustness (both 75\% decrease), while VICReg becomes unexpectedly vulnerable (84\% decrease), contradicting its recognition performance.

COCO dataset results show improved robustness overall, with Barlow Twins maintaining the strongest performance (50\% decrease) and the Supervised model remaining the least robust (67\% decrease). This consistent weakness in the Supervised model suggests inherent robustness benefits from self-supervised pretraining.

These findings indicate that task-specific architectures significantly influence adversarial robustness, and robustness in ImageNet classification doesn't necessarily transfer to detection tasks. This highlights the importance of task-specific evaluations and suggests that backbone architecture becomes less critical than overall model design for downstream applications. The full results are in Section 4.3 of the Supp. Mat.

\section{Conclusion}
Our evaluation shows that SSL models generally offer greater adversarial robustness than supervised counterparts in image classification, with MoCoV3 performing particularly well—likely due to its momentum encoder and prediction head. This robustness advantage is less evident in segmentation and detection, where task-specific architectures play a larger role. Architectural effects also vary by objective: DINO benefits from ViTs, while MoCoV3 performs better with CNNs. Extended training does not compromise robustness and may slightly improve it. We find that vulnerability to UAPs transfers more consistently across learning paradigms, while susceptibility to IAAs can be mitigated through fine-tuning. These results highlight the complex interplay between SSL objectives, architectures, and downstream tasks, underscoring the need for further study. We hope our findings support continued progress toward more robust visual representation learning.

\section*{Acknowledgements}
This work was funded by the KUIS AI Center at Koç University, Turkey.

{
    \small
    \bibliographystyle{ieeenat_fullname}
    \bibliography{main}
}
\clearpage  
\onecolumn
\section*{Supplementary Materials}
\setcounter{section}{0}

\section{Adversarial Attacks}
\subsection{Instance Adversarial Attacks}\label{app:iaa}
Instance adversarial methods, or per-instance generation, involve crafting distinct perturbations for each individual image within the dataset on which the model has been trained or fine-tuned. The generation of these perturbations relies on various techniques, which are determined by the specific goals of the attack, the level of access granted to the model—such as full access to model weights, predictions alone, or prediction scores (logits)—and the distance metrics employed. While multiple classification schemes for adversarial attacks exist, we adopt the widely accepted taxonomy for clarity and consistency.

White-box attacks, in this context, presume complete access to the model, including its architecture and parameters. The primary approach utilizes the gradients derived from the loss function to generate adversarial perturbations. These perturbations are then applied to the image within the constraints of specific distance metrics, such as \( l_0 \), \( l_1 \), \( l_2 \), or \( l_\infty \). Specifically, \( l_0 \) measures the number of altered pixels, \( l_1 \) quantifies the total absolute difference between images, \( l_2 \) computes the Euclidean distance, and \( l_\infty \) captures the magnitude of the largest perturbation applied to any pixel.

Gradient-based methods exploit the gradient of the neural network’s loss function with respect to the input data, strategically altering the input to increase the loss and induce misclassification. The foundational work in this domain is attributed to the Fast Gradient Sign Method (FGSM) \citep{goodfellow2014explainingandharnessingadversarialexamples}, which represents the first successful application of gradient-based adversarial perturbations. The Fast Gradient Sign Method (FGSM) computes the gradient of the cross-entropy loss with respect to the input image to determine the perturbation direction that maximizes the loss. The adversarial example is then generated by applying this perturbation in a single step. Due to this one-step update, FGSM is classified as a single-step adversarial attack.

  Other FGSM methods apply gradeient update more than one time (iteratively) using much smaller step sizes to remain in predetermined \( lp\) ball. PI-FGSM modify the gradient update rule by focusing on patch-based rather than pixel-wise perturbations \citep{gao2020patchwiseattackforfoolingdeepneuralnetwork, gao2020patchwise++perturbationforadversarialtargetedattacks}. DI-FGSM employs random padding and resizing operations to enhance data input for auxiliary models \citep{xie2019improvingtransferabilityofadversarialexamples}. TAP also tries to increase cross-model transferability by introducing distance maximization between intermediate feature maps of the adversarial and benign datapoints. It also regularize the images to reduce high frequency perturbations as they claim Convolution may act as a smoother, and it will increase the black-box transferability performance of perturbation \citep{zhou2018transferableadversarialperturbations}. TI-FGSM is also another iterative FGSM method, which uses translated version of benign input to enhance black-box transferability of adversarial attacks to models which are defended with various methods. TIFGSM suggest that these defended models uses different discriminative regions than the model on which adversarial examples are generated, which makes these adversarial examples less effective. TIFGSM uses diversified (shifted and padded) inputs which are obtained by approximating the gradients with convolution kernels \cite{dong2019evadingdefensestotransferableadversarialexamples}.

On the other hand, Projected Gradient Descent (PGD) employs an iterative approach, projecting updates back onto the \( l_\infty \) ball of the original data point to generate adversarial perturbations \citep{madry2017towardsdeeplearningmodelsresitanttoadversarialattacks}. The key distinction between PGD and FGSM variants, lies in the fact that PGD treats each iteration as a solution to the same optimization problem. PGD ensures that each iterative step remains within the neighborhood of the original data point, while iterative FGSM methods use the newly generated steps to continue further processing.

As with FGSM-based adversarial attacks, several improvements have been made to PGD to address specific needs \citep{madry2017towardsdeeplearningmodelsresitanttoadversarialattacks}. For example, \text{PGD-$l_2$} incorporates the \( l_2 \) norm instead of the \( l_\infty \) norm to better fool target models \citep{madry2017towardsdeeplearningmodelsresitanttoadversarialattacks}.

The Jitter attack introduces a novel objective function for adversarial perturbation generation, departing from the conventional Cross-Entropy objective. The study suggests that many adversarial attacks predominantly fool a limited set of classes rather than broadly deceiving the entire model. The proposed objective seeks to enhance the fooling rate across a wider range of classes, aiming for more generalized misclassification \citep{schwinn2023exploringmissclassificationsofrobustneuralnetworks}.

Improving the transferability of per-instance attacks can, however, lead to reduced effectiveness against auxiliary models, and vice-versa \citep{tramer2017ensembleadversarialtraining, gao2020patchwiseattackforfoolingdeepneuralnetwork}. Therefore, various strategies have been proposed to optimize attack performance based on the level of access to the target model.

In contrast, optimization-based attacks approach the generation of adversarial examples as an optimization problem, where a specific objective is minimized subject to given constraints. While gradient-based methods update images directly using gradient information and typically rely on the \( l_\infty \) norm as a boundary, optimization-based methods employ a more formal problem definition that allows for the use of advanced optimization techniques. Consequently, different \( lp \) norm is utilized in these methods alongside with \( l_\infty\) norms.

 \citep{carlini2017towardsevaluatingtherobustnessofneuralnetworks} constructs a minimization problem—focusing on minimizing the distance between adversarial examples and the original data points across several \( l \) norms—to develop the CW attack, one of the most prominent adversarial attack methods.

The EADEN attacks adopt a similar approach to the CW attack but introduce a modification to the loss function by incorporating an additional \( l_1 \) distance term in the minimization problem. The \( l_1 \) distance, which measures the total variation of the perturbation, promotes sparsity in the adversarial perturbation. While sparsity is not widely employed in adversarial example generation, it is commonly used in image denoising and restoration techniques. These methods utilize the Iterative Shrinkage-Thresholding Algorithm (ISTA) to solve the corresponding optimization problem \citep{chen2018eadelasticnetattackstodeepneuralnetworks}.



While white-box attacks exploit full access to the model, this is often not a realistic scenario. In many cases, model weights are not shared, or gradient information is unavailable. Although efforts have been made to enhance cross-model transferability, as discussed previously, there are also specific attack schemes designed to target models in black-box settings. For example, the Square Attack leverages random search combined with model scores—probability distributions over class predictions—to generate perturbations. In essence, the algorithm makes random modifications to the input data and retains changes that yield progress toward the objective function \citep{andriushchenko2020squareattackaqueryefficientblackboxadversairalattack}.

These black-box attacks leverages gradient-free approaches remain relatively underexplored. For instance, the Simultaneous Perturbation Stochastic Approximation (SPSA) method estimates gradients by perturbing the input in random directions, enabling the approximation of gradients for objectives that cannot be differentiated analytically. This approach offers deeper insights into the model's behavior, with the paper also claiming that the stochastic perturbations introduced by sampling allow algorithms to converge toward a global minimum \citep{uesato2018adversarialriskandthedangersofevaluatingweakattacks}.

Among black-box attacks, some methods focus on \( l_0 \) norm-based perturbations. Pixle, for instance, is a black-box attack that utilizes random search and the \( l_0 \) norm, altering a small number of pixels to generate adversarial examples \citep{pomponi2022pixleafastandeffectiveblackboxattack}. On a more constrained scale, the OnePixel attack modifies only a single pixel, maintaining an \( l_0 \) norm of 1, and despite its simplicity, it is capable of fooling models to some extent. However, it is less effective than other methods due to its significant restrictions. This raises important questions about our understanding of Deep Neural Networks and their vulnerability to minimal perturbations \citep{su2019onepixelattackforfoolingdeepneuralnetwork}.

\subsection{Universal Adversarial Perturbations}\label{app:uap}

The Universal Adversary (UAP) represents a singular perturbation crafted for an entire image dataset. The rationale behind UAP is to identify a perturbation, subject to specified constraints, capable of deceiving the model across a majority of images in the dataset, as initially demonstrated by \citep{moosavi2017universaladversarialperturbations}, which utilizes DeepFool to create an average perturbation for the entire dataset. It has been empirically observed that universal adversaries exhibit heightened transferability across diverse models and datasets compared to instance methods. UAP's are important as they are independent from the input - to some extend - they reveal intrinsic chracteristics of models of interest \citep{chaubey2020universaladversarialperturbations, ye2023fguapfeaturegatheringuniversaladversarialperturbation}.

Two primary techniques are employed for crafting UAPs: (1) generation with generative models, as evidenced by works such as \citep{hayes2018learninguniversaladversarialperturbationswithgenerativemodels, mopuri2018nagnetworkforadversarygeneration}, and (2) learning a perturbation designed to disrupt the representations acquired by the models.

UAPs can be further categorized into two classes: data-dependent attacks, which require a comprehensive and general dataset that the attacker seeks to compromise (e.g., ImageNet), and data-independent attacks, which do not rely on any specific dataset. 

The first example of UAP, referred to here as UAP-DeepFool (to avoid confusion with the broader class of UAP attacks), utilizes the DeepFool per-instance adversarial attack method which computes perturbations by manipulating the geometry of decision boundaries. UAP-DeepFool iteratively determines the worst-case direction for each data point, and aggregating the results into a universal perturbation - if it is succesfull -, which is then projected onto an \( l_\infty \) ball \citep{moosavi2017universaladversarialperturbations}. Following this work, UAPEPGD replaces the DeepFool approach with Projected Gradient Descent (PGD), an optimization-based adversarial attack method, to craft stronger adversarial examples \citep{deng2020universaladversarialattackviaenhanvedprojectedgradientdescent}.

ASV - to our best knowledge - is the first UAP that does not require label information, relying solely on images to generate UAPs. Adversarial Semantic Vectors (ASVs) represent one of the first UAP methods that do not require label information, relying solely on images to generate UAPs. The study suggests that since adversarial perturbations typically exhibit small magnitudes, perturbations in the non-linear maps computed by deep neural networks (DNNs) can be approximated using the Jacobian matrix \citep{khrulkov2018artofsingularvectorsanduniversaladversarialperturbations}. Similarly, the STD (Dispersion Reduction) attack seeks to reduce the "contrast" of the internal feature map by targeting the lower layers of Convolutional Neural Networks (CNNs). These lower layers typically detect simple image features such as edges and textures, which are common across datasets and CNN models. By reducing the contrast (measured as the standard deviation of feature maps), the resulting images become indistinguishable to the model \citep{lu2020enhancingcrosstaskblackboxtransferabilityofadversarialexamples}.

Self-Supervised Perturbation (SSP) takes a different approach, arguing that adversarial examples generated through gradients using labels fail to capture intrinsic properties of models. SSP aims to maximize "feature distortion," the changes in the network's internal representation caused by adversarial examples compared to the original image, in order to fool subsequent layers in the model \citep{naseer2020selfsupervisedapproachforadversarialrobustness}.

FG-UAP builds upon this by exploiting a phenomenon referred to as "Neural Collapse," where, as noted, different class activations converge to class means, allowing a single common perturbation to fool the model across a wide range of images. This collapse happens primarily in the final layers of the model, and FG-UAP targets these regions to generate effective UAPs \citep{ye2023fguapfeaturegatheringuniversaladversarialperturbation}.

Another label-independent UAP method, L4A, focuses on the success of adversarial perturbations during cross-finetuning. L4A targets the lower layers of models, which remain more stable during finetuning (as they detect simple features), and utilizes the Frobenius norm for optimization, with variants such as L4A-base, L4A-fuse, and L4A-ugs. L4A-base attacks the lowest layer, L4A-fuse attacks lowest 2 layers and L4A-ugs uses samples from a Gaussian distribution where mean and standard deviation is in close range of downstream task \citep{ban2022pretrainedadversarialperturbations}.

Data-independent UAP methods do not utilize any dataset for adversarial perturbation generation, instead focusing on the intrinsic characteristics of models. Fast Feature Fool (FFF) was the first adversarial attack method that did not use a dataset. It aims to disrupt the features learned at individual CNN layers, proposing that non-discriminative activations can lead to eventual misclassification. FFF over-saturates the learned features at multiple layers, misleading subsequent layers in the network \citep{mopuri2017fastfeaturefooladataindependentapproachtouniversaladversarialperturbations}. Following that work GD-UAP, changes the objective a little bit and add other variations such as "mean-std" and "sampled" versions to improve perturbation performance. The "mean-std" variant uses the mean and standard deviation of the test dataset to better align perturbations with dataset characteristics to prevent perturbation dataset mismatch, while the "sampled" version employs a small sample from the dataset to capture its statistics and semantics \citep{mopuri2018generalizabledatafreeobjectiveforcraftinguniversaladversarialperturbations}. In our work, we have also integrated "mean-std" and "one-sample" versions of GD-UAP to FFF, since they are highlt similar as GD-UAP is a follow-up work FFF. PD-UAP, another data-independent method, focuses on predictive uncertainty rather than any specific image data, aligning perturbations with task-specific objectives \citep{mopuri2017fastfeaturefooladataindependentapproachtouniversaladversarialperturbations}.

To accommodate both Vision Transformers (ViTs) and ResNets, we have adapted some of these attacks, originally designed for CNNs, to work with ViTs. For low-level layer attacks, we applied them to the first few blocks of the ViT model, following methods like SSP and L4A. For FFF, which typically uses mean of ReLU activations and a logarithmic operation, we modified the procedure to suit ViTs, which employ GeLU activations (capable of taking values below zero), by applying an absolute value operator between the mean and logarithmic functions. In conducting these experiments, we strove to maintain fair comparisons and minimized the introduction of tweaks to the original methodologies.

\subsection{FGSM and PGD versions}\label{app:version}

\begin{table}[ht]
    \centering
    \small
    \begin{tabular}{c|c|c|c|c}
        \toprule
        \textbf{Attack Version} & \textbf{Attack Type} & $\varepsilon$ & \textbf{Step Count} & \textbf{Norm} \\ 
        \midrule
        $FGSM_1$        & FGSM        & 0.25          & -          & $\infty$ \\ 
        $FGSM_2$        & FGSM        & 1             & -          & $\infty$ \\ 
        $PGD_1$            & PGD         & 0.25          & 20         & $\infty$ \\ 
        $PGD_2$            & PGD         & 1             & 20         & $\infty$ \\ 
        $PGD_3$            & PGD         & 0.25          & 40         & $\infty$ \\ 
        $PGD_4$            & PGD         & 1             & 40         & $\infty$ \\ 
        $PGD_5$           & PGD       & 0.5           & 40         & $\|\cdot\|_2$ \\ 
        \midrule
    \end{tabular}
    \caption{Hyperparameters of the different FGSM and PGD attacks that we use in ImageNet and transfer learning.}
    \label{tab:adversarial_attacks}
\end{table}

\subsection{Categories}
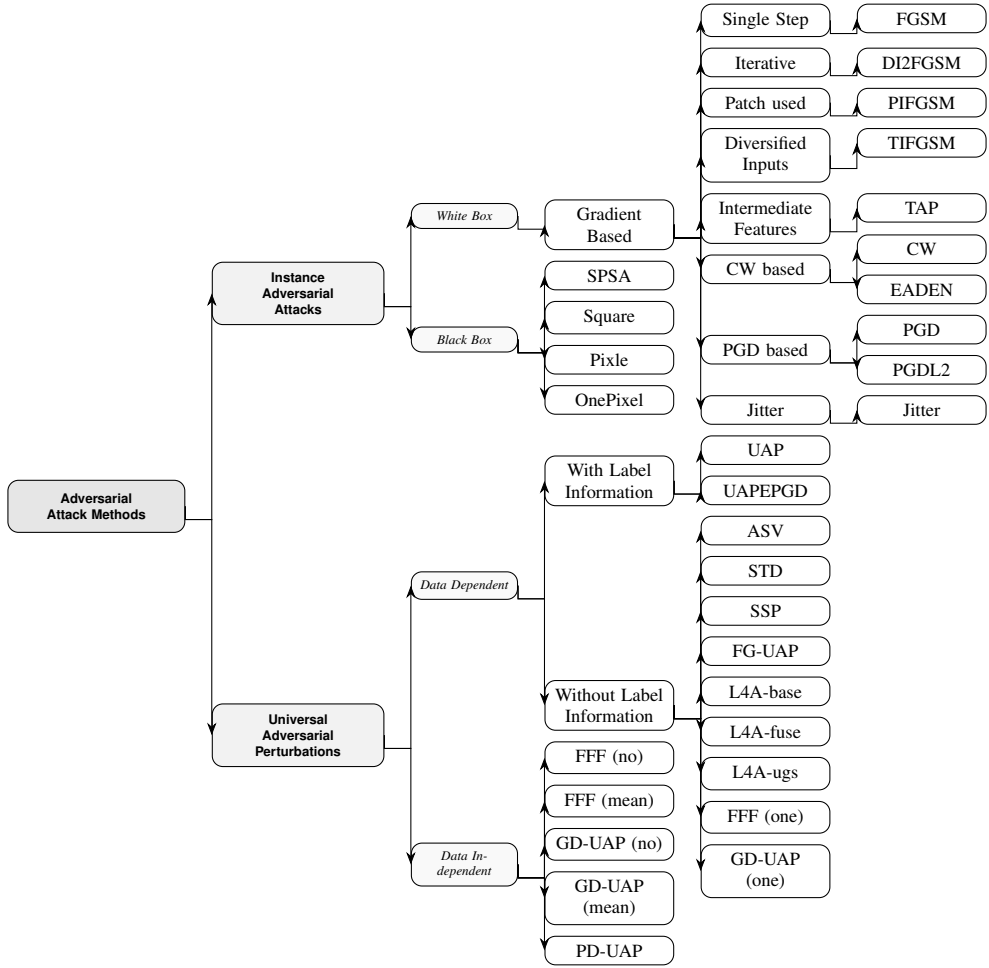
\begin{figure*}[h]
\centering
\tiny
\begin{forest}
  for tree={
    font=\scriptsize,
    draw,
    rounded corners,
    node options={align=center, inner sep=3pt},
    text width=1.5cm,
    edge={-Stealth},
    l sep=10pt,
    s sep=4pt,
    grow'=0,
    edge path={
      \noexpand\path [draw, \forestoption{edge}] (!u.parent anchor) -- +($(0,-5pt)$) -| (.child anchor)\forestoption{edge label};
    },
    parent anchor=east,
    child anchor=west,
    if level=0{
      text width=2.0cm,
      inner sep=5pt,
      font=\sffamily\bfseries,
      fill=gray!20,
    }{},
    if level=1{
      text width=2cm,
      inner sep=4pt,
      font=\sffamily\bfseries,
      fill=gray!10,
    }{},
    if level=2{
      font=\itshape,
      text width=1.2cm,
      fill=gray!5,
    }{},
  }
  [Adversarial\\Attack Methods
    [Instance\\Adversarial\\Attacks
      [White Box
        [Gradient Based
          [Single Step
            [FGSM]
          ]
          [Iterative
            [DI2FGSM]
          ]
          [Patch used
            [PIFGSM]
          ]
          [Diversified\\Inputs
            [TIFGSM]
          ]
          [Intermediate\\Features
            [TAP]
          ]
          [CW based
            [CW]
            [EADEN]
          ]
          [PGD based
            [PGD]
            [PGDL2]
          ]
          [Jitter
            [Jitter]
          ]
        ]
      ]
      [Black Box
        [SPSA]
        [Square]
        [Pixle]
        [OnePixel]
      ]
    ]
    [Universal\\Adversarial\\Perturbations
      [Data Dependent
        [With Label\\Information
          [UAP]
          [UAPEPGD]
        ]
        [Without Label\\Information
          [ASV]
          [STD]
          [SSP]
          [FG-UAP]
          [L4A-base]
          [L4A-fuse]
          [L4A-ugs]
          [FFF (one)]
          [GD-UAP (one)]
        ]
      ]
      [Data Independent
        [FFF (no)]
        [FFF (mean)]
        [GD-UAP (no)]
        [GD-UAP (mean)]
        [PD-UAP]
      ]
    ]
  ]
\end{forest}
\caption{Classification of Selected Adversarial Attack Methods}
\end{figure*}

\section{DINOv2 and MAE}
The primary challenge in comparing DINOv2 to other SSL models lies in its training data. DINOv2 is trained on a significantly larger dataset containing over 140 million images, which introduces a major discrepancy in scale compared to other models. Additionally, only the largest DINOv2 model is trained in a fully self-supervised manner; the smaller variants are obtained via distillation from this larger model. As a result, they are not inherently self-supervised and tend to perform worse than the original, potentially due to these non-uniform training procedures.

In contrast, MAE is entirely trained on ImageNet, but its performance under linear probing is relatively weak. To mitigate this, the authors apply partial fine-tuning to improve results. This complicates direct comparisons with models evaluated purely via linear probing, though MAE still achieves roughly average performance overall.

Given the diversity of training and evaluation setups, it is difficult to make definitive comparisons between these models. Nevertheless, we include them in our analysis to ensure broader coverage. The complete results can be found in Section~\ref{res:dinov2}.

\section{Checkpoints and Repositories}
The self-supervised learning checkpoints used in our experiments are from the following repositories: \href{https://github.com/facebookresearch/barlowtwins}{Barlow Twins}, \href{https://github.com/google-deepmind/deepmind-research/tree/master/byol}{BYOL}, \href{https://github.com/facebookresearch/dino}{DINO}, \href{https://github.com/facebookresearch/dinov2}{DINOv2}, \href{https://github.com/facebookresearch/moco-v3}{MoCo v3}, \href{https://github.com/google-research/simclr}{SimCLR}, \href{https://github.com/facebookresearch/swav}{SwAV}, \href{https://github.com/facebookresearch/vicreg}{VICReg}, and \href{https://github.com/facebookresearch/mae}{MAE}.

We have used \href{https://github.com/Harry24k/adversarial-attacks-pytorch}{Torchattacks} library for IAAs and the repository of  \citet{ban2022pretrainedadversarialperturbations} for UAPs

\newpage
\section{Regression Visualization}
\begin{figure}[htbp]
    \centering
    \includegraphics[width=\textwidth]{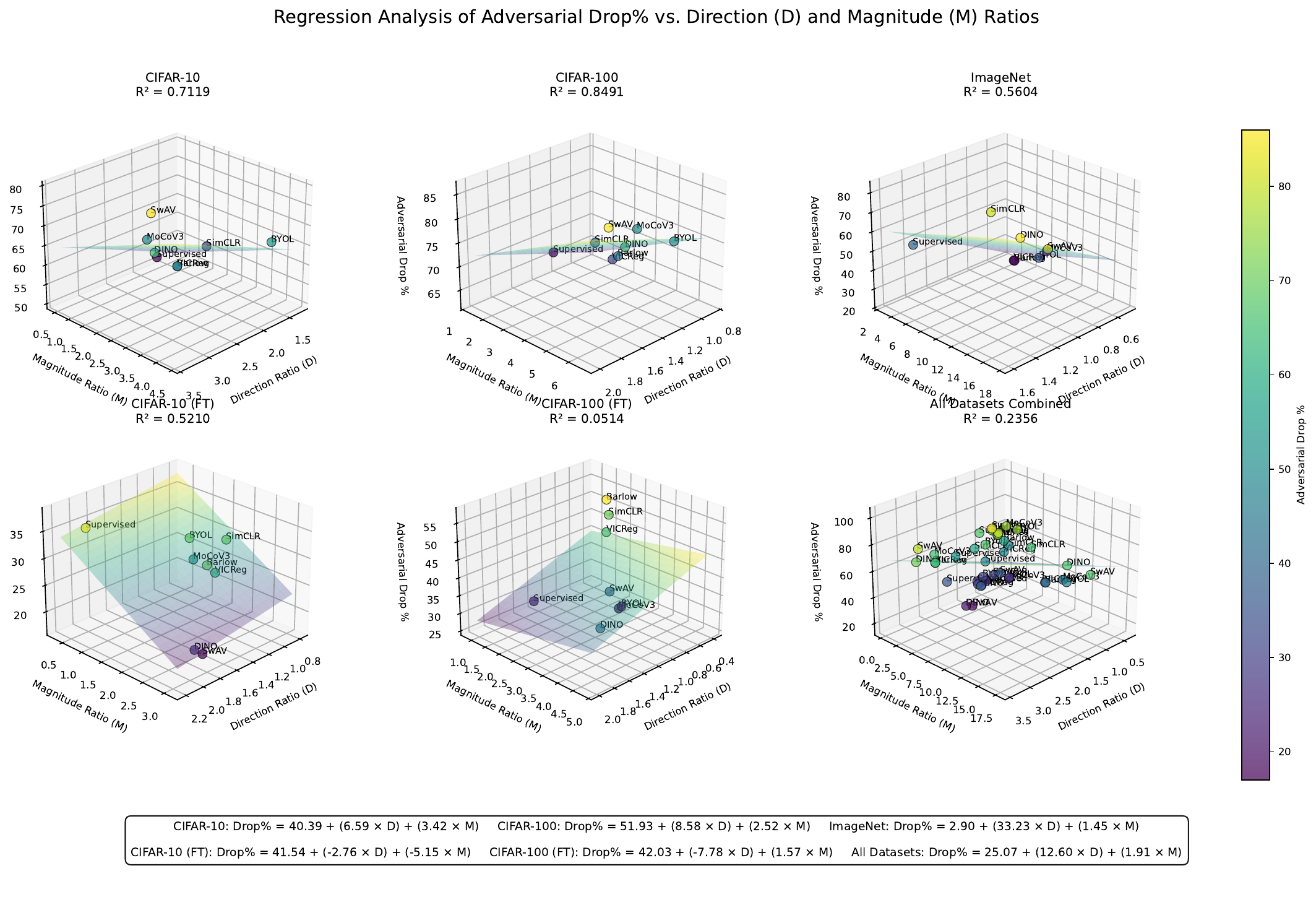}
    \caption{Regression analysis of adversarial performance drop percentage as a function of Direction Ratio (D) and Magnitude Ratio (M) across different datasets. The 3D plots show the fitted regression planes for CIFAR-10, CIFAR-100, and ImageNet datasets (top row), their fine-tuned counterparts (bottom left and center), and the combined analysis (bottom right). Each data point represents a different self-supervised learning method, color-coded by drop percentage. The R$^2$ values highlight the strong explanatory power for non-fine-tuned datasets (0.56-0.85) compared to fine-tuned ones (0.05-0.52). Note the positive coefficients for D and M in non-fine-tuned scenarios versus negative coefficients in fine-tuned contexts, suggesting fundamentally different robustness mechanisms. Regression equations are displayed below each corresponding plot.}
    \label{fig:regression_analysis}
\end{figure}
\newpage
\section{t-SNE and Inter/Intra Class Distance Visualization}
\begin{figure*}[h]
    \begin{subfigure}[h]{0.35\textwidth}
        \includegraphics[width=\textwidth]{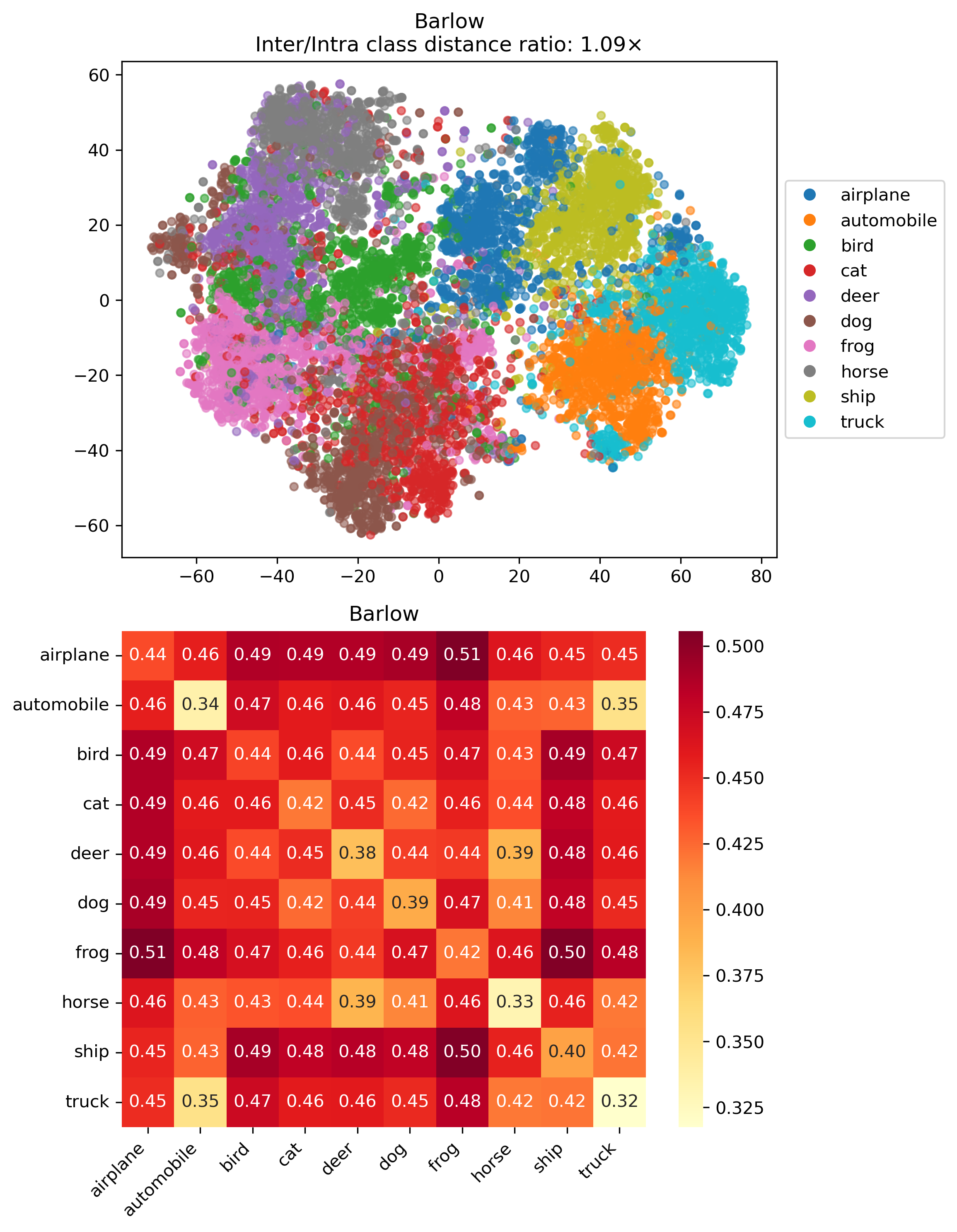}
        \caption{}
        \label{fig:subfig1}
    \end{subfigure}
    \hfill
    \begin{subfigure}[h]{0.35\textwidth}
        \includegraphics[width=\textwidth]{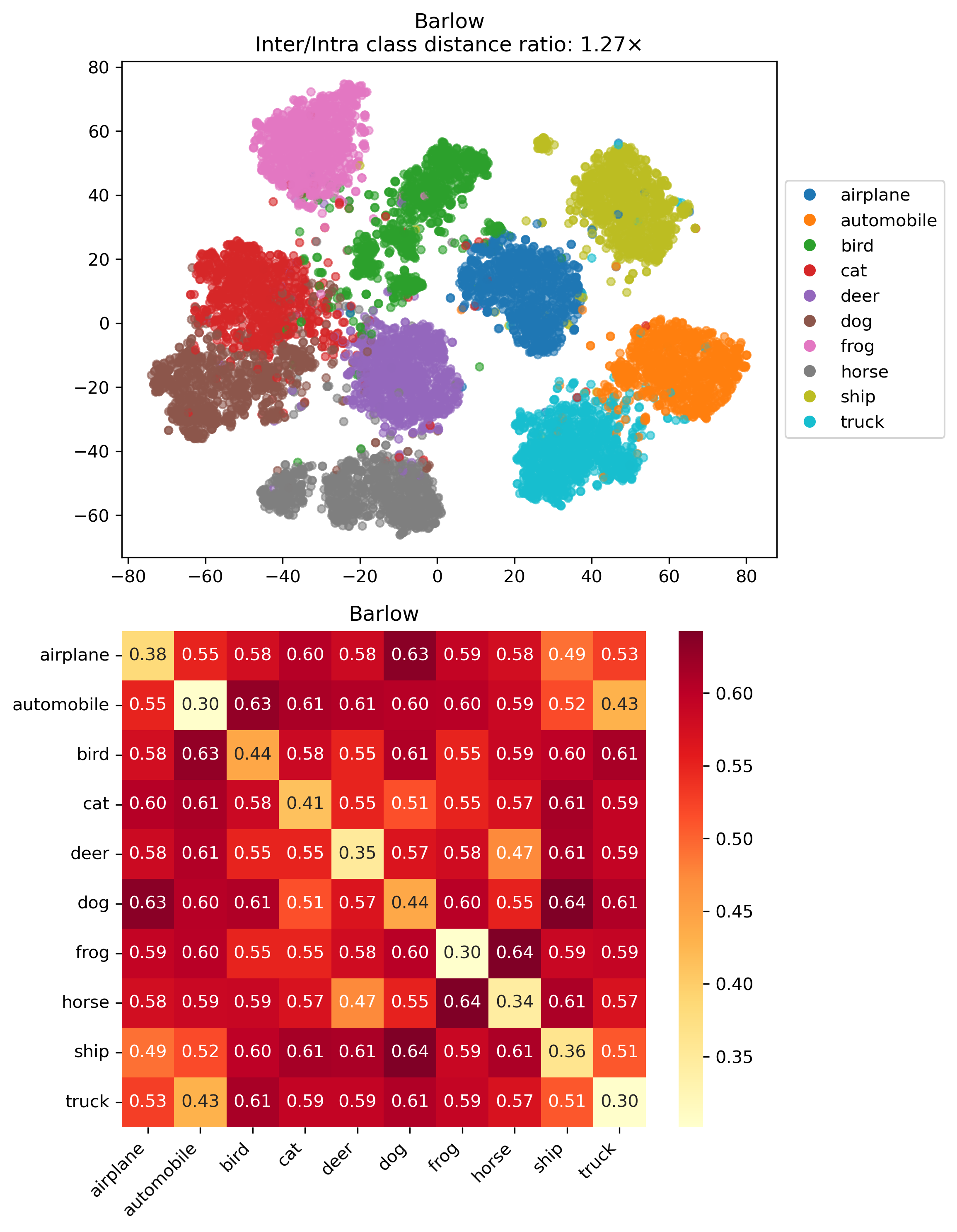}
        \caption{}
        \label{fig:subfig2}
    \end{subfigure}
    
    \vspace{0.5em}
    \begin{subfigure}[h]{0.35\textwidth}
        \includegraphics[width=\textwidth]{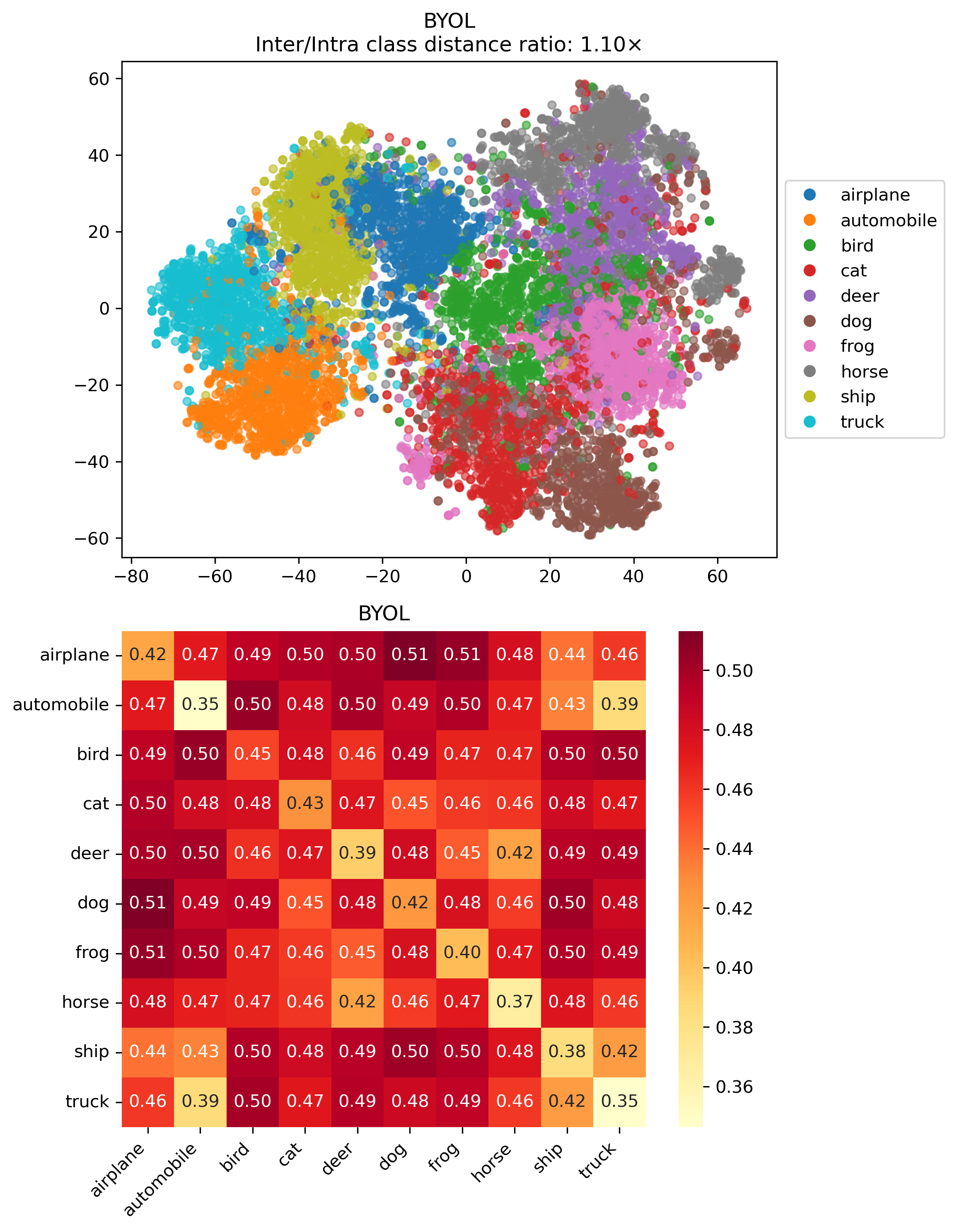}
        \caption{}
        \label{fig:subfig3}
    \end{subfigure}
    \hfill
    \begin{subfigure}[h]{0.35\textwidth}
        \includegraphics[width=\textwidth]{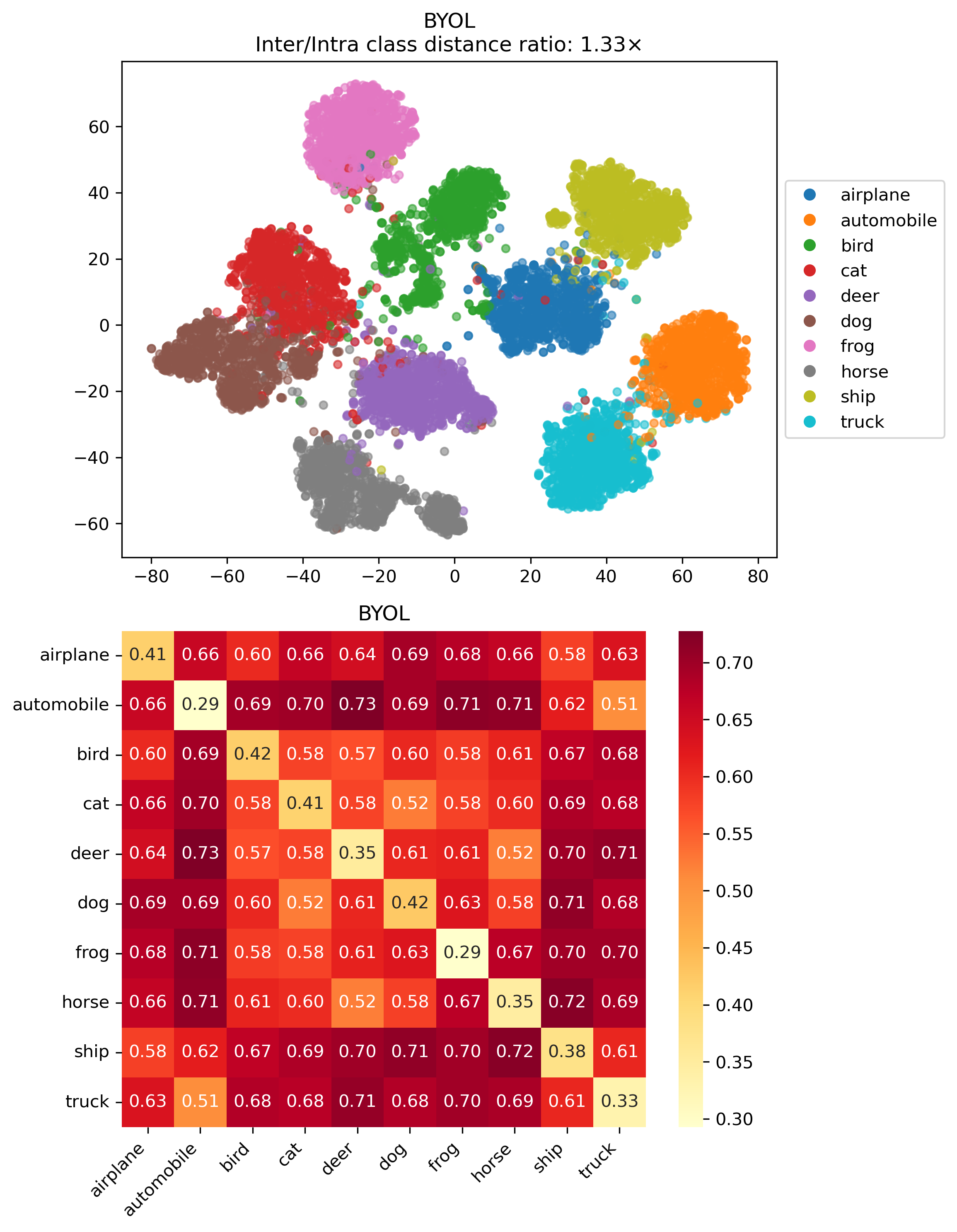}
        \caption{}
        \label{fig:subfig4}
    \end{subfigure}

    \caption{Comparison of feature representations for CIFAR-10 images using ResNet50 with different self-supervised learning (SSL) methods. \textbf{Top row:} Results from Barlow. \textbf{Bottom row:} Results from BYOL. For each SSL method, \textbf{left} panels show results from \textbf{probed} models and \textbf{right} panels show results from \textbf{fine-tuned} models. Within each panel, the \textbf{upper} plots display t-SNE visualizations of the 2048-dimensional feature vectors using Euclidean distance, with points colored by class and Inter/Intra class distance ratios indicated. The \textbf{lower} plots show the corresponding class-wise distance matrices computed using cosine similarity, with the average distances between samples from each pair of classes. Higher Inter/Intra class distance ratios indicate better class separation in the feature space.}

    \label{fig:main}
\end{figure*}

\begin{figure*}[h]
    \begin{subfigure}[h]{0.35\textwidth}
        \includegraphics[width=\textwidth]{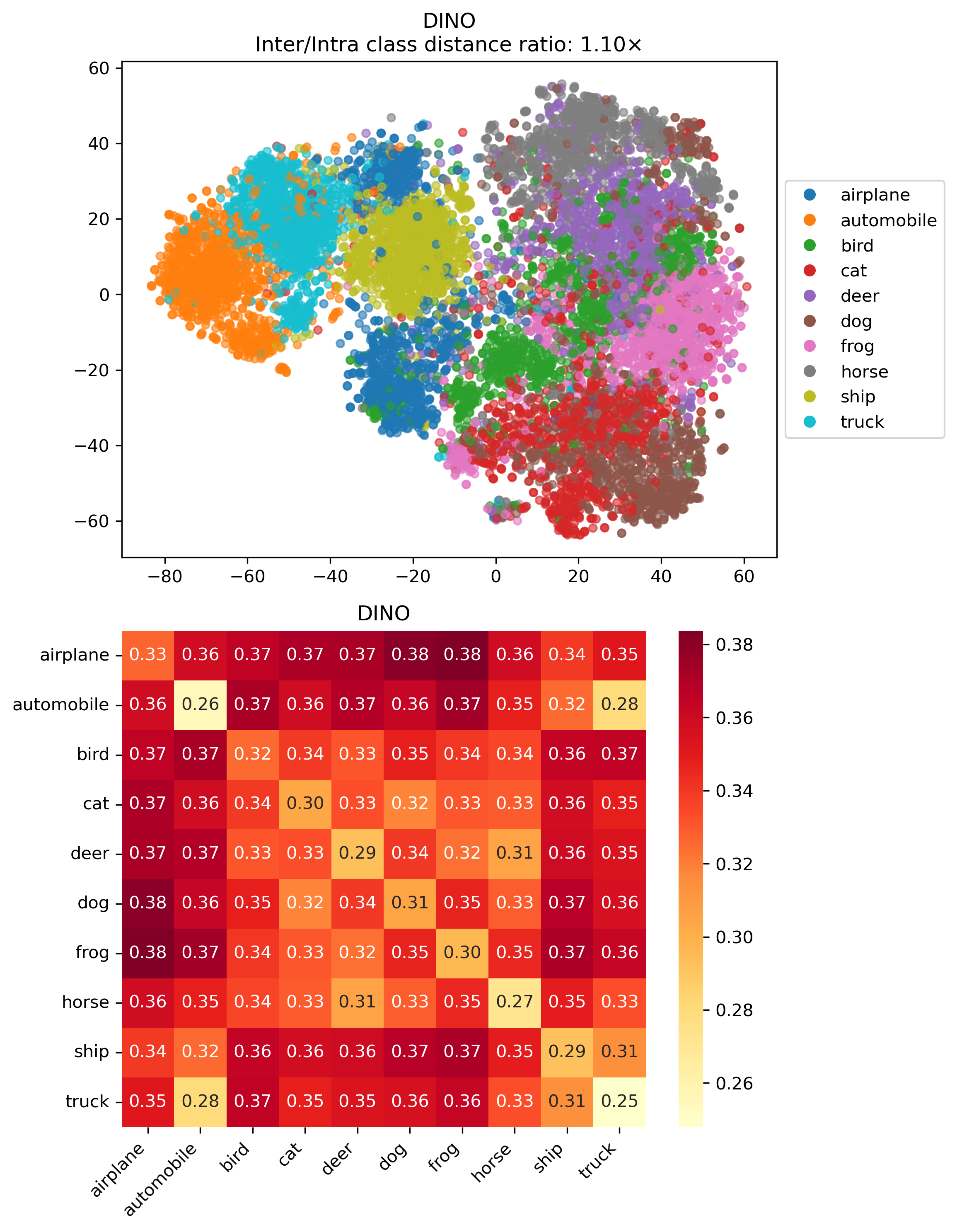}
        \caption{}
        \label{fig:subfig1}
    \end{subfigure}
    \hfill
    \begin{subfigure}[h]{0.35\textwidth}
        \includegraphics[width=\textwidth]{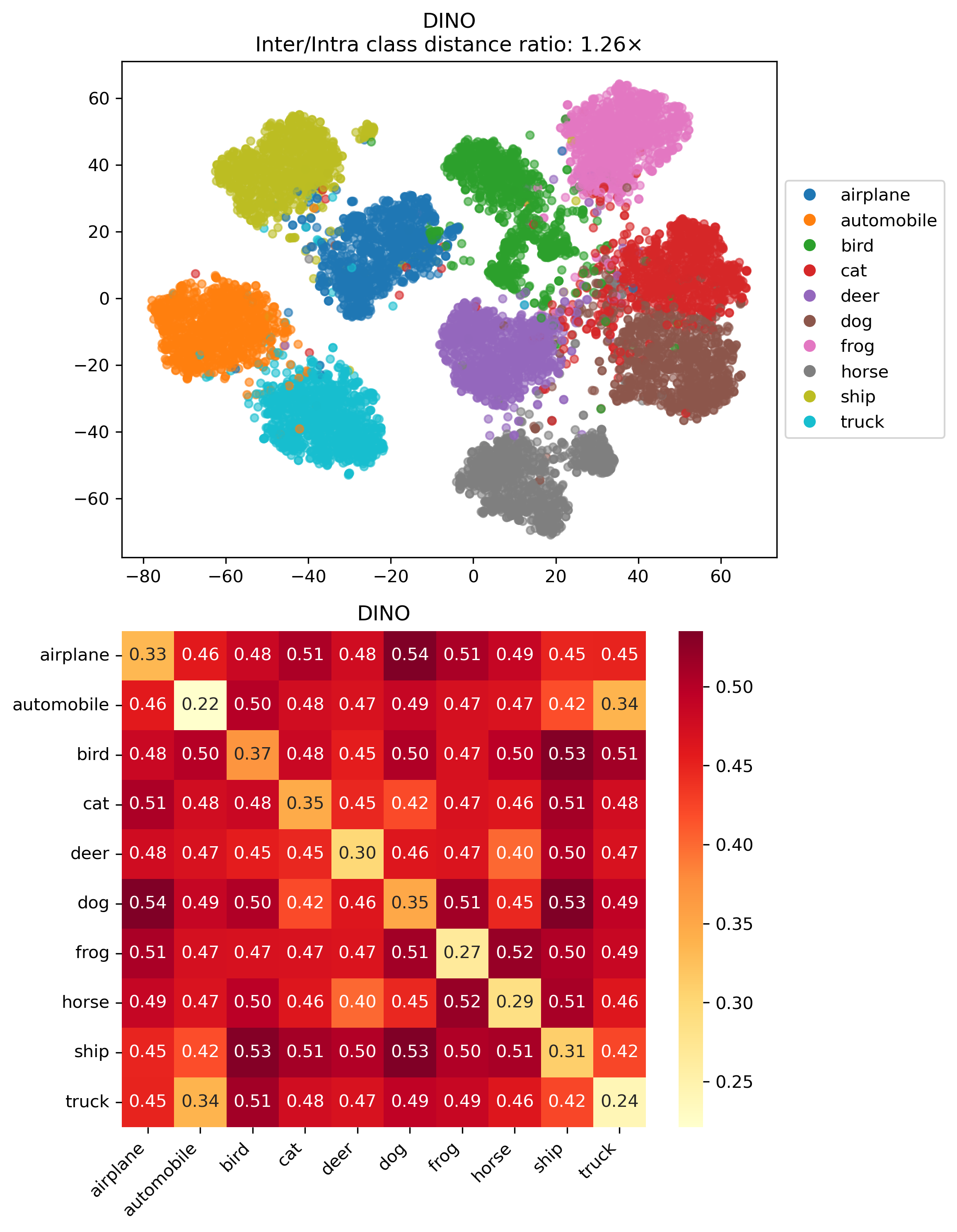}
        \caption{}
        \label{fig:subfig2}
    \end{subfigure}
    
    \vspace{0.5em}
    \begin{subfigure}[h]{0.35\textwidth}
        \includegraphics[width=\textwidth]{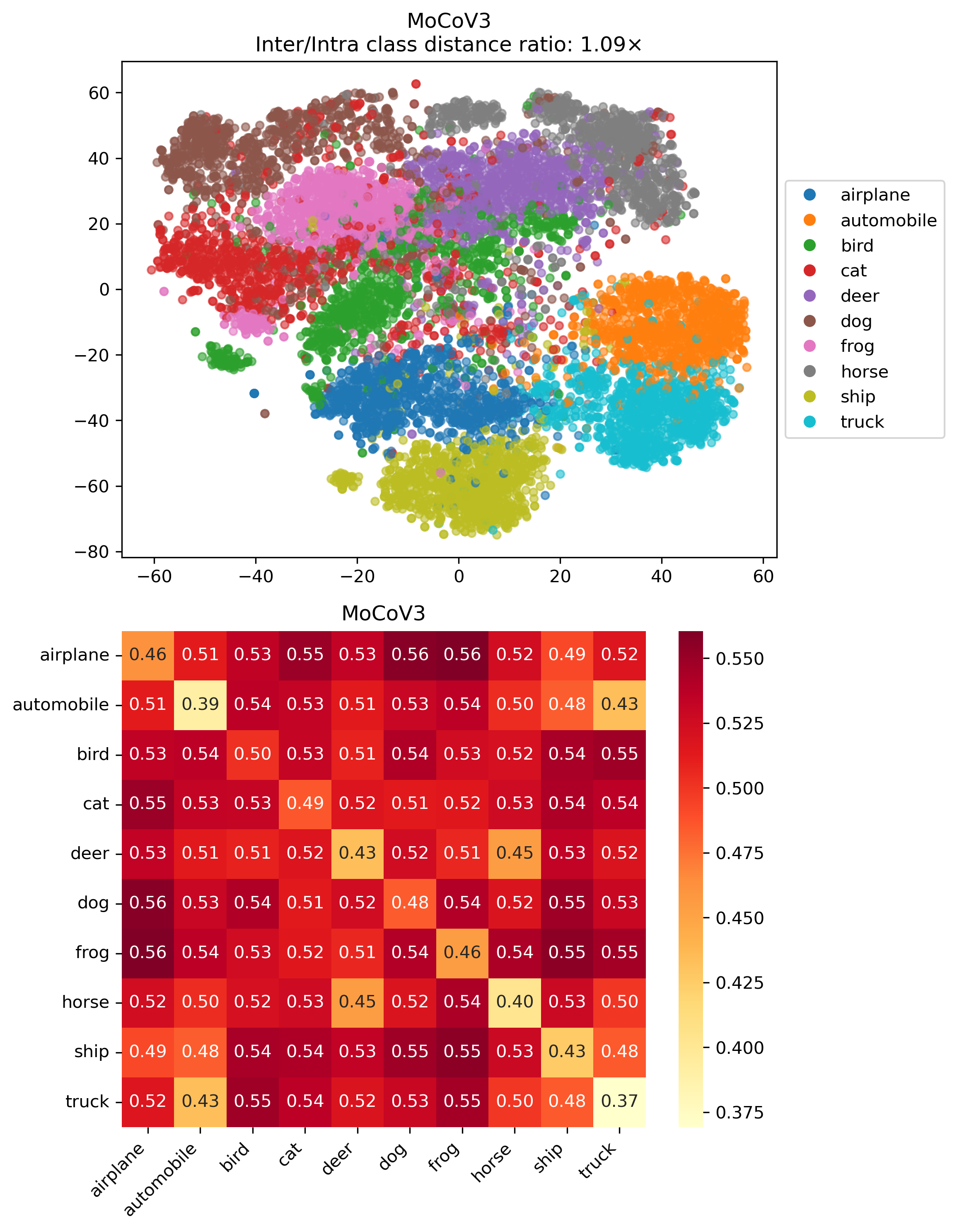}
        \caption{}
        \label{fig:subfig3}
    \end{subfigure}
    \hfill
    \begin{subfigure}[h]{0.35\textwidth}
        \includegraphics[width=\textwidth]{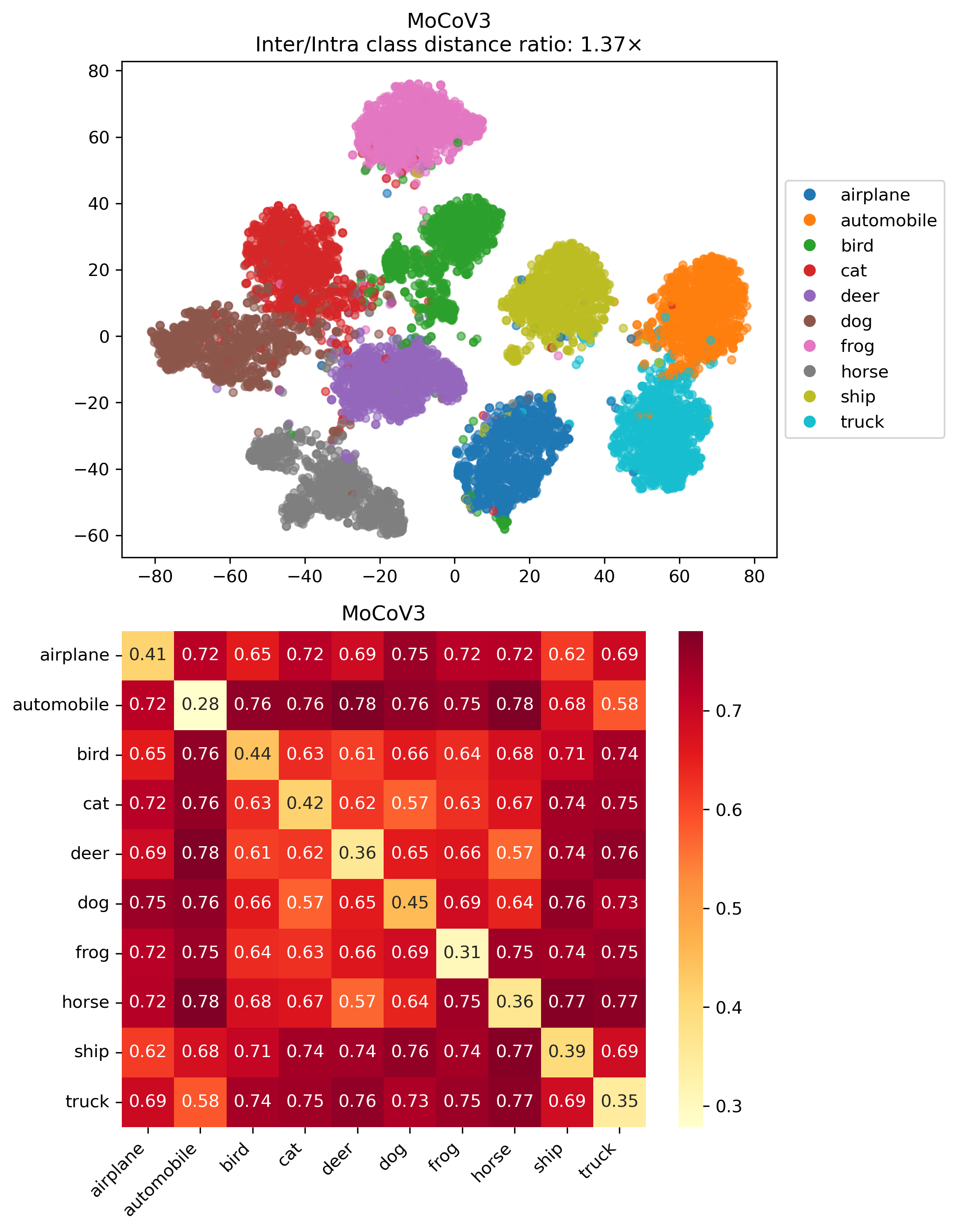}
        \caption{}
        \label{fig:subfig4}
    \end{subfigure}

    \caption{Comparison of feature representations for CIFAR-10 images using ResNet50 with different self-supervised learning (SSL) methods. \textbf{Top row:} Results from DINO. \textbf{Bottom row:} Results from MoCoV3. For each SSL method, \textbf{left} panels show results from \textbf{probed} models and \textbf{right} panels show results from \textbf{fine-tuned} models. Within each panel, the \textbf{upper} plots display t-SNE visualizations of the 2048-dimensional feature vectors using Euclidean distance, with points colored by class and Inter/Intra class distance ratios indicated. The \textbf{lower} plots show the corresponding class-wise distance matrices computed using cosine similarity, with the average distances between samples from each pair of classes. Higher Inter/Intra class distance ratios indicate better class separation in the feature space.}    \label{fig:main}
\end{figure*}

\begin{figure*}[h]
    \begin{subfigure}[h]{0.35\textwidth}
        \includegraphics[width=\textwidth]{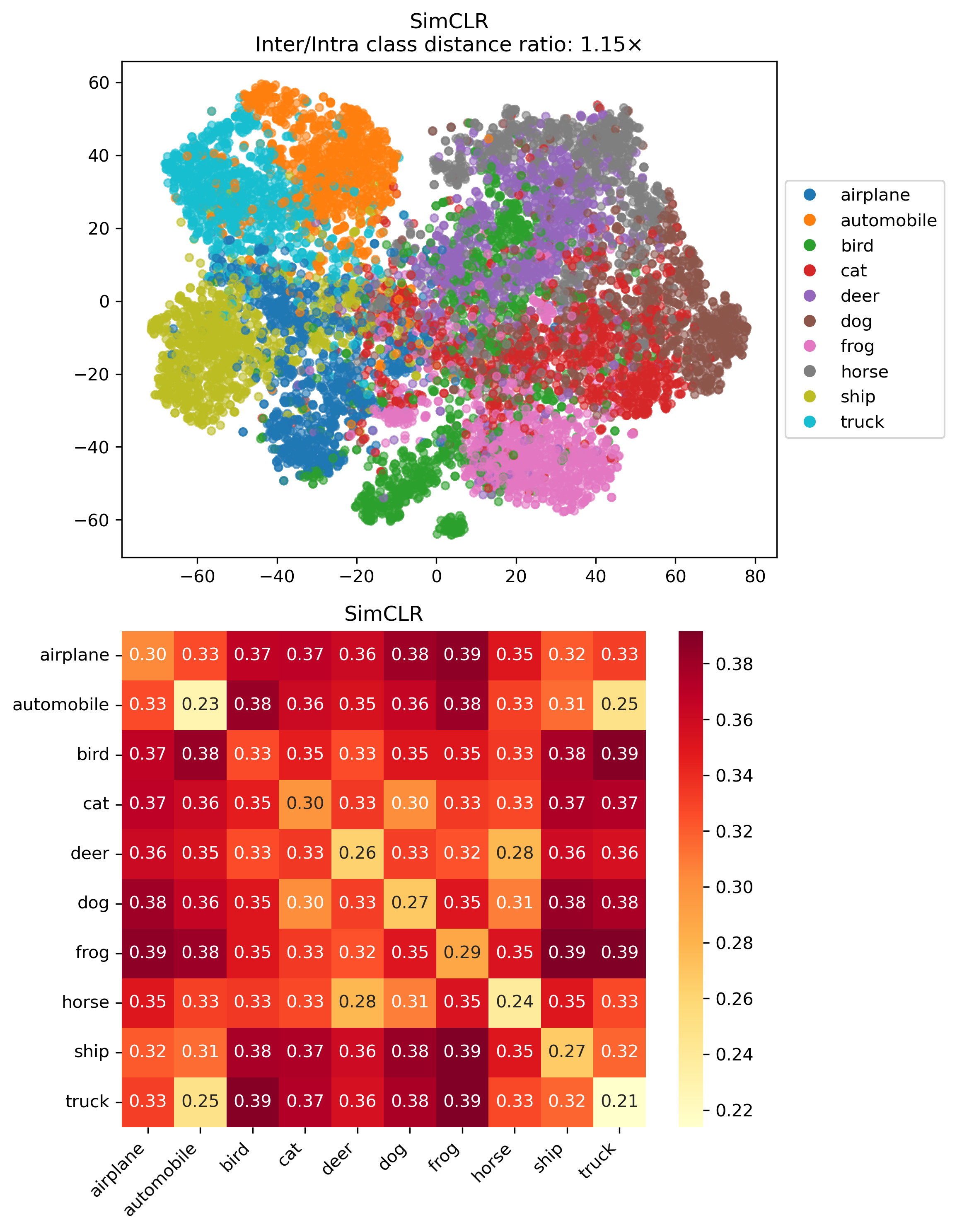}
        \caption{}
        \label{fig:subfig1}
    \end{subfigure}
    \hfill
    \begin{subfigure}[h]{0.35\textwidth}
        \includegraphics[width=\textwidth]{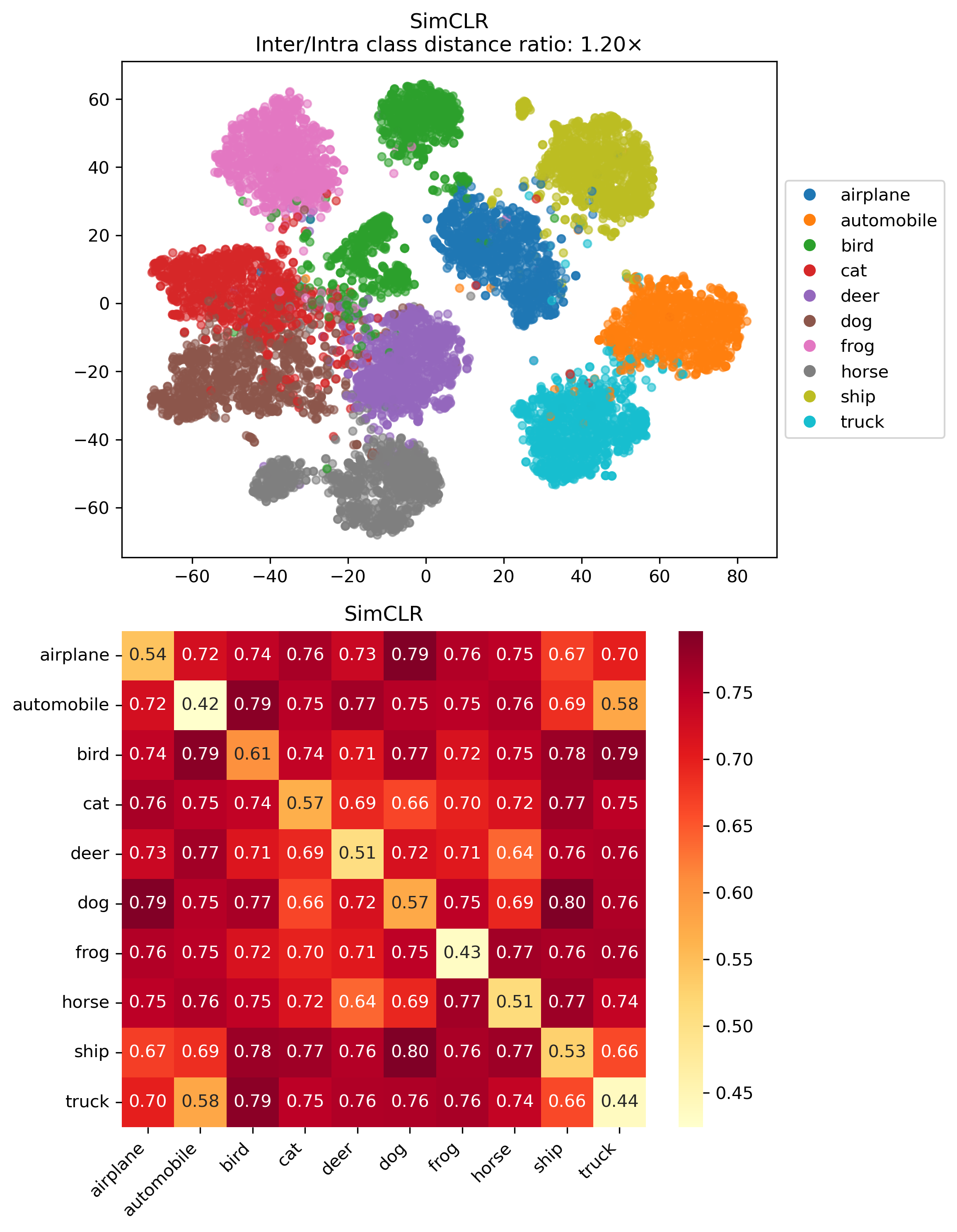}
        \caption{}
        \label{fig:subfig2}
    \end{subfigure}
    
    \vspace{0.5em}
    \begin{subfigure}[h]{0.35\textwidth}
        \includegraphics[width=\textwidth]{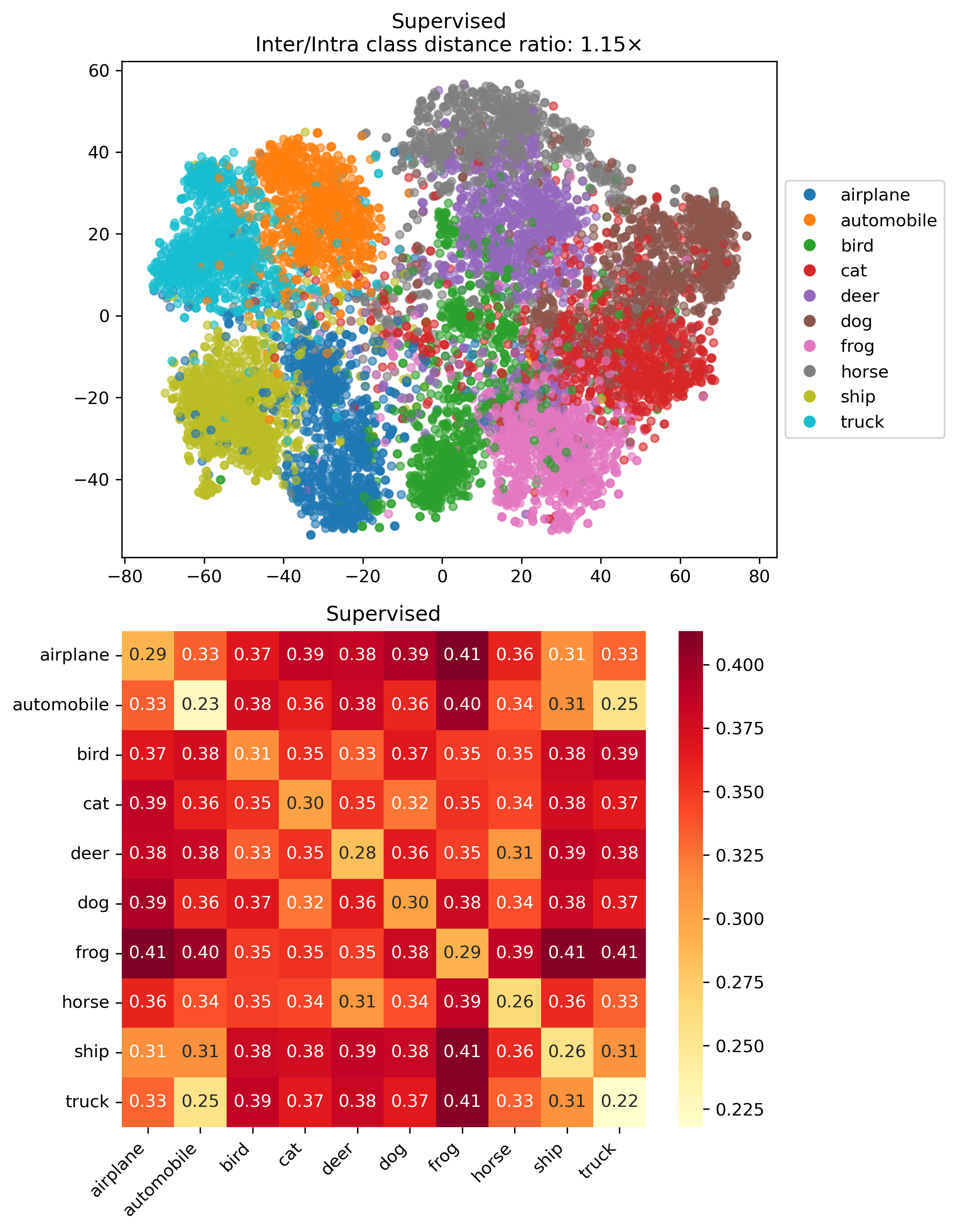}
        \caption{}
        \label{fig:subfig3}
    \end{subfigure}
    \hfill
    \begin{subfigure}[h]{0.35\textwidth}
        \includegraphics[width=\textwidth]{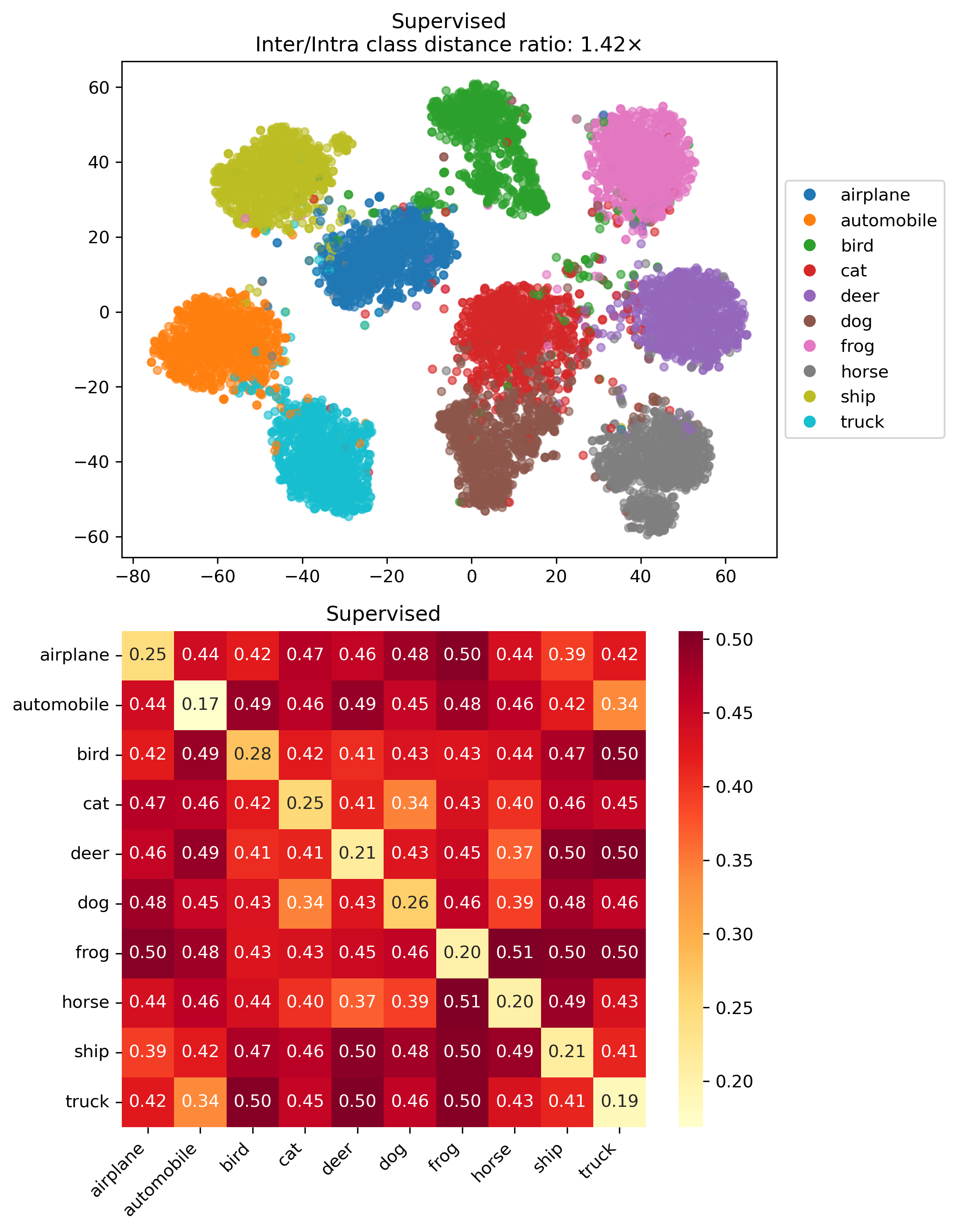}
        \caption{}
        \label{fig:subfig4}
    \end{subfigure}

    \caption{Comparison of feature representations for CIFAR-10 images using ResNet50 with different self-supervised learning (SSL) methods. \textbf{Top row:} Results from SimCLR. \textbf{Bottom row:} Results from Supervised. For each SSL method, \textbf{left} panels show results from \textbf{probed} models and \textbf{right} panels show results from \textbf{fine-tuned} models. Within each panel, the \textbf{upper} plots display t-SNE visualizations of the 2048-dimensional feature vectors using Euclidean distance, with points colored by class and Inter/Intra class distance ratios indicated. The \textbf{lower} plots show the corresponding class-wise distance matrices computed using cosine similarity, with the average distances between samples from each pair of classes. Higher Inter/Intra class distance ratios indicate better class separation in the feature space.}
    \label{fig:main}
\end{figure*}

\begin{figure*}[h]
    \begin{subfigure}[h]{0.35\textwidth}
        \includegraphics[width=\textwidth]{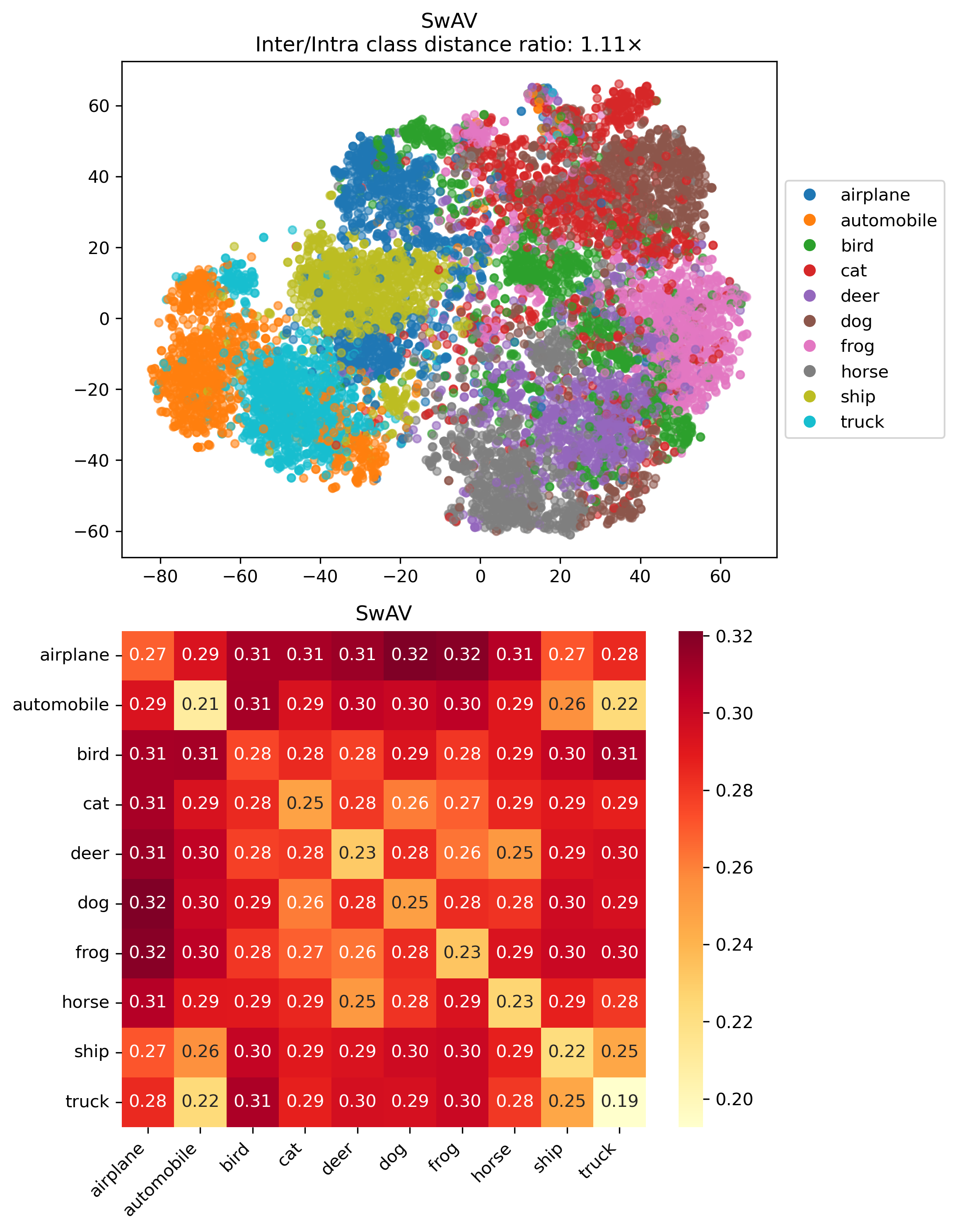}
        \caption{}
        \label{fig:subfig1}
    \end{subfigure}
    \hfill
    \begin{subfigure}[h]{0.35\textwidth}
        \includegraphics[width=\textwidth]{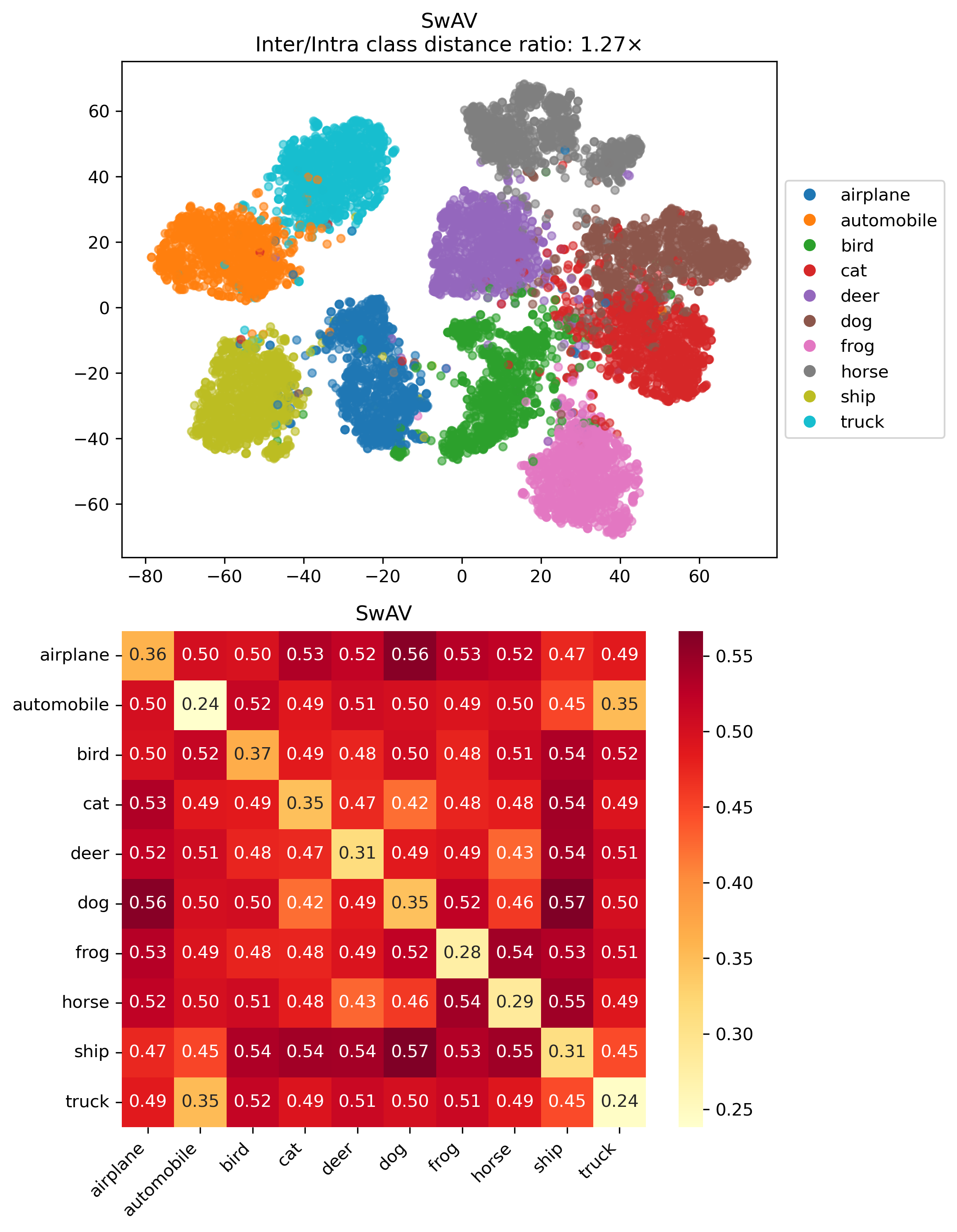}
        \caption{}
        \label{fig:subfig2}
    \end{subfigure}
    
    \vspace{0.5em}
    \begin{subfigure}[h]{0.35\textwidth}
        \includegraphics[width=\textwidth]{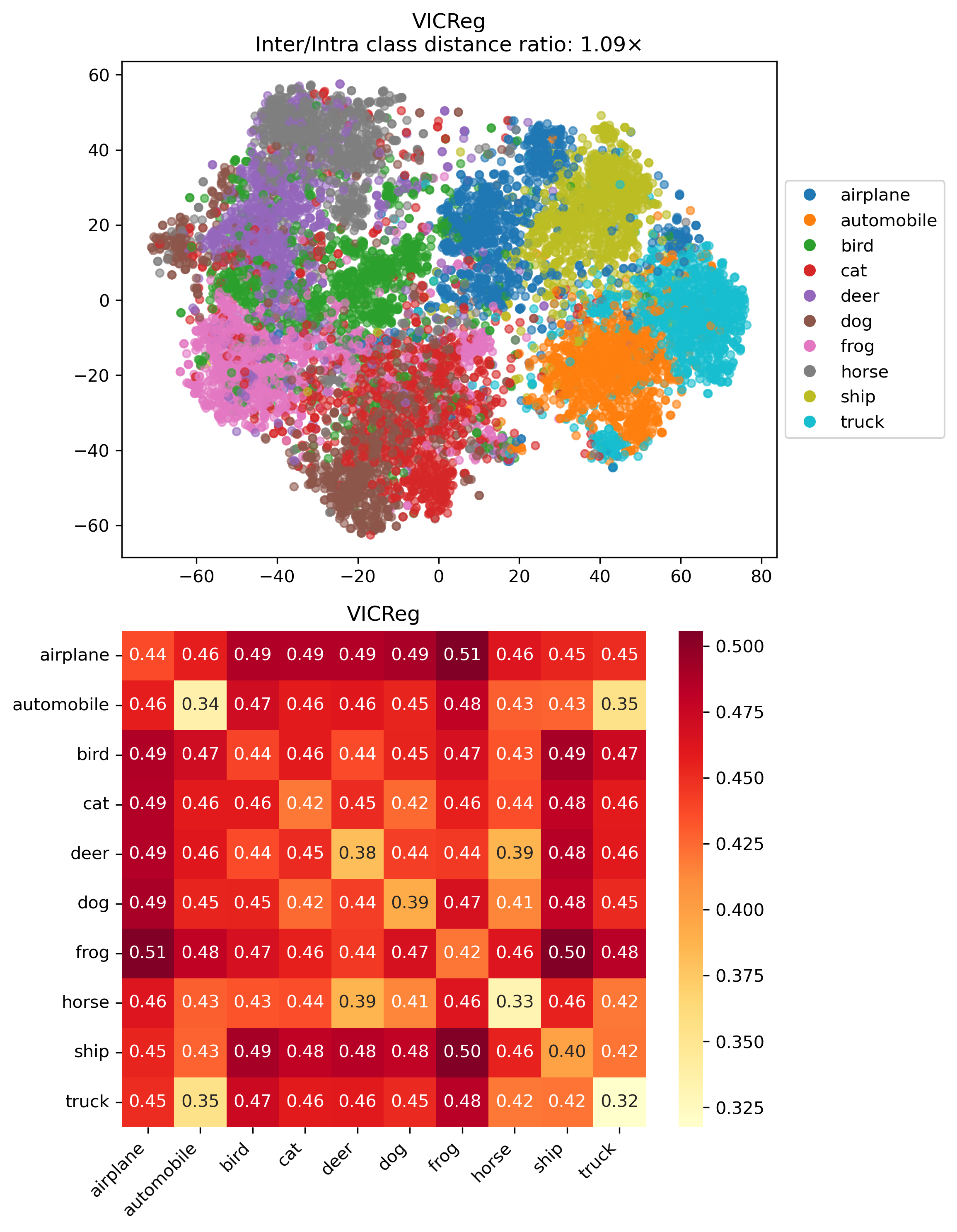}
        \caption{}
        \label{fig:subfig3}
    \end{subfigure}
    \hfill
    \begin{subfigure}[h]{0.35\textwidth}
        \includegraphics[width=\textwidth]{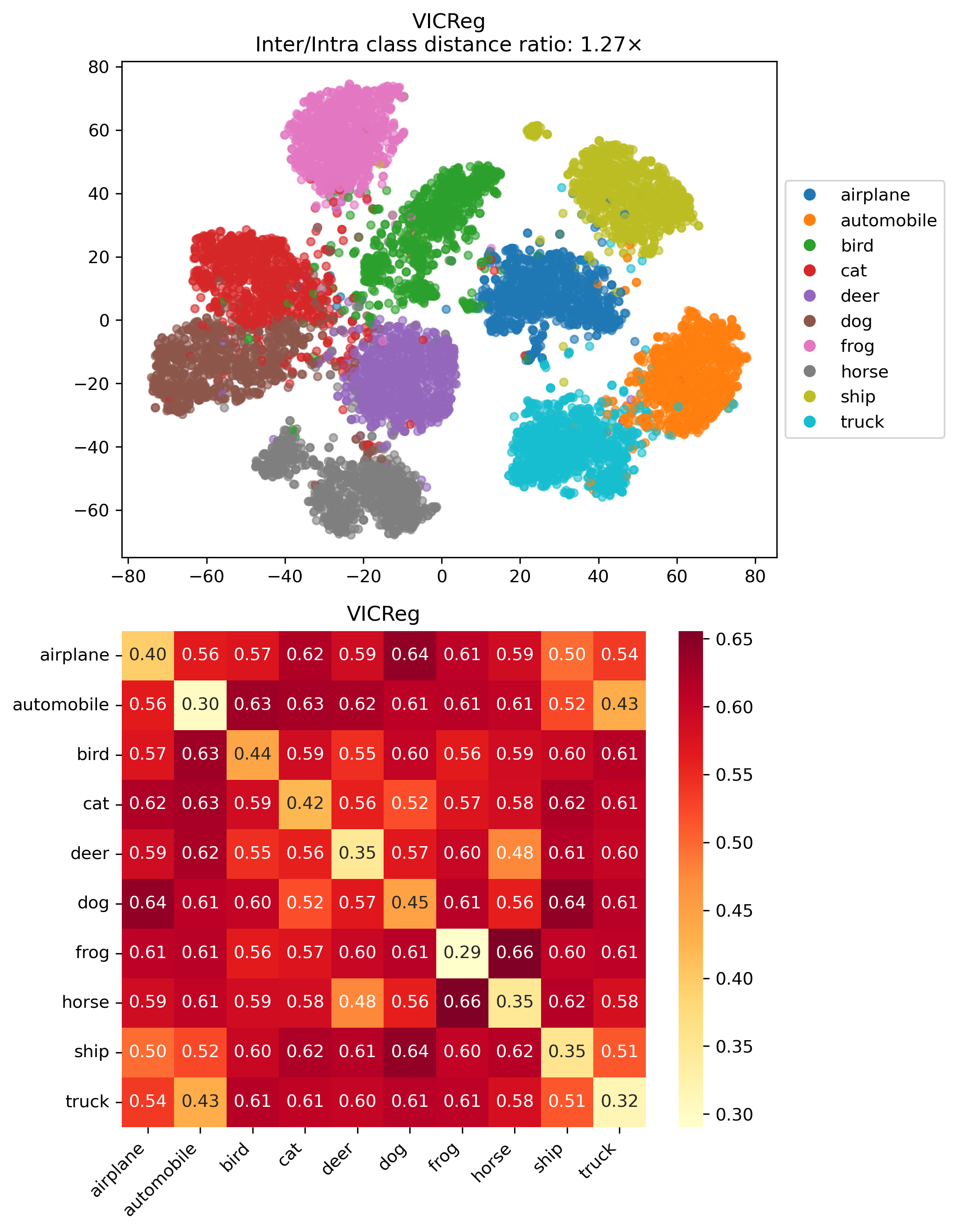}
        \caption{}
        \label{fig:subfig4}
    \end{subfigure}

    \caption{Comparison of feature representations for CIFAR-10 images using ResNet50 with different self-supervised learning (SSL) methods. \textbf{Top row:} Results from SwAV. \textbf{Bottom row:} Results from VICReg. For each SSL method, \textbf{left} panels show results from \textbf{probed} models and \textbf{right} panels show results from \textbf{fine-tuned} models. Within each panel, the \textbf{upper} plots display t-SNE visualizations of the 2048-dimensional feature vectors using Euclidean distance, with points colored by class and Inter/Intra class distance ratios indicated. The \textbf{lower} plots show the corresponding class-wise distance matrices computed using cosine similarity, with the average distances between samples from each pair of classes. Higher Inter/Intra class distance ratios indicate better class separation in the feature space.}
    \label{fig:main}
\end{figure*}
\clearpage
\section{Full Results}\label{app:full-res}
\subsection{ImageNet}\label{res:in1k}
\begin{table*}[h]
    \centering
    \scriptsize
    \caption{This table presents the results of various instance and universal adversarial perturbation (UAP) attacks on the Imagenet-1k dataset, with all UAP attack names in \textit{italics}. Different configurations of FGSM and PGD are denoted, such as $FGSM_1$ and $PGD_1$. Average results for universal adversarial perturbations (UAP Avg.), instance adversarial attacks (IAA Avg.), and overall adversarial performance (Adv Avg.) are reported at the bottom, including percentage drops relative to clean accuracy.}
    \label{tab:few_shot_small_domain_shift}

    \begin{adjustbox}{width=\textwidth}
    \begin{tabular}{lllllllll}
    \toprule
    {} & Barlow & BYOL & DINO & MoCoV3 & SimCLR & Supervised & SwAV & VICReg \\
    \midrule
    
    $FGSM_1$  & 42.41 & 39.41 & 24.68  & 42.67 & 24.29 & 38.83 & 24.71  & 42.42\\
    $FGSM_2$  & 18.11 & 13.47 & 5.66 & 15.53 & 8.84 & 12.18 & 6.35 & 18.11\\
    
    $PGD_1$  & 42.38 & 39.63 & 25.65 & 42.39 & 26.6 & 35.26 & 26.48 & 42.41\\
    $PGD_2$  & 1.48 & 0.65 & 0.18 & 1.06 & 0.25 & 0.37 & 0.18 & 1.5\\
    
    $PGD_3$  & 42.6 & 39.82 & 25.85 & 42.56 & 26.79 & 35.39 & 26.73 & 42.6\\
    $PGD_4$  & 1.19 & 0.5 & 0.14 & 0.82 & 0.2 & 0.28 & 0.14 & 1.2\\
    
    $PGD_5$ & 5.18 & 3.44 & 0.67 & 4.79 & 0.9 & 1.9 & 0.69 & 5.15\\
    
    DIFGSM  & 52 & 52.71 & 41.12 & 54.09 & 42.57 & 51.43 & 45.65 & 52.49\\
    
    CW & 0.18 & 0.02 & 0 & 0.02 & 0.02 & 0.02 & 0 & 0.19\\
    
    Jitter & 59.83 & 61.92 & 60.26 & 62.47 & 56.4 & 62.75 & 61.16 & 59.84\\
    
    TIFGSM & 61.04 & 62.27 & 56.98 & 61.47 & 55.63 & 62.16 & 60.07 & 59.91\\
    
    PIFGSM & 34.38 & 29.83 & 14.54 & 34.1 & 13.34 & 28.64 & 14.12 & 34.43\\
    
    EADEN & 0 & 0 & 0 & 0 & 0 & 0 & 0 & 0\\
    OnePixel & 69.34 & 72.5 & 72.83 & 72.64 & 66.47 & 73.27 & 72.73 & 69.38\\
    Pixle & 25.22 & 28.67 & 19.41 & 31.45 & 21.75 & 23.21 & 16.95 & 25.23\\
    SPSA & 66.59 & 69.59 & 68.11 & 69.93 & 63.01 & 69.48 & 68.61 & 66.63\\
    Square & 4.44 & 2.62 & 1.3 & 3.15 & 4.22 & 0.87 & 1.99 & 4.49\\
    TAP & 70.31 & 74.36 & 73.78 & 73.72 & 68.1 & 68.98 & 75.05 & 70.33\\
    
    \textit{ASV} & 62.17 &	68.46 &	68.69 &	67.59 &	61.84 &	64.28 &	67.66 &	62.51\\

    \textit{FFF (mean-std)} & 53.72	&62.38&	59.45&	46.78&	56.79	&53.45&	61.14&	54.76\\

    \textit{FFF (no-data)} & 39.64	&31.94&	40.73&	57.39&	53.52&	42.49&	32.15&	38.68\\
    
    \textit{FFF (one-sample)} & 30.53 &	54.41 &	16.15 &	57.18 &	28.63 &	40.57 &	29.96  &	30.14\\
    
    \textit{FG-UAP} & 4.34 &	1.83	&2.3	&3.08	&1.03	&1.89	&2.17&	1.95\\

    \textit{GD-UAP (mean-std)} & 56.92	&61.1	&49.85	&53.59	&55.51	&59.6	&57.38	&58.07\\
    
    \textit{GD-UAP (no-data)} &33.41	&37.96	&26.59	&43.66	&37.05	&36.01	&36.07	&31.29\\

    \textit{GD-UAP (one-sample)} & 42.17&	37.54&	25.9&	35.1&	14.11&	25.27&	36.37&	40.08\\

    \textit{L4A-base} & 42.17&	36.35&	51.13&	41.09&	7.83&	26.73&	10.9&	28.36\\
    
    \textit{L4A-fuse} & 28.33&	36.01&	50.69&	40.39&	8.34&	26.01&	10.94&	28.55\\
    
    \textit{L4A-ugs} & 27.56&	60.03&	56.83&	65.27&	53.08&	34.27&	49.91&	58.23\\
    
    \textit{PD-UAP} & 65.06&	54.06&	45.06&	56.248&	60.96&	46.79&	51.75&	65.07\\
    
    \textit{SSP} & 27.7	&41.05	&38.55	&48.56	&35.06	&9.43	&23.27	&33.26\\
    
    \textit{STD} & 49.19&	58.08&	52.44&	53.44&	50.3&	56.16&	56.08&	48.88\\
    
    \textit{UAP (DeepFool)} & 13.23&	20.25&	15.18&	24.25&	23.72&	13.67&	8.59&	13.22 \\
    
    \textit{UAPEPGD} & 67.74 &	67.21&	70.6&	71.57&	64.36&	71.26&	70.98&	67.88\\
    \hline
    Clean Accuracy & 71.20 & 74.57 & 75.28 & 74.57 & 68.90 & 76.13 & 75.27 & 71.26 \\
    IAA Avg. & 33.14 \textcolor{red}{\tiny{$\downarrow$54\%}}  & 32.86\textcolor{red}{\tiny{$\downarrow$56\%}} & 27.28\textcolor{red}{\tiny{$\downarrow$64\%}}  & 34.04\textcolor{red}{\tiny{$\downarrow$54\%}} & 26.63\textcolor{red}{\tiny{$\downarrow$61\%}} & 31.39\textcolor{red}{\tiny{$\downarrow$59\%}} & 27.87\textcolor{red}{\tiny{$\downarrow$63\%}}  & 33.12\textcolor{red}{\tiny{$\downarrow$54\%}} \\
    UAP Avg. & 40.24 \textcolor{red}{\tiny{$\downarrow$43\%}}  & 45.79\textcolor{red}{\tiny{$\downarrow$39\%}} & 41.88\textcolor{red}{\tiny{$\downarrow$44\%}}  & 47.82\textcolor{red}{\tiny{$\downarrow$36\%}} & 38.25\textcolor{red}{\tiny{$\downarrow$44\%}} & 37.99\textcolor{red}{\tiny{$\downarrow$50\%}} & 37.83\textcolor{red}{\tiny{$\downarrow$50\%}}  & 41.30\textcolor{red}{\tiny{$\downarrow$42\%}} \\

    Adv Avg. & 36.48 \textcolor{red}{\tiny{$\downarrow$49\%}}  & 38.94\textcolor{red}{\tiny{$\downarrow$48\%}} & 34.16\textcolor{red}{\tiny{$\downarrow$55\%}}  & 40.53\textcolor{red}{\tiny{$\downarrow$46\%}} & 32.10\textcolor{red}{\tiny{$\downarrow$53\%}} & 34.50\textcolor{red}{\tiny{$\downarrow$55\%}} & 32.56\textcolor{red}{\tiny{$\downarrow$57\%}}  & 36.98\textcolor{red}{\tiny{$\downarrow$48\%}} \\

    \midrule
    \end{tabular}
    \end{adjustbox}
\end{table*}

\clearpage
\subsection{Segmentation}\label{res:seg}
\subsubsection{Pascal VOC 2012}

\begin{table*}[h]
    \centering
    \begin{adjustbox}{width=0.9\textwidth}
    \begin{tabular}{lcccccccc}
    \toprule
    \textbf{Metric} & \textbf{Barlow} & \textbf{BYOL} & \textbf{DINO} & \textbf{MocoV3} & \textbf{SimCLR} & \textbf{Supervised} & \textbf{SwAV} & \textbf{VICReg} \\
    \midrule
    \multicolumn{9}{c}{\textbf{Alma}} \\
    mIOU ($\uparrow$)  &  0.35 & 0.33 & 0.34 & 0.4 & 0.31 & 0.26 & 0.38 & 0.39 \\
    APSR ($\downarrow$)  & 99.02 & 99.01 & 99.02 & 98.91 & 99 & 99.01 & 99.01 & 98.99 \\
    \midrule
    \multicolumn{9}{c}{\textbf{Asma}} \\
    mIOU ($\uparrow$)  &49.4 & 63.39 & 61.36 & 61.57 & 32.06 & 77.3 & 62.12 & 50.38\\
    APSR ($\downarrow$)  &15.39 & 10.95 & 11.38 & 12.18 & 22.78 & 5.29 & 11.56 & 14.48\\
    \midrule
    \multicolumn{9}{c}{\textbf{DAG}} \\
    mIOU ($\uparrow$)  & 0.02 & 0.02 & 0.02 & 0.02 & 0.03 & 0.05 & 0.02 & 0.02 \\
    APSR ($\downarrow$)  & 99.87 & 99.91 & 99.89 & 99.88 & 99.83 & 99.74 & 99.89 & 99.89 \\
    \midrule
    \multicolumn{9}{c}{\textbf{DDN}} \\
    mIOU ($\uparrow$)  & 5.62 & 4.64 & 5.11 & 7.16 & 1.67 & 1.52 & 6.91 & 4.94 \\
    APSR ($\downarrow$)  & 89.66 & 92.6 & 92.75 & 88.01 & 97.24 & 88.56 & 90.77 & 87.23 \\
    \midrule
    \multicolumn{9}{c}{\textbf{FGSM}} \\
    mIOU ($\uparrow$)  & 30.35 & 29.28 & 30.41 & 29.43 & 32.15 & 38.31 & 29.4 & 29.84 \\
    APSR ($\downarrow$)  & 35.91 & 45.62 & 39.66 & 41.71 & 33.55 & 21.36 & 42.94 & 39.31 \\
    \midrule
    \multicolumn{9}{c}{\textbf{FMN}} \\
    mIOU ($\uparrow$)  & 5.4 & 5.29 & 4.86 & 5.19 & 5.07 & 2.74 & 4.9 & 6.2 \\
    APSR ($\downarrow$)  &91.18 & 92.25 & 91.02 & 91.42 & 89.88 & 93.53 & 91.94 & 89.99 \\
    \midrule
    \multicolumn{9}{c}{\textbf{PGD}} \\
    mIOU ($\uparrow$)  & 12.67 & 13.16 & 12.75 & 13.06 & 12.88 & 10.92 & 12.98 & 13.04 \\
    APSR ($\downarrow$)  & 70.07 & 82 & 77 & 79.27 & 71.15 & 67.4 & 77.31 & 72.43 \\
    \midrule
    \textbf{Clean mIOU ($\uparrow$)}  & 72.63 & 70.37 & 71.65 & 71.25 & 71.96 & 77.35 & 70.8 & 70.33 \\
    \textbf{Clean APSR ($\downarrow$)}  & 7.18 & 8.29 & 7.64 & 7.83 & 7.2 & 5.27 & 8.21 & 8.01 \\
    \textbf{Adversarial mIOU ($\uparrow$)}  & 14.83\textcolor{red}{\scriptsize{$\downarrow$80\%}}  & 16.59\textcolor{red}{\scriptsize{$\downarrow$78\%}}  & 16.41\textcolor{red}{\scriptsize{$\downarrow$77\%}}  & 16.69\textcolor{red}{\scriptsize{$\downarrow$77\%}}  & 12.02\textcolor{red}{\scriptsize{$\downarrow$83\%}}  & 18.73\textcolor{red}{\scriptsize{$\downarrow$76\%}}  & 16.67\textcolor{red}{\scriptsize{$\downarrow$77\%}}  & 14.97\textcolor{red}{\scriptsize{$\downarrow$79\%}}  \\
    \textbf{Adversarial APSR ($\downarrow$)}  & 71.59\textcolor{red}{\scriptsize{$\uparrow$64\%}}  & 74.62 \textcolor{red}{\scriptsize{$\uparrow$66\%}}& 72.96 \textcolor{red}{\scriptsize{$\uparrow$65\%}}& 73.05\textcolor{red}{\scriptsize{$\uparrow$65\%}} & 73.35\textcolor{red}{\scriptsize{$\uparrow$66\%}} & 67.84\textcolor{red}{\scriptsize{$\uparrow$64\%}} & 73.35\textcolor{red}{\scriptsize{$\uparrow$65\%}} & 71.76\textcolor{red}{\scriptsize{$\uparrow$64\%}} \\
    \bottomrule
    \end{tabular}
    \end{adjustbox}

    \caption{Performance metrics (mIOU and APSR) for various self-supervised and supervised models under different adversarial attacks, using unfrozen backbones. Clean and adversarial scores are reported, with percentage changes in adversarial performance noted. Higher mIOU and lower APSR indicate better results}
    \label{tab:my_label}
\end{table*}

\begin{table*}[h]
    \centering
    \begin{adjustbox}{width=0.9\textwidth}
    \begin{tabular}{lcccccccc}
    
    \toprule
    \textbf{Metric} & \textbf{Barlow} & \textbf{BYOL} & \textbf{DINO} & \textbf{MocoV3} & \textbf{SimCLR} & \textbf{Supervised} & \textbf{SwAV} & \textbf{VICReg} \\
    \midrule
    \multicolumn{9}{c}{\textbf{Alma}} \\
    mIOU ($\uparrow$)  & 0.39   & 0.31   & 0.37   & 0.37   & 0.55  & 0.28   & 0.35   & 0.41 \\
    APSR ($\downarrow$)  & 99.02  & 99.02  & 99.02  & 99.02  & 98.45  & 99.01  & 99.02  & 99.02 \\
    \midrule
    \multicolumn{9}{c}{\textbf{Asma}} \\
    mIOU ($\uparrow$)  & 76.06  & 72.84  & 75.32  & 72.84  & 70.42  & 69.84  & 74.09  & 76.74 \\
    APSR ($\downarrow$)  & 6.01   & 7.23   & 6.14   & 7.58   & 5.98  & 8.18   & 6.75   & 5.98 \\
    \midrule
    \multicolumn{9}{c}{\textbf{DAG}} \\
    mIOU ($\uparrow$)  & 0.03   & 0.04   & 0.02   & 0.04   & 0.04   & 0.02   & 0.03   & 0.03 \\
    APSR ($\downarrow$)  & 99.90  & 99.87  & 99.89  & 99.87  & 99.82  & 99.87  & 99.88  & 99.89 \\
    \midrule
    \multicolumn{9}{c}{\textbf{DDN}} \\
    mIOU ($\uparrow$)  & 10.81  & 9.76   & 6.91   & 10.74  & 6.62   & 2.95   & 8.57   & 11.12 \\
    APSR ($\downarrow$)  & 79.62  & 75.93  & 82.58  & 78.71  & 75.20  & 87.30  & 83.48  & 80.41 \\
    \midrule
    \multicolumn{9}{c}{\textbf{FGSM}} \\
    mIOU ($\uparrow$)  & 35.16  & 31.90  & 30.88  & 35.18  & 36.25  & 27.70  & 32.37  & 34.99 \\
    APSR ($\downarrow$)  & 33.29  & 33.63  & 36.12  & 33.63  & 27.35  & 36.99  & 36.10  & 33.69 \\
    \midrule
    \multicolumn{9}{c}{\textbf{FMN}} \\
    mIOU ($\uparrow$)  & 6.63   & 6.23   & 6.22   & 6.42   & 8.92  & 4.23   & 6.48   & 6.56 \\
    APSR ($\downarrow$)  & 87.73  & 87.10  & 87.12  & 87.70  & 81.28  & 91.30  & 87.23  & 87.23 \\
    \midrule
    \multicolumn{9}{c}{\textbf{PGD}} \\
    mIOU ($\uparrow$)  & 14.13  & 12.12  & 12.12  & 13.25  & 12.23  & 10.49  & 12.31  & 13.51 \\
    APSR ($\downarrow$)  & 76.16  & 75.49  & 75.49  & 76.60  & 73.38  & 78.37  & 80.82  & 77.62 \\
    \midrule
    \textbf{Clean mIOU ($\uparrow$)}  & 76.90  & 76.69  & 77.01  & 76.19  & 75.62  & 74.20  & 76.54  & 77.89 \\
    \textbf{Clean APSR ($\downarrow$)}  & 5.75   & 5.74   & 5.38  & 6.01   & 5.98   & 6.35   & 5.79   & 5.48 \\
    \textbf{Adversarial mIOU ($\uparrow$)}  & 20.46\textcolor{red}{\scriptsize{$\downarrow$73\%}}   & 19.03\textcolor{red}{\scriptsize{$\downarrow$75\%}}   & 18.83\textcolor{red}{\scriptsize{$\downarrow$76\%}}   & 19.83\textcolor{red}{\scriptsize{$\downarrow$74\%}}   & 19.29\textcolor{red}{\scriptsize{$\downarrow$74\%}}   & 16.50 \textcolor{red}{\scriptsize{$\downarrow$78\%}}  & 19.17 \textcolor{red}{\scriptsize{$\downarrow$75\%}}  & 20.48\textcolor{red}{\scriptsize{$\downarrow$74\%}}  \\
    \textbf{Adversarial APSR ($\downarrow$)}  & 68.82\textcolor{red}{\scriptsize{$\uparrow$63\%}}  & 68.32\textcolor{red}{\scriptsize{$\uparrow$63\%}} & 69.48 \textcolor{red}{\scriptsize{$\uparrow$64\%}} & 69.02\textcolor{red}{\scriptsize{$\uparrow$63\%}}  & 65.92 \textcolor{red}{\scriptsize{$\uparrow$60\%}} & 71.57 \textcolor{red}{\scriptsize{$\uparrow$65\%}} & 70.47\textcolor{red}{\scriptsize{$\uparrow$65\%}}  & 69.12\textcolor{red}{\scriptsize{$\uparrow$64\%}} \\
    \bottomrule
    \end{tabular}
    \end{adjustbox}

    \caption{Performance metrics (mIOU and APSR) for various self-supervised and supervised models under different adversarial attacks, using frozen backbones. Clean and adversarial scores are reported, with percentage changes in adversarial performance noted. Higher mIOU and lower APSR indicate better results.}
    \label{tab:my_label}
\end{table*}

\clearpage

\subsubsection{CityScapes}

\begin{table*}[h]
    \centering
    \begin{adjustbox}{width=0.9\textwidth}
    \begin{tabular}{lcccccccc}
    \toprule
    \textbf{Metric} & \textbf{Barlow} & \textbf{BYOL} & \textbf{DINO} & \textbf{MocoV3} & \textbf{SimCLR} & \textbf{Supervised} & \textbf{SwAV} & \textbf{VICReg} \\
    \midrule
    \textbf{IOU (Higher Better)} & 65.48 & 62.05 & 65.87 & 63.88 & 58.82 & 62.57 & 66.48 & 65.64 \\
    \textbf{APSR (Lower Better)} & 6.4 & 7.05 & 6.08 & 6.68 & 7.35 & 7.38 & 6.27 & 6.29 \\
    \midrule
    \multicolumn{9}{c}{\textbf{Alma}} \\
    IOU & 3.82 & 3.45 & 3.72 & 2.99 & 8.43 & 4.58 & 5.51 & 4.08 \\
    APSR & 91.4 & 91.89 & 91.09 & 93.99 & 77.65 & 89.24 & 86.44 & 90.41 \\
    \midrule
    \multicolumn{9}{c}{\textbf{Asma}} \\
    IOU & 50.9 & 18.62 & 64.88 & 51.93 & 33.41 & 58.43 & 63.07 & 65.38 \\
    APSR & 17.3 & 44.51 & 7.49 & 13.76 & 21.71 & 8.87 & 8.75 & 6.73 \\
    \midrule
    \multicolumn{9}{c}{\textbf{DAG}} \\
    IOU & 0.19 & 0.27 & 0.21 & 0.33 & 0.27 & 0.21 & 0.15 & 0.24 \\
    APSR & 99.8 & 99.74 & 99.75 & 99.72 & 99.57 & 99.6 & 99.76 & 99.79 \\
    \midrule
    \multicolumn{9}{c}{\textbf{DDN}} \\
    IOU & 1.49 & 1.99 & 1.66 & 1.18 & 1.12 & 1.59 & 0.51 & 1.36 \\
    APSR & 84.79 & 82.15 & 84.35 & 82.85 & 94.06 & 81 & 97.32 & 83.05 \\
    \midrule
    \multicolumn{9}{c}{\textbf{FGSM}} \\
    IOU & 30.31 & 26.22 & 28.75 & 30.92 & 30.64 & 28.19 & 28.68 & 20.33 \\
    APSR & 31.97 & 39.57 & 30.72 & 26.8 & 24.97 & 31.38 & 33.62 & 31.99 \\
    \midrule
    \multicolumn{9}{c}{\textbf{FMN}} \\
    IOU & 8.03 & 9.19 & 6.04 & 8.2 & 16.77 & 7.81 & 8.48 & 8.2 \\
    APSR & 77.75 & 72.11 & 81.38 & 77.35 & 58.44 & 74.76 & 77.34 & 75.44 \\
    \midrule
    \multicolumn{9}{c}{\textbf{PGD}} \\
    IOU & 10 & 10.71 & 7.76 & 12.23 & 9.94 & 11.59 & 11.43 & 9.62 \\
    APSR & 71.92 & 66.84 & 75.79 & 72.8 & 70.34 & 68.75 & 71.46 & 70.16 \\
    \midrule
    \textbf{Average IOU} & 14.96\textcolor{red}{\scriptsize{$\downarrow$77\%}} & 10.06\textcolor{red}{\scriptsize{$\downarrow$84\%}} & 16.15\textcolor{red}{\scriptsize{$\downarrow$75\%}} & 15.40\textcolor{red}{\scriptsize{$\downarrow$76\%}} & 14.37\textcolor{red}{\scriptsize{$\downarrow$76\%}} & 16.06\textcolor{red}{\scriptsize{$\downarrow$74\%}} & 16.83\textcolor{red}{\scriptsize{$\downarrow$75\%}} & 15.60{\scriptsize{$\downarrow$76\%}} \\
    \textbf{Average APSR} & 67.85\textcolor{red}{\scriptsize{$\uparrow$61\%}} & 70.97\textcolor{red}{\scriptsize{$\uparrow$64\%}} & 67.22\textcolor{red}{\scriptsize{$\uparrow$61\%}} & 66.75\textcolor{red}{\scriptsize{$\uparrow$56\%}} & 63.82\textcolor{red}{\scriptsize{$\uparrow$57\%}} & 64.80\textcolor{red}{\scriptsize{$\uparrow$62\%}} & 67.81\textcolor{red}{\scriptsize{$\uparrow$61\%}} & 65.37\textcolor{red}{\scriptsize{$\uparrow$60\%}} \\
     
    \bottomrule
    \end{tabular}
    \end{adjustbox}
    \caption{Updated performance metrics (mIOU and APSR) for various self-supervised and supervised models under different adversarial attacks, using frozen backbones. Clean and adversarial scores are reported, with percentage changes in adversarial performance noted. Higher mIOU and lower APSR indicate better results.}
    \label{tab:my_label}
\end{table*}

\clearpage

\clearpage
\subsection{Detection}\label{res:det}

\subsubsection{INRIA Person}
\begin{table*}[h]
    \centering
    \small
    \caption{Adversarial Attack Results on Detection using unfrozen SSL and Supervised Models as backbones. The table presents performance metrics under clean and adversarial conditions for various attack types (Optim, BIM, MIM, SGD, PGD, Optim-Adam, Optim-Nesterov). The last two rows display clean mean Average Precision (mAP) and the average performance under adversarial attacks, with the percentage decrease in performance highlighted in red}
    \label{tab:table8}

    \begin{adjustbox}{width=\textwidth}
    \begin{tabular}{lllllllll}
    \toprule
    {} & Barlow & BYOL & DINO & MocoV3 & SimCLR & Supervised & SwAV & VICReg \\    
    \midrule
    Optim & 6.18 & 1.68 & 1.77 & 4.87 & 1.06 & 1.54 & 4.27 & 2.12 \\
    BIM & 32.78 & 26.93 & 31.82 & 21.63 & 12.6 & 1.75 & 40.84 & 23.22 \\
    MIM & 11.89 & 26.24 & 5.2 & 10.7 & 3.1 & 1.94 & 10.69 & 7.85 \\
    SGD & 6.13 & 2.89 & 7.59 & 20.15 & 2.45 & 2.4 & 13.71 & 2.99 \\
    PGD & 84.58 & 78.44 & 80.97 & 81.96 & 80.45 & 57.76 & 80.54 & 77.52 \\
    Optim-Adam & 6.43 & 1.49 & 2.07 & 7.49 & 1.31 & 1.32 & 4.47 & 1.99 \\
    Optim-Nesterov & 2.34 & 1.58 & 1.31 & 5.24 & 1.43 & 2.55 & 4.34 & 1.42 \\
    \hline
    Clean mAP & 89.14 & 88.98 & 89.74 & 89.74 & 88.16 & 86.45 & 88.60 & 89.45 \\
    Adv mAP. & 21.48\textcolor{red}{\scriptsize{$\downarrow$76\%}} & 19.89\textcolor{red}{\scriptsize{$\downarrow$78\%}} & 18.68\textcolor{red}{\scriptsize{$\downarrow$79\%}} & 21.72\textcolor{red}{\scriptsize{$\downarrow$76\%}} & 14.63\textcolor{red}{\scriptsize{$\downarrow$83\%}} & 9.89\textcolor{red}{\scriptsize{$\downarrow$89\%}} & 22.69\textcolor{red}{\scriptsize{$\downarrow$72\%}} & 16.73\textcolor{red}{\scriptsize{$\downarrow$81\%}} \\
    
    \midrule
    \end{tabular}  
    \end{adjustbox}
\end{table*}

\begin{table*}[h]
    \centering
    \small
    \caption{Adversarial Attack Results on Detection using frozen SSL and Supervised Models as backbones. The table presents performance metrics under clean and adversarial conditions for various attack types (Optim, BIM, MIM, SGD, PGD, Optim-Adam, Optim-Nesterov). The last two rows display clean mean Average Precision (mAP) and the average performance under adversarial attacks, with the percentage decrease in performance highlighted in red}
    \label{tab:table8}

    \begin{adjustbox}{width=\textwidth}
    \begin{tabular}{lllllllll}
    \toprule
    {} & Barlow & BYOL & DINO & MocoV3 & SimCLR & Supervised & SwAV & VICReg \\    
    \midrule
    Optim & 3.98 & 1.05 & 2 & 2.6 & 0.78 & 0.56 & 1.51 & 0.39 \\
    BIM & 44.87 & 32.24 & 54.93 & 26.72 & 6.79 & 42.8 & 44.47 & 10.32 \\
    MIM & 11.37 & 3.04 & 10.32 & 5.72 & 6.0 & 4.73 & 10.87 & 2.68 \\
    SGD & 3.21 & 1.28 & 2.95 & 9.44 & 1.42 & 1.02 & 2.85 & 1.72 \\
    PGD & 83.08 & 80.83 & 79.65 & 79.83 & 74.68 & 75.29 & 79.14 & 81.27 \\
    Optim-Adam & 4.71 & 0.76 & 3.5 & 2.03 & 0.75 & 0.87 & 3.46 & 0.67 \\
    Optim-Nesterov & 1.75 & 0.64 & 0.97 & 2.77 & 0.94 & 0.72 & 1.1 & 0.64 \\
    \hline
    Clean mAP & 88.39 & 87.44 & 87.63 & 87.36 & 87.67 & 87.67 & 86.55 & 88.43 \\
    Adv mAP & 21.85\textcolor{red}{\scriptsize{$\downarrow$75\%}} & 17.12\textcolor{red}{\scriptsize{$\downarrow$80\%}} & 22.05\textcolor{red}{\scriptsize{$\downarrow$75\%}} & 18.44\textcolor{red}{\scriptsize{$\downarrow$79\%}} & 13.05\textcolor{red}{\scriptsize{$\downarrow$85\%}} & 18.00\textcolor{red}{\scriptsize{$\downarrow$79\%}} & 20.49\textcolor{red}{\scriptsize{$\downarrow$76\%}} & 13.96\textcolor{red}{\scriptsize{$\downarrow$84\%}} \\
    \midrule
    \end{tabular}  
    \end{adjustbox}
\end{table*}
\clearpage
\subsubsection{COCO}

\begin{table*}[h]
    \centering
    \small
    \caption{Adversarial Attack Results on Detection using frozen SSL and Supervised Models as backbones. The table presents performance metrics under clean and adversarial conditions for various attack types (Optim, BIM, MIM, SGD, PGD, Optim-Adam, Optim-Nesterov). The last two rows display clean mean Average Precision (mAP) and the average performance under adversarial attacks, with the percentage decrease in performance highlighted in red}
    \label{tab:table8}

    \begin{adjustbox}{width=\textwidth}
    \begin{tabular}{lllllllll}
    \toprule
    {} & Barlow & BYOL & DINO & MocoV3 & SimCLR & Supervised & SwAV & VICReg \\   
    \midrule
    Optim & 18.6 & 16.47 & 11.05 & 15.43 & 11.71 & 9.12 & 13.8 & 13.97 \\
    BIM & 18.6 & 18.58 & 21.02 & 18.23 & 13.06 & 16.74 & 20.03 & 18.47 \\
    MIM & 21.07 & 18.37 & 16.38 & 16.47 & 14.05 & 13.01 & 17.55 & 17.67 \\
    SGD & 17.62 & 18.38 & 19.19 & 21.7 & 12.96 & 9.22 & 17.41 & 16.69 \\
    PGD & 21.05 & 20.5 & 21.68 & 22.22 & 14.2 & 19.61 & 21.44 & 20.96 \\
    Optim-Adam & 19.15 & 16.36 & 11.12 & 15.83 & 12 & 9.06 & 13.92 & 14.09 \\
    Optim-Nesterov & 15.76 & 16.62 & 13.37 & 15.31 & 11.67 & 9.08 & 13.91 & 13.44 \\
    \hline
    Clean mAP & 37.93 & 39.49 & 38.57 & 40.49 & 32.17 & 37.01 & 38.45 & 38.55 \\
    Adv Avg. & 18.84\textcolor{red}{\scriptsize{$\downarrow$50\%}} & 17.90\textcolor{red}{\scriptsize{$\downarrow$55\%}} & 16.26\textcolor{red}{\scriptsize{$\downarrow$58\%}} & 17.88\textcolor{red}{\scriptsize{$\downarrow$56\%}} & 12.80\textcolor{red}{\scriptsize{$\downarrow$60\%}} & 12.26\textcolor{red}{\scriptsize{$\downarrow$67\%}} & 16.87\textcolor{red}{\scriptsize{$\downarrow$56\%}} & 16.47\textcolor{red}{\scriptsize{$\downarrow$57\%}} \\
    \midrule
    \end{tabular}  
    \end{adjustbox}
\end{table*}
\clearpage
\subsection{ResNet vs ViT}\label{res:vit}
\begin{table*}[h]
    \centering
    \scriptsize
    \caption{This table presents the results of various instance and universal adversarial perturbation (UAP) attacks on the Imagenet-1k dataset, with all UAP attack names in \textit{italics}. Different configurations of FGSM and PGD are denoted, such as $FGSM_1$ and $PGD_1$. Average results for universal adversarial perturbations (UAP Avg.), instance adversarial attacks (IAA Avg.), and overall adversarial performance (Adv Avg.) are reported at the bottom, including percentage drops relative to clean accuracy}
    \label{tab:few_shot_small_domain_shift}
    
    \begin{adjustbox}{width=0.99\textwidth}
    \begin{tabular}{llllllll}
    \toprule
    {} & MoCoV3-ViT-B & DINO-ViT-B & MoCo-ViT & DINO-ViT & DINO-ResNet & MoCoV3-ResNet &\\
    \midrule
    $FGSM_1$ & 46.69&	58.36& 34.63 & 51.42 & 24.68 & 42.67  \\
    $FGSM_2$ & 13.21&	22.17 & 0.32 & 0.97 & 5.66& 15.53\\
    
    $PGD_1$ & 45.06&	57.88 & 33.35 & 50.98 & 25.65& 42.39\\
    $PGD_2$ & 1.14&	8.58& 0.00 & 0.00 & 0.18& 1.06\\
    
    $PGD_3$ & 45.1&	57.89 & 33.46 & 50.95 & 25.85& 42.56\\
    $PGD_4$ & 1.05	&8.36 & 0.17 & 3.84 & 0.14& 0.82\\
    
    $PGD_5$ & 6.86	&22.58& 2.12 & 13.57 & 0.67 & 4.79\\
    
    DIFGSM & 45.13&	57.89& 51.91 & 59.81 & 41.12& 54.09\\
    
    CW &0	&0& 0 & 0 & 0.02\\
    
    Jitter & 63.52	&68.96 & 58.25 & 66.30 & 60.26& 62.47\\
    
    TIFGSM  & 66.17	&68.58 & 61.84 & 65.23 & 56.98 & 61.47\\
    
    PIFGSM & 38.14	&55.06 & 25.78 & 47.64 & 14.54& 34.10\\
    
    EADEN & 0 & 0& 0 & 0 & 0& 0\\
    OnePixel & 74.93&	76.67 & 71.28 & 75.47 & 72.83& 72.64\\
    Pixle  & 42.85&	49.51& 34.69 & 44.08 & 19.41& 31.45\\
    SPSA & 71.18&	74.35 & 66.20 & 72.47 & 68.11& 69.93\\
    Square &1.87&	2.78 & 1.22 & 1.67 & 1.30& 3.15\\
    TAP & 75.66	&76.78 & 72.34 & 75.60 & 73.78 & 73.72\\
    \textit{ASV}               & 0.10 & 0.10 & 0.10 & 0.10 & 68.69 & 67.59 \\
    \textit{FFF (mean-std)}      & 76.38 & 77.45 & 72.55 & 76.71 & 59.45 & 46.78 \\
    
    \textit{FFF (no-data)}   & 76.34 & 77.45 & 72.66 & 76.71 & 40.73 & 57.39 \\
    \textit{FFF (one-sample)}  & 76.34 & 77.45 & 72.55 & 76.71 & 16.15 & 57.18 \\
    \textit{FG-UAP}            & 1.37  & 6.79  & 0.72  & 3.51  & 2.30  & 3.08  \\
    \textit{GD-UAP (no-data)}  & 53.08 & 43.49 & 1.16  & 7.30  & 26.59 & 43.66 \\
    \textit{GD-UAP (mean-std)} & 56.97 & 68.29 & 28.80 & 64.17 & 49.85 & 53.59 \\
    \textit{GD-UAP (one-sample)} & 16.00 & 64.12 & 2.04  & 52.85 & 25.90 & 35.10 \\
    \textit{L4A-base}         & 23.14 & 19.58 & 0.69  & 46.28 & 51.13 & 41.09 \\
    \textit{L4A-fuse}         & 24.13 & 19.06 & 0.73  & 46.30 & 50.69 & 40.39 \\
    \textit{L4A-ugs}          & 6.01  & 21.65 & 2.35  & 0.27  & 56.83 & 65.27 \\
    \textit{PD-UAP}           & 74.72 & 76.81 & 70.61 & 74.91 & 45.06 & 56.25 \\
    \textit{SSP}              & 29.27 & 60.91 & 2.52  & 56.53 & 38.55 & 48.56 \\
    \textit{STD}              & 54.52 & 72.43 & 23.01 & 71.66 & 52.44 & 53.44 \\
    \textit{UAP (DeepFool)}   & 2.19  & 8.50  & 1.15  & 8.56  & 15.18 & 24.25 \\
    \textit{UAPEPGD}          & 73.86 & 76.26 & 69.34 & 74.51 & 70.60 & 71.57 \\
    \hline
    Clean Accuracy  & 76.66 & 77.99 & 73.21 & 76.95 & 75.28 & 74.57 \\
    IAA Avg. & 35.47 \textcolor{red}{\tiny{$\downarrow$54\%}}  & 42.58 \textcolor{red}{\tiny{$\downarrow$45\%}}  & 30.42\textcolor{red}{\tiny{$\downarrow$58\%}} & 37.78\textcolor{red}{\tiny{$\downarrow$51\%}}  & 27.29\textcolor{red}{\tiny{$\downarrow$64\%}}  & 34.05 \textcolor{red}{\tiny{$\downarrow$54\%}}\\
    UAP Avg. & 40.28 \textcolor{red}{\tiny{$\downarrow$47\%}} & 48.15 \textcolor{red}{\tiny{$\downarrow$38\%}}   & 26.31\textcolor{red}{\tiny{$\downarrow$64\%}} & 46.07\textcolor{red}{\tiny{$\downarrow$40\%}}  & 41.88\textcolor{red}{\tiny{$\downarrow$44\%}} & 47.82 \textcolor{red}{\tiny{$\downarrow$36\%}} \\

    Adv Avg. &  37.73 \textcolor{red}{\tiny{$\downarrow$51\%}} &  45.19 \textcolor{red}{\tiny{$\downarrow$42\%}}  & 28.49\textcolor{red}{\tiny{$\downarrow$61\%}} & 41.67\textcolor{red}{\tiny{$\downarrow$46\%}}  & 34.15\textcolor{red}{\tiny{$\downarrow$55\%}} &  40.53 \textcolor{red}{\tiny{$\downarrow$46\%}} \\

    \midrule
    \end{tabular}
    \end{adjustbox}
\end{table*}

\clearpage
\subsection{DINOv2 and MAE}\label{res:dinov2}

\begin{table*}[h]
    \centering
    \scriptsize
    \caption{This table presents the results of various instance and universal adversarial perturbation (UAP) attacks on the Imagenet-1k dataset, with all UAP attack names in \textit{italics}. Different configurations of FGSM and PGD are denoted, such as $FGSM_1$ and $PGD_1$. Average results for universal adversarial perturbations (UAP Avg.), instance adversarial attacks (IAA Avg.), and overall adversarial performance (Adv Avg.) are reported at the bottom, including percentage drops relative to clean accuracy}
    \label{tab:few_shot_small_domain_shift}
    
    \begin{adjustbox}{width=0.7\textwidth}
    \begin{tabular}{lllll}
    \toprule
    {} & DINOv2-S & DINOv2-B & MAE &\\
    \midrule
    $FGSM_1$ & 43.00 &	52.25 & 56.72  \\
    $FGSM_2$ & 6.05 &	14.16 & 41.61 \\
    
    $PGD_1$ & 38.84 & 48.53 & 45.13 \\
    $PGD_2$ & 0.26 &	0.75 & 2.73 \\
    
    $PGD_3$ & 38.93 & 48.60 & 45.33 \\
    $PGD_4$ & 0.25 & 0.69 & 1.92 \\
    
    $PGD_5$ & 2.83	& 4.17 8& 3.39 \\
    
    DIFGSM & 56.25 &	60.55 & 63.10 \\
    
    CW &0	& 0 & 0.06 \\
    
    Jitter & 63.10	& 69.52 & 69.4 \\
    
    TIFGSM  & 68.83	& 73.35 & 72.17 \\
    
    PIFGSM & 37.35	& 45.16 & 44.42 \\
    
    EADEN & 0 & 0& 0 \\
    OnePixel & 79.48 & 82.84 & 82.39 \\
    Pixle  & 47.93 &	61.67 & 61.66 \\
    SPSA & 76.05 & 80.22 & 78.56 \\
    Square & 0.33 &	1.96 & 0.12 \\
    TAP & 80.49	& 84.05 & 82.95\\
    \textit{ASV}               & 0.10 & 0.10 & 0.10 \\
    \textit{FFF (mean-std)}      & 79.68 & 83.78 & 74.19 \\
    
    \textit{FFF (no-data)}   & 79.68 & 83.80 & 78.05 \\
    \textit{FFF (one-sample)}  & 79.68 & 83.78 & 80.00 &\\
    \textit{FG-UAP}            & 1.29  & 1.59  & 1.55    \\
    \textit{GD-UAP (no-data)}  & 13.57  & 74.98  & 78.06 \\
    \textit{GD-UAP (mean-std)} & 0.14 &  35.22  & 78.89  \\
    \textit{GD-UAP (one-sample)} & 3.16 & 51.71 & 79.52 \\
    \textit{L4A-base}         & 51.89 & 9.56 & 14.67  \\
    \textit{L4A-fuse}         & 52.07 & 8.51 & 21.18  \\
    \textit{L4A-ugs}          & 5.37  & 53.86 & 0.43  \\
    \textit{PD-UAP}           & 18.74 & 83.33 & 17.66 \\
    \textit{SSP}              & 0.13 & 21.93 & 28.77  \\
    \textit{STD}              & 68.38 & 80.17 & 77.00 \\
    \textit{UAP (DeepFool)}   & 12.24  & 20.24 & 28.99  \\
    \textit{UAPEPGD}          & 79.51 & 83.45 & 82.25 \\
    \hline
    Clean Accuracy  & 81.33 & 84.42 & 83.58 \\
    IAA Avg. & 35.55 \textcolor{red}{\tiny{$\downarrow$56\%}}  & 40.47 \textcolor{red}{\tiny{$\downarrow$52\%}}  & 41.76\textcolor{red}{\tiny{$\downarrow$50\%}} \\
    UAP Avg. & 32.10 \textcolor{red}{\tiny{$\downarrow$61\%}} & 45.65 \textcolor{red}{\tiny{$\downarrow$46\%}}   & 43.61\textcolor{red}{\tiny{$\downarrow$48\%}} \\

    Adv Avg. &  34.87 \textcolor{red}{\tiny{$\downarrow$57\%}} &  44.25 \textcolor{red}{\tiny{$\downarrow$48\%}}  & 43.91\textcolor{red}{\tiny{$\downarrow$47\%}} \\

    \midrule
    \end{tabular}
    \end{adjustbox}
\end{table*}
\clearpage
\subsection{ImageNet Across Training Epochs}\label{res:dur}
\begin{table*}[h]
    \centering
    \scriptsize
    \caption{This table presents the results of various instance and universal adversarial perturbation (UAP) attacks on the Imagenet-1k dataset, with all UAP attack names in \textit{italics}. Different configurations of FGSM and PGD are denoted, such as $FGSM_1$ and $PGD_1$. Average results for universal adversarial perturbations (UAP Avg.), instance adversarial attacks (IAA Avg.), and overall adversarial performance (Adv Avg.) are reported at the bottom, including percentage drops relative to clean accuracy.}
    \label{tab:few_shot_small_domain_shift}

    \begin{adjustbox}{width=0.7\textwidth}
    \begin{tabular}{llllllll}
    \toprule
    {} & MoCoV3-100 & MoCoV3-300 & MoCoV3-1000 \\
    \midrule
    
    $FGSM_1$ & 38.87 & 42.6 & 42.67\\
    $FGSM_2$ & 7.94 & 8.38 & 15.53\\
    
    $PGD_1$ & 37.89 & 41.99 & 42.39\\
    $PGD_2$ & 0.49 & 0.09 & 1.06\\
    
    $PGD_3$ & 38.06 & 42.14 & 42.56\\
    $PGD_4$ & 1.75 & 1.22 & 0.82\\
    
    $PGD_5$ & 5.4 & 5.49 & 4.79\\
    
    DIFGSM & 49.21 & 52.65 & 54.09\\
    
    CW & 0.02 & 0.02 & 0.02 \\
    
    Jitter & 56.45 & 60.53 & 62.47\\
    
    TIFGSM & 57.39 & 61.86 & 61.47\\
    
    PIFGSM & 31.24 & 34.41 & 34.1\\
    
    EADEN & 0 & 0 & 0\\
    OnePixel & 66.79 & 70.76 & 72.64\\
    Pixle & 26.27 & 29.41 & 31.45\\
    SPSA & 64.05 & 68.02 & 69.93\\
    Square & 2.05 & 2.01 & 3.15\\
    TAP & 67.85 & 71.9 & 73.72\\
    \textit{ASV} & 62.59 & 66.11 & 67.59\\
    \textit{FFF (mean-std)} & 50.05 & 59.76 &  46.78\\
    \textit{FFF (no-data)} & 32.96 & 36.39 & 57.39\\
    \textit{FFF (one-sample)} & 40.81 & 58.32 & 57.18\\
    \textit{FG-UAP} & 2.72 & 5.07 & 3.08\\
    \textit{GD-UAP (mean-std)} & 49.56 & 53.45 & 43.66\\
    \textit{GD-UAP (no-data)} & 37.00 & 46.68 & 53.59\\
    \textit{GD-UAP (one-sample)} & 32.04 & 37.17 & 35.10\\
    \textit{L4A-base} & 40.09 & 33.66 & 41.09\\
    \textit{L4A-fuse} & 40.07 & 33.70 & 40.39\\
    \textit{L4A-ugs} & 51.79 & 64.64 & 65.27\\
    \textit{PD-UAP} & 40.96 & 55.29 & 56.25\\
    \textit{SSP} & 42.24 & 26.80 & 48.56\\
    \textit{STD} & 49.03 & 51.07 & 53.44\\
    \textit{UAP (DeepFool)} & 24.71 & 26.99 & 24.25\\
    \textit{UAPEPGD} & 65.51 & 69.75 & 71.57\\
    \hline
    Clean Accuracy & 68.91 & 72.82 & 74.57 \\
    IAA Avg. & 30.65 \textcolor{red}{\tiny{$\downarrow$56\%}}  & 32.97\textcolor{red}{\tiny{$\downarrow$55\%}} & 34.05\textcolor{red}{\tiny{$\downarrow$54\%}}  \\
    UAP Avg. & 41.39 \textcolor{red}{\tiny{$\downarrow$40\%}}  & 45.30\textcolor{red}{\tiny{$\downarrow$38\%}} & 4782\textcolor{red}{\tiny{$\downarrow$36\%}}  \\

    Adv Avg. &  35.70 \textcolor{red}{\tiny{$\downarrow$48\%}}  & 38.77\textcolor{red}{\tiny{$\downarrow$47\%}} & 40.53\textcolor{red}{\tiny{$\downarrow$47\%}}  \\
    \midrule
    \end{tabular}
    \end{adjustbox}
\end{table*}

    %
%
%
%
%
%
%
%
%
%
%
%
%
%
\clearpage
\begin{table*}[h]
    \centering
    \scriptsize
    \caption{This table presents the results of various instance and universal adversarial perturbation (UAP) attacks on the Imagenet-1k dataset, with all UAP attack names in \textit{italics}. Different configurations of FGSM and PGD are denoted, such as $FGSM_1$ and $PGD_1$. Average results for universal adversarial perturbations (UAP Avg.), instance adversarial attacks (IAA Avg.), and overall adversarial performance (Adv Avg.) are reported at the bottom, including percentage drops relative to clean accuracy}
    \label{tab:few_shot_small_domain_shift}
    
    \begin{adjustbox}{width=0.7\textwidth}
    \begin{tabular}{lllll}
    \toprule
    {} & SwAV-100 & SwAV-200 & SwAV-400 & SwAV-800\\
    \midrule
    
    $FGSM_1$ & 18.08 & 19.99 & 21.9 & 24.71\\
    $FGSM_2$ & 4.01 & 4.34 & 5.2 & 6.35\\
    
    $PGD_1$ & 18.94 & 21.3 & 23.7 & 26.48\\
    $PGD_2$ & 0.31 & 0.17 & 0.17 & 0.18\\
    
    $PGD_3$ & 19.08 & 21.44 & 23.88 & 26.73\\
    $PGD_4$ & 0.3 & 0.15 & 0.14 & 0.14\\
    
    $PGD_5$& 0.73 & 0.59 & 0.52 & 0.69\\
    
    DIFGSM & 39.31 & 42.01 & 42.31 & 45.65\\
    
    CW & 0.0 & 0.0 & 0.0 & 0.0 \\
    
    Jitter & 56.67 & 59.15 & 60.43 & 61.16\\
    
    TIFGSM & 53.11 & 55.14 & 56.44 & 60.07\\
    
    PIFGSM  & 10 & 10.87 & 11.76 & 14.12\\
    
    EADEN & 0 & 0 & 0 & 0\\
    OnePixel & 68.73 & 70.83 & 71.64 & 72.73\\
    Pixle  & 13.21 & 16.03 & 18.08 & 16.95\\
    SPSA & 63.94 & 66.25 & 67.38 & 68.61\\
    Square & 0.35 & 0.36 & 0.5 & 1.99\\
    TAP & 71.79 & 73.56 & 74.37 & 75.05\\
    \textit{ASV} & 64.00 & 63.90 & 67.06 & 67.66 \\
    \textit{FFF (mean-std)} & 55.81 & 58.52 & 62.60 & 61.14 \\
    \textit{FFF (no-data)} & 35.02 & 24.83 & 35.81 &  32.15 \\
    \textit{FFF (one-sample)} & 24.75 & 24.22 & 34.59 & 29.96 \\
    \textit{FG-UAP} & 1.86 & 3.56 & 2.29 & 2.17 \\
    \textit{GD-UAP (mean-std)} & 54.07 & 58.14 & 58.08 & 57.38 \\
    \textit{GD-UAP (no-data)} & 22.09 & 29.45 & 26.57 & 36.07\\
    \textit{GD-UAP (one-sample)} & 21.33 & 35.80 & 37.18 & 36.37 \\
    \textit{L4A-base} & 18.63 & 17.66 & 33.43 & 10.90 \\
    \textit{L4A-fuse} & 18.58 & 17.908 & 34.73 & 10.94 \\
    \textit{L4A-ugs} & 34.02 & 37.03 & 50.20 & 49.91 \\
    \textit{PD-UAP} & 41.58 & 54.83 & 51.53 & 51.75 \\
    \textit{SSP} & 12.12 & 25.74 & 16.642 & 23.27 \\
    \textit{STD} & 45.74 & 53.71 & 45.88 & 56.08 \\
    \textit{UAP (DeepFool)} & 10.17 & 10.09 & 10.79 & 8.59 \\
    \textit{UAPEPGD} & 67.13 & 69.01 & 70.07 & 70.98 \\

    \hline
    Clean Accuracy & 72.02 & 73.82 & 74.57 & 75.27 \\
    IAA Avg. & 24.36 \textcolor{red}{\tiny{$\downarrow$66\%}}  & 25.68\textcolor{red}{\tiny{$\downarrow$65\%}} & 26.58\textcolor{red}{\tiny{$\downarrow$64\%}}  & 27.87\textcolor{red}{\tiny{$\downarrow$63\%}} \\
    UAP Avg. & 32.93 \textcolor{red}{\tiny{$\downarrow$54\%}}  & 36.52\textcolor{red}{\tiny{$\downarrow$51\%}} & 39.84\textcolor{red}{\tiny{$\downarrow$47\%}}  & 37.83\textcolor{red}{\tiny{$\downarrow$50\%}} \\

    Adv Avg. &  28.40 \textcolor{red}{\tiny{$\downarrow$60\%}}  & 30.78\textcolor{red}{\tiny{$\downarrow$58\%}} & 32.82\textcolor{red}{\tiny{$\downarrow$56\%}}  & 32.55\textcolor{red}{\tiny{$\downarrow$57\%}} \\
    \midrule
    \end{tabular}
    \end{adjustbox}
\end{table*}

\clearpage
\subsection{Imagenet With Different MoCo Versions}\label{res:moco}
\begin{table*}[h]
    \centering
    \scriptsize

    \caption{This table presents the results of various instance and universal adversarial perturbation (UAP) attacks on the Imagenet-1k dataset, with all UAP attack names in \textit{italics}. Different configurations of FGSM and PGD are denoted, such as $FGSM_1$ and $PGD_1$. Average results for universal adversarial perturbations (UAP Avg.), instance adversarial attacks (IAA Avg.), and overall adversarial performance (Adv Avg.) are reported at the bottom, including percentage drops relative to clean accuracy}
    \label{tab:few_shot_small_domain_shift}
    
    \begin{adjustbox}{width=0.7\textwidth}
    \begin{tabular}{llllllll}
    \toprule
    {} & MoCoV1 & MoCoV2  & MoCoV3\\
    \midrule
    $FGSM_1$ & 15.91 & 22.01 & 42.67 \\
    $FGSM_2$ & 6.25 & 5.17 & 15.53 \\
    
    $PGD_1$ & 17.89 & 24.00 & 42.39 \\
    $PGD_2$ & 0.09 & 0.54 & 1.06\\
    
    $PGD_3$ & 17.96 & 24.14 & 42.56\\
    $PGD_4$ & 0.06 & 0.52 & 0.82 \\
    
    $PGD_5$ & 0.21 & 1.33 & 4.79 \\
    
    DIFGSM  & 34.85 & 40.39 & 54.09 \\
    
    CW  & 0 & 0 & 0.02 & \\
    
    Jitter & 50.04 & 53.09 & 62.47 \\
    
    TIFGSM & 48.70 & 49.50 & 61.47 \\
    
    PIFGSM & 8.53 & 13.20 & 34.10 \\
    
    EADEN & 0 & 0 & 0 \\
    OnePixel & 56.67 & 64.63 & 72.64 \\
    Pixle  & 3.10 & 17.85 & 31.45\\
    SPSA & 50.62 & 60.57 & 69.93 \\
    Square & 0.80 & 0.42 & 3.15 \\
    TAP & 58.55 & 65.24 & 73.72 \\
    \textit{ASV} & 18.756 & 60.99 & 67.59\\
    \textit{FFF (mean-std)} & 30.34 & 46.83 & 46.78\\
    \textit{FFF (no-data)} & 19.47 & 33.01 & 57.39\\
    \textit{FFF (one-sample)} & 5.40 & 42.08 & 57.18\\
    \textit{FG-UAP} & 0.838 & 3.78 & 3.08\\
    \textit{GD-UAP (mean-std)} & 5.88 & 51.28 &   53.39\\
    \textit{GD-UAP (no-data)} & 18.34 & 44.34 & 43.66\\
    \textit{GD-UAP (one-sample)} & 3.50 & 39.28 & 35.10\\
    \textit{L4A-base} & 2.15 & 37.09 & 41.09\\
    \textit{L4A-fuse} & 2.19 & 36.79 & 40.39\\
    \textit{L4A-ugs} & 2.25 & 30.16 & 65.27\\
    \textit{PD-UAP} & 33.96 & 50.06 & 56.25\\
    \textit{SSP} & 4.59 & 42.66 & 48.56\\
    \textit{STD} & 23.09 & 46.31 & 53.44\\
    \textit{UAP (DeepFool)} & 4.34 & 30.64 & 24.25\\
    \textit{UAPEPGD} & 47.67 & 63.44 & 71.57\\

    \hline
    Clean Accuracy & 60.64 & 67.72 & 74.57 \\
    IAA Avg. & 20.56 \textcolor{red}{\tiny{$\downarrow$66\%}}  & 24.59\textcolor{red}{\tiny{$\downarrow$64\%}} & 34.05\textcolor{red}{\tiny{$\downarrow$54\%}}  \\
    UAP Avg. & 13.92 \textcolor{red}{\tiny{$\downarrow$77\%}}  & 41.17\textcolor{red}{\tiny{$\downarrow$39\%}} & 47.82\textcolor{red}{\tiny{$\downarrow$36\%}}  \\

    Adv Avg. &  17.44 \textcolor{red}{\tiny{$\downarrow$71\%}}  & 32.40\textcolor{red}{\tiny{$\downarrow$52\%}} & 40.53\textcolor{red}{\tiny{$\downarrow$46\%}}  \\
    
    \midrule
    \end{tabular}
    \end{adjustbox}
\end{table*}

\clearpage
\subsection{BYOL Ablations}\label{res:byol_abl}
\begin{table*}[h]
    \centering
    \scriptsize
    \caption{This table presents the results of various instance and universal adversarial perturbation (UAP) attacks on the Imagenet-1k dataset, with all UAP attack names in \textit{italics}. Different configurations of FGSM and PGD are denoted, such as $FGSM_1$ and $PGD_1$. Average results for universal adversarial perturbations (UAP Avg.), instance adversarial attacks (IAA Avg.), and overall adversarial performance (Adv Avg.) are reported at the bottom, including percentage drops relative to clean accuracy.}
    \label{tab:few_shot_small_domain_shift}

    \begin{adjustbox}{width=0.9\textwidth}
    \begin{tabular}{llllllll}
    \toprule
    {} & BYOL-NC & BYOL-CC & BYOL-128 & BYOL-512 & BYOL  \\
    \midrule
    
    $FGSM_1$  &23.45&	31.23&	35.33	&37.51 & 39.41 \\
    $FGSM_2$  &10.6	&11.36	&13.62	&13.92& 13.47\\
    
    $PGD_1$  &25.57&	30.86	&35.32&	37.69 & 39.63\\
    $PGD_2$  & 0.68	&0.79&	1.3&	1& 0.65 \\
    
    $PGD_3$  & 25.83&	30.89&	35.52	&37.69 & 39.82 \\
    $PGD_4$  & 0.59	&0.66&	1.1	&0.83 & 0.5 \\
    
    $PGD_5$ & 1.22	&1.89&	3.51&	3.69 & 3.44 \\
    
    DIFGSM  & 35.73	&45.67	&47.83&	49.61& 52.71 \\
    
    CW & 0.02&	0.01	&0.1	&0.02 & 0.02 \\
    
    Jitter & 53	&56.81&	57.7	&60.02 & 61.92 \\
    
    TIFGSM & 40.89&	58.33	&56.77	&60.11 & 62.27\\
    
    PIFGSM & 12.27&	22.2	&25.86	&28.48 & 29.83 \\
    
    EADEN & 0	&0	&0&	0 & 0 \\
    OnePixel & 61.18&	66.27&	67.31	&68.89 & 72.5 \\
    Pixle & 27.1	&9.84&	25&	26.13 & 28.67 \\
    SPSA & 55.76	&63.8&	64.24	&66.94 & 69.59 \\
    Square & 0.28	&2.13&	2.11	&2.04 & 2.62 \\
    TAP & 39.38	&68.58&	69.31&	71.71 & 74.36 \\
    \textit{ASV} & 51.27 & 51.41 & 62.74 & 66.16 & 68.46 \\
    \textit{FFF (mean-std)} & 43.19 & 51.74 & 51.84 & 60.94 &  62.38 \\
    
    \textit{FFF (no-data)} & 42.24 & 30.11 & 39.31 & 43.11 & 31.94 \\
    \textit{FFF (one-sample)} & 16.55 & 23.80 & 36.39 & 37.32 & 54.41 \\
    
    \textit{FG-UAP} & 1.15 & 1.85 & 3.20 & 3.53 & 1.83 \\
    \textit{GD-UAP (no-data)} & 39.36 & 54.65 & 53.00 & 56.74 & 61.10\\
    
    \textit{GD-UAP (mean-std)} & 32.64 & 38.54 & 41.85 & 42.88 & 37.96 \\
    
    \textit{GD-UAP (one-sample)} & 33.28 & 25.60 & 37.95 & 35.48 & 37.54 \\
    
    \textit{L4A-base} & 40.59 & 32.82 & 32.86 & 22.65 & 36.35 \\
    
    \textit{L4A-fuse} & 39.98 & 32.97 & 32.77 & 22.58 & 36.01 \\
    
    \textit{L4A-ugs} & 39.77 & 35.36 & 54.70 & 53.27 & 60.03 \\
    
    \textit{PD-UAP} & 39.90 & 41.47 & 50.45 & 50.02 & 54.06 \\
    
    \textit{SSP} & 21.82 & 35.16 & 26.31 & 22.75 & 41.05 \\
    
    \textit{STD} & 36.97 & 47.11 & 53.42 & 53.72 & 58.08 \\
    
    \textit{UAP (DeepFool)} & 18.58 & 10.96 & 19.48 & 22.53 & 20.25 \\
    
    \textit{UAPEPGD} & 56.11 & 65.48 & 66.16 & 68.83 & 71.21 \\
    Clean Accuracy & 63.77	&69.15&	69.67 &	72.09 & 74.57  \\
    \midrule
    IAA Avg. & 22.98 \textcolor{red}{\tiny{$\downarrow$60\%}}  & 27.85\textcolor{red}{\tiny{$\downarrow$60\%}} & 30.11\textcolor{red}{\tiny{$\downarrow$57\%}}
    & 31.46\textcolor{red}{\tiny{$\downarrow$56\%}} & 32.86\textcolor{red}{\tiny{$\downarrow$56\%}}\\
    UAP Avg. & 34.59 \textcolor{red}{\tiny{$\downarrow$46\%}}& 36.19\textcolor{red}{\tiny{$\downarrow$48\%}}  & 41.40\textcolor{red}{\tiny{$\downarrow$41\%}} & 41.41\textcolor{red}{\tiny{$\downarrow$43\%}}
    & 45.79\textcolor{red}{\tiny{$\downarrow$39\%}}\\

    Adv Avg. & 28.44\textcolor{red}{\tiny{$\downarrow$55\%}}  & 31.78\textcolor{red}{\tiny{$\downarrow$54\%}}& 35.42\textcolor{red}{\tiny{$\downarrow$49\%}}
    & 36.14\textcolor{red}{\tiny{$\downarrow$50\%}} & 38.94\textcolor{red}{\tiny{$\downarrow$48\%}}\\

    \midrule
    \end{tabular}
    \end{adjustbox}
\end{table*}
\clearpage
\subsection{Transfer Learning (Linear)}\label{res:combined}
\begin{table*}[htbp]
\centering
\scriptsize
\caption{Combined results from transfer learning datasets showing Clean accuracy, UAP Avg., IAA Avg., and Adv Avg. with percentage drops relative to Clean accuracy.}
\label{tab:results}
\begin{tabular}{@{}llllllllll@{}}
\toprule
{} & {} & Barlow & BYOL & DINO & MocoV3 & SimCLR & Supervised & SwAV & VICReg \\
\midrule
\multirow{4}{*}{\rotatebox{90}{Aircraft}} 
     & Clean & 56.88 & 56.34 & 60.25 & 58.75 & 46.77 & 44.89 & 54.01 & 56.43 \\
     & IAA & 16.29\textcolor{red}{\tiny{$\downarrow$71\%}} & 14.87\textcolor{red}{\tiny{$\downarrow$73\%}} & 15.27\textcolor{red}{\tiny{$\downarrow$75\%}} & 17.41\textcolor{red}{\tiny{$\downarrow$70\%}} & 11.93\textcolor{red}{\tiny{$\downarrow$74\%}} & 9.82\textcolor{red}{\tiny{$\downarrow$78\%}} & 13.82\textcolor{red}{\tiny{$\downarrow$74\%}} & 16.38\textcolor{red}{\tiny{$\downarrow$71\%}} \\
     & UAP & 20.22\textcolor{red}{\tiny{$\downarrow$64\%}} & 19.64\textcolor{red}{\tiny{$\downarrow$65\%}} & 19.25\textcolor{red}{\tiny{$\downarrow$68\%}} & 25.57\textcolor{red}{\tiny{$\downarrow$56\%}} & 16.30\textcolor{red}{\tiny{$\downarrow$65\%}} & 15.54\textcolor{red}{\tiny{$\downarrow$65\%}} & 16.01\textcolor{red}{\tiny{$\downarrow$70\%}} & 19.82\textcolor{red}{\tiny{$\downarrow$64\%}} \\
     & Adv & 18.14\textcolor{red}{\tiny{$\downarrow$68\%}} & 17.11\textcolor{red}{\tiny{$\downarrow$70\%}} & 17.14\textcolor{red}{\tiny{$\downarrow$72\%}} & 21.25\textcolor{red}{\tiny{$\downarrow$64\%}} & 13.99\textcolor{red}{\tiny{$\downarrow$70\%}} & 12.51\textcolor{red}{\tiny{$\downarrow$72\%}} & 14.85\textcolor{red}{\tiny{$\downarrow$73\%}} & 18.00\textcolor{red}{\tiny{$\downarrow$68\%}} \\
\midrule
\multirow{4}{*}{\rotatebox{90}{Caltech}} 
     & Clean & 90.54 & 90.99 & 90.31 & 92.89 & 89.10 & 90.25 & 90.36 & 90.57 \\
     & IAA & 53.60\textcolor{red}{\tiny{$\downarrow$41\%}} & 54.06\textcolor{red}{\tiny{$\downarrow$41\%}} & 47.42\textcolor{red}{\tiny{$\downarrow$47\%}} & 58.23\textcolor{red}{\tiny{$\downarrow$37\%}} & 49.79\textcolor{red}{\tiny{$\downarrow$44\%}} & 44.10\textcolor{red}{\tiny{$\downarrow$51\%}} & 45.55\textcolor{red}{\tiny{$\downarrow$50\%}} & 53.64\textcolor{red}{\tiny{$\downarrow$41\%}} \\
     & UAP & 66.48\textcolor{red}{\tiny{$\downarrow$27\%}} & 69.06\textcolor{red}{\tiny{$\downarrow$16\%}} & 65.18\textcolor{red}{\tiny{$\downarrow$28\%}} & 74.95\textcolor{red}{\tiny{$\downarrow$7\%}} & 66.50\textcolor{red}{\tiny{$\downarrow$23\%}} & 65.08\textcolor{red}{\tiny{$\downarrow$27\%}} & 59.24\textcolor{red}{\tiny{$\downarrow$20\%}} & 65.95\textcolor{red}{\tiny{$\downarrow$24\%}} \\
     & Adv & 59.66\textcolor{red}{\tiny{$\downarrow$34\%}} & 61.12\textcolor{red}{\tiny{$\downarrow$33\%}} & 55.78\textcolor{red}{\tiny{$\downarrow$38\%}} & 66.10\textcolor{red}{\tiny{$\downarrow$29\%}} & 57.66\textcolor{red}{\tiny{$\downarrow$35\%}} & 53.97\textcolor{red}{\tiny{$\downarrow$40\%}} & 51.99\textcolor{red}{\tiny{$\downarrow$42\%}} & 59.44\textcolor{red}{\tiny{$\downarrow$34\%}} \\
\midrule
\multirow{4}{*}{\rotatebox{90}{Cars}} 
     & Clean & 64.20 & 57.62 & 65.62 & 63.61 & 43.81 & 47.10 & 59.78 & 64.12 \\
     & IAA & 19.90\textcolor{red}{\tiny{$\downarrow$69\%}} & 15.84\textcolor{red}{\tiny{$\downarrow$73\%}} & 17.54\textcolor{red}{\tiny{$\downarrow$73\%}} & 20.12\textcolor{red}{\tiny{$\downarrow$68\%}} & 11.14\textcolor{red}{\tiny{$\downarrow$75\%}} & 9.56\textcolor{red}{\tiny{$\downarrow$80\%}} & 14.95\textcolor{red}{\tiny{$\downarrow$75\%}} & 19.66\textcolor{red}{\tiny{$\downarrow$69\%}} \\
     & UAP & 30.44\textcolor{red}{\tiny{$\downarrow$53\%}} & 26.86\textcolor{red}{\tiny{$\downarrow$54\%}} & 25.52\textcolor{red}{\tiny{$\downarrow$61\%}} & 34.77\textcolor{red}{\tiny{$\downarrow$45\%}} & 15.93\textcolor{red}{\tiny{$\downarrow$64\%}} & 17.44\textcolor{red}{\tiny{$\downarrow$63\%}} & 19.95\textcolor{red}{\tiny{$\downarrow$67\%}} & 28.17\textcolor{red}{\tiny{$\downarrow$56\%}} \\
     &Adv & 24.86\textcolor{red}{\tiny{$\downarrow$61\%}} & 21.02\textcolor{red}{\tiny{$\downarrow$64\%}} & 22.13\textcolor{red}{\tiny{$\downarrow$66\%}} & 27.01\textcolor{red}{\tiny{$\downarrow$57\%}} & 13.90\textcolor{red}{\tiny{$\downarrow$68\%}} & 13.27\textcolor{red}{\tiny{$\downarrow$72\%}} & 17.93\textcolor{red}{\tiny{$\downarrow$70\%}} & 24.58\textcolor{red}{\tiny{$\downarrow$62\%}} \\
\midrule
\multirow{4}{*}{\rotatebox{90}{CIFAR 10}} 
     & Clean & 92.78 & 93.05 & 93.85 & 94.67 & 90.98 & 91.40 & 93.90 & 92.79 \\
     & IAA & 32.34\textcolor{red}{\tiny{$\downarrow$65\%}} & 31.19\textcolor{red}{\tiny{$\downarrow$66\%}} & 28.07\textcolor{red}{\tiny{$\downarrow$70\%}} & 32.85\textcolor{red}{\tiny{$\downarrow$65\%}} & 30.00\textcolor{red}{\tiny{$\downarrow$67\%}} & 31.74\textcolor{red}{\tiny{$\downarrow$65\%}} & 27.37\textcolor{red}{\tiny{$\downarrow$71\%}} & 32.45\textcolor{red}{\tiny{$\downarrow$65\%}} \\
     & UAP & 23.40\textcolor{red}{\tiny{$\downarrow$75\%}} & 22.13\textcolor{red}{\tiny{$\downarrow$76\%}} & 25.83\textcolor{red}{\tiny{$\downarrow$72\%}} & 25.63\textcolor{red}{\tiny{$\downarrow$73\%}} & 27.29\textcolor{red}{\tiny{$\downarrow$70\%}} & 18.81\textcolor{red}{\tiny{$\downarrow$79\%}} & 21.49\textcolor{red}{\tiny{$\downarrow$77\%}} & 22.63\textcolor{red}{\tiny{$\downarrow$76\%}} \\
     & Adv & 28.14\textcolor{red}{\tiny{$\downarrow$70\%}} & 26.93\textcolor{red}{\tiny{$\downarrow$71\%}} & 27.02\textcolor{red}{\tiny{$\downarrow$71\%}} & 29.46\textcolor{red}{\tiny{$\downarrow$69\%}} & 28.73\textcolor{red}{\tiny{$\downarrow$68\%}} & 25.66\textcolor{red}{\tiny{$\downarrow$72\%}} & 24.61\textcolor{red}{\tiny{$\downarrow$74\%}} & 27.83\textcolor{red}{\tiny{$\downarrow$70\%}} \\
\midrule
\multirow{4}{*}{\rotatebox{90}{CIFAR 100}} 
     & Clean & 77.86 & 78.18 & 76.67 & 80.19 & 72.97 & 73.86 & 79.41 & 77.79 \\
     & IAA & 23.34\textcolor{red}{\tiny{$\downarrow$70\%}} & 22.65\textcolor{red}{\tiny{$\downarrow$71\%}} & 20.45\textcolor{red}{\tiny{$\downarrow$74\%}} & 22.77\textcolor{red}{\tiny{$\downarrow$72\%}} & 18.36\textcolor{red}{\tiny{$\downarrow$75\%}} & 21.72\textcolor{red}{\tiny{$\downarrow$71\%}} & 19.59\textcolor{red}{\tiny{$\downarrow$75\%}} & 24.05\textcolor{red}{\tiny{$\downarrow$69\%}} \\
     & UAP & 10.93\textcolor{red}{\tiny{$\downarrow$86\%}} & 11.78\textcolor{red}{\tiny{$\downarrow$85\%}} & 12.55\textcolor{red}{\tiny{$\downarrow$84\%}} & 12.49\textcolor{red}{\tiny{$\downarrow$84\%}} & 10.60\textcolor{red}{\tiny{$\downarrow$85\%}} & 8.27\textcolor{red}{\tiny{$\downarrow$89\%}} & 9.55\textcolor{red}{\tiny{$\downarrow$88\%}} & 11.19\textcolor{red}{\tiny{$\downarrow$86\%}} \\
     & Adv & 17.50\textcolor{red}{\tiny{$\downarrow$77\%}} & 17.54\textcolor{red}{\tiny{$\downarrow$77\%}} & 16.68\textcolor{red}{\tiny{$\downarrow$79\%}} & 17.94\textcolor{red}{\tiny{$\downarrow$78\%}} & 14.71\textcolor{red}{\tiny{$\downarrow$80\%}} & 15.39\textcolor{red}{\tiny{$\downarrow$76\%}} & 14.87\textcolor{red}{\tiny{$\downarrow$75\%}} & 18.00\textcolor{red}{\tiny{$\downarrow$77\%}} \\
\midrule
\multirow{4}{*}{\rotatebox{90}{DTD}} 
     & Clean & 79.97 & 76.76 & 77.02 & 75.43 & 73.19 & 72.13 & 77.45 & 77.61 \\
     & IAA & 40.02\textcolor{red}{\tiny{$\downarrow$50\%}} & 37.65\textcolor{red}{\tiny{$\downarrow$51\%}} & 38.88\textcolor{red}{\tiny{$\downarrow$50\%}} & 40.14\textcolor{red}{\tiny{$\downarrow$50\%}} & 33.50\textcolor{red}{\tiny{$\downarrow$54\%}} & 33.86\textcolor{red}{\tiny{$\downarrow$53\%}} & 38.96\textcolor{red}{\tiny{$\downarrow$50\%}} & 41.30\textcolor{red}{\tiny{$\downarrow$47\%}} \\
     & UAP & 63.57\textcolor{red}{\tiny{$\downarrow$17\%}} & 61.54\textcolor{red}{\tiny{$\downarrow$20\%}} & 61.52\textcolor{red}{\tiny{$\downarrow$20\%}} & 62.51\textcolor{red}{\tiny{$\downarrow$17\%}} & 59.00\textcolor{red}{\tiny{$\downarrow$19\%}} & 53.20\textcolor{red}{\tiny{$\downarrow$26\%}} & 59.85\textcolor{red}{\tiny{$\downarrow$23\%}} & 64.98\textcolor{red}{\tiny{$\downarrow$16\%}} \\
     & Adv & 51.11\textcolor{red}{\tiny{$\downarrow$34\%}} & 48.90\textcolor{red}{\tiny{$\downarrow$36\%}} & 49.53\textcolor{red}{\tiny{$\downarrow$36\%}} & 50.67\textcolor{red}{\tiny{$\downarrow$33\%}} & 45.51\textcolor{red}{\tiny{$\downarrow$38\%}} & 42.96\textcolor{red}{\tiny{$\downarrow$40\%}} & 48.79\textcolor{red}{\tiny{$\downarrow$37\%}} & 52.45\textcolor{red}{\tiny{$\downarrow$32\%}} \\
\midrule
\multirow{4}{*}{\rotatebox{90}{Flowers}} 
     & Clean & 94.92 & 93.36 & 95.23 & 94.07 & 90.57 & 90.59 & 93.84 & 94.92 \\
     & IAA & 47.71\textcolor{red}{\tiny{$\downarrow$50\%}} & 43.94\textcolor{red}{\tiny{$\downarrow$53\%}} & 43.76\textcolor{red}{\tiny{$\downarrow$54\%}} & 47.25\textcolor{red}{\tiny{$\downarrow$50\%}} & 40.25\textcolor{red}{\tiny{$\downarrow$56\%}} & 34.86\textcolor{red}{\tiny{$\downarrow$62\%}} & 39.92\textcolor{red}{\tiny{$\downarrow$58\%}} & 47.94\textcolor{red}{\tiny{$\downarrow$50\%}} \\
     & UAP & 57.51\textcolor{red}{\tiny{$\downarrow$39\%}} & 59.58\textcolor{red}{\tiny{$\downarrow$36\%}} & 58.50\textcolor{red}{\tiny{$\downarrow$39\%}} & 66.71\textcolor{red}{\tiny{$\downarrow$29\%}} & 55.73\textcolor{red}{\tiny{$\downarrow$38\%}} & 46.55\textcolor{red}{\tiny{$\downarrow$49\%}} & 46.21\textcolor{red}{\tiny{$\downarrow$51\%}} & 57.29\textcolor{red}{\tiny{$\downarrow$40\%}} \\
     & Adv & 52.32\textcolor{red}{\tiny{$\downarrow$45\%}} & 51.30\textcolor{red}{\tiny{$\downarrow$45\%}} & 50.70\textcolor{red}{\tiny{$\downarrow$47\%}} & 56.41\textcolor{red}{\tiny{$\downarrow$40\%}} & 47.54\textcolor{red}{\tiny{$\downarrow$48\%}} & 40.36\textcolor{red}{\tiny{$\downarrow$55\%}} & 42.88\textcolor{red}{\tiny{$\downarrow$54\%}} & 52.34\textcolor{red}{\tiny{$\downarrow$45\%}} \\
\midrule
\multirow{4}{*}{\rotatebox{90}{Food}} 
     & Clean & 76.09 & 73.07 & 78.42 & 73.83 & 67.24 & 69.05 & 76.51 & 75.81 \\
     & IAA & 27.50\textcolor{red}{\tiny{$\downarrow$64\%}} & 24.15\textcolor{red}{\tiny{$\downarrow$67\%}} & 24.09\textcolor{red}{\tiny{$\downarrow$69\%}} & 27.69\textcolor{red}{\tiny{$\downarrow$62\%}} & 21.03\textcolor{red}{\tiny{$\downarrow$69\%}} & 19.81\textcolor{red}{\tiny{$\downarrow$71\%}} & 23.39\textcolor{red}{\tiny{$\downarrow$69\%}} & 26.37\textcolor{red}{\tiny{$\downarrow$65\%}} \\
     & UAP & 40.06\textcolor{red}{\tiny{$\downarrow$47\%}} & 37.88\textcolor{red}{\tiny{$\downarrow$48\%}} & 39.60\textcolor{red}{\tiny{$\downarrow$50\%}} & 42.42\textcolor{red}{\tiny{$\downarrow$43\%}} & 30.05\textcolor{red}{\tiny{$\downarrow$55\%}} & 25.37\textcolor{red}{\tiny{$\downarrow$63\%}} & 30.51\textcolor{red}{\tiny{$\downarrow$60\%}} & 38.54\textcolor{red}{\tiny{$\downarrow$49\%}} \\
     & Adv & 33.41\textcolor{red}{\tiny{$\downarrow$56\%}} & 30.61\textcolor{red}{\tiny{$\downarrow$58\%}} & 31.39\textcolor{red}{\tiny{$\downarrow$60\%}} & 34.62\textcolor{red}{\tiny{$\downarrow$53\%}} & 25.28\textcolor{red}{\tiny{$\downarrow$62\%}} & 22.43\textcolor{red}{\tiny{$\downarrow$68\%}} & 26.74\textcolor{red}{\tiny{$\downarrow$65\%}} & 32.10\textcolor{red}{\tiny{$\downarrow$58\%}} \\
\midrule
\multirow{4}{*}{\rotatebox{90}{Pets}} 
     & Clean & 89.13 & 89.08 & 89.15 & 90.77 & 83.23 & 92.06 & 87.47 & 89.13 \\
     & IAA & 45.87\textcolor{red}{\tiny{$\downarrow$49\%}} & 44.48\textcolor{red}{\tiny{$\downarrow$50\%}} & 39.48\textcolor{red}{\tiny{$\downarrow$56\%}} & 50.74\textcolor{red}{\tiny{$\downarrow$44\%}} & 37.75\textcolor{red}{\tiny{$\downarrow$55\%}} & 41.79\textcolor{red}{\tiny{$\downarrow$55\%}} & 36.73\textcolor{red}{\tiny{$\downarrow$58\%}} & 45.95\textcolor{red}{\tiny{$\downarrow$48\%}} \\
     & UAP & 69.99\textcolor{red}{\tiny{$\downarrow$21\%}} & 72.36\textcolor{red}{\tiny{$\downarrow$19\%}} & 67.33\textcolor{red}{\tiny{$\downarrow$24\%}} & 76.55\textcolor{red}{\tiny{$\downarrow$16\%}} & 66.09\textcolor{red}{\tiny{$\downarrow$21\%}} & 71.36\textcolor{red}{\tiny{$\downarrow$22\%}} & 64.99\textcolor{red}{\tiny{$\downarrow$26\%}} & 70.52\textcolor{red}{\tiny{$\downarrow$21\%}} \\
     & Adv & 57.22\textcolor{red}{\tiny{$\downarrow$36\%}} & 57.60\textcolor{red}{\tiny{$\downarrow$35\%}} & 52.58\textcolor{red}{\tiny{$\downarrow$41\%}} & 62.88\textcolor{red}{\tiny{$\downarrow$31\%}} & 51.09\textcolor{red}{\tiny{$\downarrow$39\%}} & 55.71\textcolor{red}{\tiny{$\downarrow$39\%}} & 50.03\textcolor{red}{\tiny{$\downarrow$43\%}} & 57.52\textcolor{red}{\tiny{$\downarrow$35\%}} \\
\midrule
\multirow{4}{*}{\rotatebox{90}{\textbf{All}}} 
     & \textbf{Clean} & 80.26 & 78.71 & 80.72 & 80.47 & 73.09 & 74.59 & 79.19 & 79.90 \\
     & \textbf{IAA} & 34.06\textcolor{red}{\tiny{$\downarrow$58\%}} & 32.09\textcolor{red}{\tiny{$\downarrow$59\%}} & 30.55\textcolor{red}{\tiny{$\downarrow$62\%}} & 35.24\textcolor{red}{\tiny{$\downarrow$56\%}} & 28.19\textcolor{red}{\tiny{$\downarrow$62\%}} & 27.47\textcolor{red}{\tiny{$\downarrow$63\%}} & 28.92\textcolor{red}{\tiny{$\downarrow$63\%}} & 34.19\textcolor{red}{\tiny{$\downarrow$57\%}} \\
     & \textbf{UAP} & 41.89\textcolor{red}{\tiny{$\downarrow$49\%}} & 41.67\textcolor{red}{\tiny{$\downarrow$49\%}} & 41.32\textcolor{red}{\tiny{$\downarrow$50\%}} & 45.98\textcolor{red}{\tiny{$\downarrow$44\%}} & 38.39\textcolor{red}{\tiny{$\downarrow$50\%}} & 35.27\textcolor{red}{\tiny{$\downarrow$55\%}} & 36.20\textcolor{red}{\tiny{$\downarrow$55\%}} & 41.72\textcolor{red}{\tiny{$\downarrow$49\%}} \\
     & \textbf{Adv} & 37.75\textcolor{red}{\tiny{$\downarrow$54\%}} & 36.60\textcolor{red}{\tiny{$\downarrow$55\%}} & 35.62\textcolor{red}{\tiny{$\downarrow$57\%}} & 40.31\textcolor{red}{\tiny{$\downarrow$51\%}} & 33.00\textcolor{red}{\tiny{$\downarrow$57\%}} & 31.15\textcolor{red}{\tiny{$\downarrow$60\%}} & 32.35\textcolor{red}{\tiny{$\downarrow$60\%}} & 37.74\textcolor{red}{\tiny{$\downarrow$54\%}} \\
\bottomrule
\end{tabular}
\end{table*}
\clearpage
\subsubsection{AirCraft}
\begin{table*}[h]
    \centering
    \scriptsize
    \caption{This table presents the results of various instance and universal adversarial perturbation (UAP) attacks on the AirCraft dataset, with all UAP attack names in \textit{italics}. Different configurations of FGSM and PGD are denoted, such as $FGSM_1$ and $PGD_1$. Average results for universal adversarial perturbations (UAP Avg.), instance adversarial attacks (IAA Avg.), and overall adversarial performance (Adv Avg.) are reported at the bottom, including percentage drops relative to clean accuracy.}
    \label{tab:few_shot_small_domain_shift}

    \begin{adjustbox}{width=\textwidth}
    \begin{tabular}{lllllllll}
    \toprule
    
    {} & Barlow & BYOL & DINO & MoCoV3 & SimCLR & Supervised & SwAV & VICReg \\
    \midrule
    
    \textbf{$FGSM_1$}& 8.92 & 5.94 & 4.84 & 11.41 & 2.7 & 2.58 & 3.64 & 8.86\\
    $FGSM_2$ & 1.52 & 0.69 & 0.45 & 1.95 & 0.78 & 0.81 & 2.57 & 1.8\\
    
    $PGD_1$ & 10.03 & 5.72 & 4.54 & 10.96 & 3.44 & 1.61 & 4 & 10.18\\
    $PGD_2$ & 0.06 & 0 & 0 & 0.12 & 0.24 & 0.18 & 0.64 & 0.06\\
    
    $PGD_3$  & 10.27 & 6.02 & 4.63 & 11.09 & 3.27 & 1.61 & 3.83 & 10.06\\
    $PGD_4$ & 0.06 & 0 & 0 & 0.12 & 0.18 & 0.12 & 0.61 & 0.06\\
    
    $PGD_5$ & 0.12 & 0.03 & 0 & 0.24 & 0.18 & 0.24 & 0.79 & 0.12\\
    
    DIFGSM & 24.56 & 24.16 & 20.83 & 28.01 & 19.39 & 19.43 & 16.74 & 27.41\\
    
    CW & 0 & 0 & 0 & 0 & 0 & 0 & 0 & 0\\
    
    Jitter & 45.87 & 44.28 & 48.39 & 45.42 & 37.43 & 31.98 & 43.75 & 44.73\\
    
    TIFGSM  & 32.78 & 31.08 & 29.68 & 35.76 & 28.31 & 18.99 & 29.83 & 33.04\\
    
    PIFGSM  & 3.62 & 2.1 & 1.62 & 4.46 & 0.9 & 0.6 & 1.71 & 3.44\\
    
    EADEN & 0 & 0 & 0 & 0 & 0 & 0 & 0 & 0\\ 
    OnePixel & 51.75 & 49.39 & 54.93 & 53.41 & 41.4 & 36.01 & 47.55 & 51.54\\
    Pixle & 3.67 & 1.9 & 2.17 & 6.16 & 2.8 & 1.48 & 2.26 & 3.8\\
    SPSA & 44.36 & 42.91 & 44.2 & 46.6 & 30.76 & 28.51 & 38.42 & 44.31\\
    Square  & 0.03 & 0 & 0 & 0.03 & 0.03 & 0 & 0 & 0.03\\
    TAP & 55.53 & 53.4 & 58.55 & 57.72 & 42.93 & 32.54 & 52.48 & 55.35\\
        \textit{ASV} & 21.31 & 25.20 & 28.18 & 39.65 & 22.87 & 20.31 & 23.38 & 21.18 \\
        \textit{FFF (mean-std)} & 17.59 & 25.94 & 23.81 & 28.10 & 17.91 & 13.76 & 20.00 & 17.40 \\
        \textit{FFF (no-data)} & 14.86 & 17.93 & 10.41 & 25.14 & 10.69 & 7.07 & 6.54 & 15.05 \\
        \textit{FFF (one-sample)} & 13.21 & 13.21 & 3.21 & 13.12 & 11.94 & 7.67 & 6.81 & 10.69 \\
        \textit{FG-UAP} & 2.36 & 1.52 & 2.14 & 1.99 & 1.55 & 1.27 & 3.32 & 2.05 \\
        \textit{GD-UAP (mean-std)} & 18.95 & 18.03 & 15.16 & 21.30 & 15.17 & 22.07 & 15.71 & 16.51 \\
        \textit{GD-UAP (no-data)} & 16.79 & 11.39 & 7.79 & 23.47 & 18.70 & 11.18 & 7.32 & 15.83 \\
        \textit{GD-UAP (one-sample)} & 3.41 & 1.90 & 3.24 & 5.61 & 1.56 & 4.48 & 6.11 & 3.48 \\
        \textit{L4A-base} & 31.12 & 24.98 & 32.35 & 37.42 & 27.29 & 15.81 & 15.24 & 31.54 \\
        \textit{L4A-fuse} & 31.47 & 25.71 & 32.00 & 38.29 & 27.28 & 15.84 & 16.27 & 31.11 \\
        \textit{L4A-ugs} & 41.55 & 36.54 & 42.27 & 40.98 & 25.42 & 26.53 & 32.97 & 41.93 \\
        \textit{PD-UAP} & 26.27 & 13.53 & 8.70 & 18.34 & 17.58 & 11.92 & 10.62 & 26.54 \\
        \textit{SSP} & 22.28 & 23.89 & 27.06 & 31.55 & 22.69 & 12.69 & 20.22 & 23.39 \\
        \textit{STD} & 20.66 & 26.74 & 26.68 & 33.18 & 12.77 & 25.20 & 23.94 & 19.70 \\
        \textit{UAP (DeepFool)} & 7.05 & 10.30 & 8.59 & 13.81 & 8.29 & 24.45 & 15.34 & 6.45 \\
        \textit{UAPEPGD} & 34.64 & 37.38 & 36.34 & 37.23 & 19.05 & 28.45 & 32.37 & 34.22 \\
    \midrule
    Clean Accuracy & 56.88 & 56.34 & 60.25  & 58.75 & 46.77 & 44.89 & 54.01 & 56.43 \\
    IAA Avg. & 16.29 \textcolor{red}{\tiny{$\downarrow$71\%}}  & 14.87\textcolor{red}{\tiny{$\downarrow$73\%}} & 15.27\textcolor{red}{\tiny{$\downarrow$75\%}}  & 17.41\textcolor{red}{\tiny{$\downarrow$70\%}} & 11.93\textcolor{red}{\tiny{$\downarrow$74\%}} & 9.82\textcolor{red}{\tiny{$\downarrow$78\%}} & 13.82\textcolor{red}{\tiny{$\downarrow$74\%}}  & 16.38\textcolor{red}{\tiny{$\downarrow$71\%}} \\
    UAP Avg. & 20.22\textcolor{red}{\tiny{$\downarrow$64\%}} & 19.64\textcolor{red}{\tiny{$\downarrow$65\%}} & 19.25\textcolor{red}{\tiny{$\downarrow$68\%}} & 25.57\textcolor{red}{\tiny{$\downarrow$56\%}} & 16.30\textcolor{red}{\tiny{$\downarrow$65\%}} & 15.54\textcolor{red}{\tiny{$\downarrow$65\%}} & 16.01\textcolor{red}{\tiny{$\downarrow$70\%}} & 19.82\textcolor{red}{\tiny{$\downarrow$64\%}} \\

    Adv Avg. & 18.14\textcolor{red}{\tiny{$\downarrow$68\%}}  & 17.11\textcolor{red}{\tiny{$\downarrow$70\%}} & 17.14\textcolor{red}{\tiny{$\downarrow$72\%}}  & 21.25\textcolor{red}{\tiny{$\downarrow$64\%}} & 13.99\textcolor{red}{\tiny{$\downarrow$70\%}} & 12.51\textcolor{red}{\tiny{$\downarrow$72\%}} & 14.85\textcolor{red}{\tiny{$\downarrow$73\%}}  & 18.00\textcolor{red}{\tiny{$\downarrow$68\%}} \\

    \midrule
    \end{tabular}
    \end{adjustbox}
\end{table*}

\clearpage
\subsubsection{Caltech 101}
\begin{table*}[h]
    \centering
    \scriptsize
    \caption{This table presents the results of various instance and universal adversarial perturbation (UAP) attacks on the Caltech 101 dataset, with all UAP attack names in \textit{italics}. Different configurations of FGSM and PGD are denoted, such as $FGSM_1$ and $PGD_1$. Average results for universal adversarial perturbations (UAP Avg.), instance adversarial attacks (IAA Avg.), and overall adversarial performance (Adv Avg.) are reported at the bottom, including percentage drops relative to clean accuracy.}
    \label{tab:few_shot_small_domain_shift}

    \begin{adjustbox}{width=\textwidth}
    \begin{tabular}{lllllllll}
    \toprule
    {} & Barlow & BYOL & DINO & MoCoV3 & SimCLR & Supervised & SwAV & VICReg \\
    \midrule
    
    $FGSM_1$ & 75.31 & 75.58 & 66.93 & 79.84 & 66.06 & 62.11 & 63.12 & 75.3\\
    $FGSM_2$ & 53.82 & 52.44 & 37.84 & 59.58 & 47.67 & 27.38 & 36.13 & 53.82\\
    
    $PGD_1$ & 74.27 & 75.19 & 65.57 & 79.35 & 64.94 & 58.96 & 61.96 & 74.34\\
    $PGD_2$  & 9.61 & 10.47 & 2.24 & 17.17 & 11.14 & 1.64 & 2.05 & 9.34\\
    
    $PGD_3$  & 74.43 & 75.39 & 65.7 & 79.81 & 65 & 59.24 & 62.28 & 74.68\\
    $PGD_4$  & 7.62 & 9 & 1.81 & 14.79 & 10.22 & 1.19 & 1.69 & 7.53\\
    
    $PGD_5$  & 17.17 & 18.64 & 5.48 & 25.45 & 13.11 & 4.35 & 3.91 & 16.86\\
    
    DIFGSM  & 80.24 & 81.09 & 76.38 & 83.66 & 76.16 & 71.28 & 75.23 & 79.97\\
    
    CW & 0.68 & 0.94 & 0.3 & 0.79 & 0.49 & 0.22 & 0.31 & 0.68\\
    
    Jitter & 83.43 & 83.41 & 81.7 & 86.82 & 80.89 & 77.36 & 79.34 & 83.85\\
    
    TIFGSM & 85.73 & 86.72 & 83.63 & 88.69 & 82.73 & 79.58 & 81.98 & 85.98\\
    
    PIFGSM  & 68.03 & 68.03 & 53.66 & 74.14 & 50.54 & 49 & 45.82 & 67.98\\
    
    EADEN & 0 & 0 & 0 & 0 & 0 & 0 & 0 & 0\\
    OnePixel & 89.85 & 90.57 & 89.43 & 92.25 & 87.67 & 88.7 & 89.52 & 89.88\\
    Pixle & 53.89 & 57.26 & 40.6 & 67.39 & 49.57 & 39.02 & 32.73 & 54.58\\
    SPSA & 88.89 & 88.82 & 87.45 & 91.08 & 86.51 & 85.73 & 87.2 & 89.04\\
    Square & 11.43 & 8.71 & 4.7 & 14.98 & 15.14 & 1.03 & 6.53 & 11.37\\
    TAP & 90.48 & 90.91 & 90.16 & 92.36 & 88.52 & 87.13 & 90.12 & 90.48\\
    \textit{ASV} & 87.36 & 88.54 & 87.05 & 91.07 & 86.34 & 84.57 & 86.72 & 87.61 \\
    \textit{FFF (mean-std)} & 83.37 & 79.83 & 83.83 & 88.14 & 81.32 & 72.29 & 80.73 & 83.66 \\
    \textit{FFF (no-data)} & 73.92 & 51.35 & 58.54 & 77.23 & 75.82 & 55.89 & 55.47 & 71.03 \\
    \textit{FFF (one-sample)} & 42.23 & 51.33 & 65.30 & 79.74 & 44.95 & 65.27 & 65.96 & 37.53 \\
    \textit{FG-UAP} & 8.78 & 6.05 & 10.58 & 15.68 & 17.41 & 5.23 & 8.04 & 9.26 \\
    \textit{GD-UAP (mean-std)} & 83.06 & 81.74 & 65.43 & 78.92 & 80.73 & 78.18 & 72.84 & 81.71 \\
    \textit{GD-UAP (no-data)} & 58.14 & 62.99 & 51.64 & 73.74 & 73.92 & 58.89 & 62.15 & 59.71 \\
    \textit{GD-UAP (one-sample)} & 33.34 & 68.44 & 18.83 & 37.63 & 27.16 & 71.42 & 43.55 & 38.20 \\
    \textit{L4A-base} & 70.40 & 79.34 & 84.44 & 86.36 & 63.87 & 66.12 & 27.52 & 70.06 \\
    \textit{L4A-fuse} & 69.97 & 78.93 & 84.56 & 85.72 & 63.84 & 64.96 & 27.34 & 69.21 \\
    \textit{L4A-ugs} & 88.72 & 88.78 & 86.75 & 92.23 & 83.99 & 82.42 & 85.82 & 88.82 \\
    \textit{PD-UAP} & 88.58 & 75.06 & 73.89 & 86.96 & 84.04 & 71.07 & 75.22 & 88.18 \\
    \textit{SSP} & 73.96 & 82.74 & 78.66 & 85.34 & 80.62 & 57.39 & 64.53 & 71.47 \\
    \textit{STD} & 88.03 & 89.12 & 84.49 & 90.76 & 84.99 & 86.23 & 85.52 & 88.12 \\
    \textit{UAP (DeepFool)} & 24.67 & 30.67 & 21.12 & 37.96 & 29.01 & 33.89 & 18.49 & 21.36 \\
    \textit{UAPEPGD} & 89.13 & 90.02 & 87.70 & 91.76 & 86.07 & 87.37 & 87.94 & 89.29 \\
    \hline
    Clean Accuracy & 90.54 & 90.99 & 90.31  & 92.89 & 89.1 & 90.25 & 90.36 & 90.57 \\
    IAA Avg. & 53.60 \textcolor{red}{\tiny{$\downarrow$41\%}}& 54.06\textcolor{red}{\tiny{$\downarrow$41\%}} & 47.42\textcolor{red}{\tiny{$\downarrow$47\%}} & 58.23\textcolor{red}{\tiny{$\downarrow$37\%}} & 49.79\textcolor{red}{\tiny{$\downarrow$44\%}} & 44.10\textcolor{red}{\tiny{$\downarrow$51\%}} & 45.55\textcolor{red}{\tiny{$\downarrow$50\%}} & 53.64\textcolor{red}{\tiny{$\downarrow$41\%}} \\
    UAP Avg. & 66.48 \textcolor{red}{\tiny{$\downarrow$26.58\%}}  & 69.06\textcolor{red}{\tiny{$\downarrow$15.8\%}} & 65.18\textcolor{red}{\tiny{$\downarrow$27.9\%}}  & 74.95\textcolor{red}{\tiny{$\downarrow$7.4\%}} & 66.50\textcolor{red}{\tiny{$\downarrow$23.2\%}} & 65.08\textcolor{red}{\tiny{$\downarrow$26.8\%}} & 59.24\textcolor{red}{\tiny{$\downarrow$20.3\%}}  & 65.95\textcolor{red}{\tiny{$\downarrow$24.2\%}} \\

    Adv Avg. & 59.66 \textcolor{red}{\tiny{$\downarrow$34.1\%}}  & 61.12\textcolor{red}{\tiny{$\downarrow$32.8\%}} & 55.78\textcolor{red}{\tiny{$\downarrow$38.2\%}}  & 66.10\textcolor{red}{\tiny{$\downarrow$28.8\%}} & 57.66\textcolor{red}{\tiny{$\downarrow$35.3\%}} & 53.97\textcolor{red}{\tiny{$\downarrow$40.2\%}} & 51.99\textcolor{red}{\tiny{$\downarrow$42.5\%}}  & 59.44\textcolor{red}{\tiny{$\downarrow$34.4\%}} \\

    \midrule
    \end{tabular}
    \end{adjustbox}
\end{table*}
\clearpage
\subsubsection{Cars}

\begin{table*}[h]
    \centering
    \scriptsize
    \caption{This table presents the results of various instance and universal adversarial perturbation (UAP) attacks on the Cars dataset, with all UAP attack names in \textit{italics}. Different configurations of FGSM and PGD are denoted, such as $FGSM_1$ and $PGD_1$. Average results for universal adversarial perturbations (UAP Avg.), instance adversarial attacks (IAA Avg.), and overall adversarial performance (Adv Avg.) are reported at the bottom, including percentage drops relative to clean accuracy.}
    \label{tab:few_shot_small_domain_shift}

    \begin{adjustbox}{width=\textwidth}
    \begin{tabular}{lllllllll}
    \toprule
    {} & Barlow & BYOL & DINO & MoCoV3 & SimCLR & Supervised & SwAV & VICReg \\
    \midrule

    $FGSM_1$ & 14.55 & 8.27 & 6.34 & 16.32 & 3.18 & 2.1 & 3.48 & 14.48 \\
    $FGSM_2$ & 1.41 & 0.6 & 0.51 & 1.39 & 0.9 & 0.16 & 0.5 & 1.42\\
    
    $PGD_1$ & 14.15 & 7.76 & 5.83 & 15.3 & 3.42 & 1.6 & 3.03 & 13.94\\
    $PGD_2$ & 0.02 & 0 & 0 & 0 & 0.19 & 0.09 & 0 & 0.02\\
    
    $PGD_3$ & 14.3 & 8.05 & 5.83 & 15.5 & 3.52 & 1.67 & 3.11 & 14.33\\
    $PGD_4$ & 0.01 & 0 & 0 & 0 & 0.17 & 0.07 & 0 & 0.01\\
    
    $PGD_5$ & 0 & 0.01 & 0 & 0 & 0.19 & 0 & 0.02 & 0\\
    
    DIFGSM & 33.68 & 24.49 & 28.83 & 32.76 & 17.55 & 14.05 & 22.04 & 30.92\\
    
    CW & 0 & 0 & 0 & 0 & 0 & 0 & 0 & 0\\
    
    Jitter & 44.21 & 36.41 & 45.44 & 42.21 & 26.4 & 23.85 & 40.67 & 43.86\\
    
    TIFGSM   & 44.26 & 35.39 & 39.8 & 43.84 & 27.62 & 22.01 & 34.77 & 44\\
    
    PIFGSM & 6.32 & 3.3 & 1.54 & 8.54 & 0.75 & 0.6 & 0.65 & 6.39\\
    
    EADEN & 0 & 0 & 0 & 0 & 0 & 0 & 0 & 0\\
    OnePixel & 60.73 & 53.74 & 61.77 & 59.99 & 39.56 & 39.56 & 55.33 & 60.63\\
    Pixle & 6.63 & 5.1 & 5.02 & 8.17 & 4.14 & 1.92 & 2.3 & 6.33\\
    SPSA & 54.22 & 45.44 & 51.11 & 54.88 & 30.33 & 31.59 & 44.47 & 54.05\\
    Square & 0.06 & 0.01 & 0 & 0.04 & 0 & 0 & 0.01 & 0.05\\
    TAP  & 63.71 & 56.62 & 63.75 & 63.26 & 42.74 & 32.92 & 58.79 & 63.51\\
  \textit{ASV} & 43.05 & 35.78 & 42.81 & 51.73 & 28.26 & 27.47 & 33.37 & 41.80 \\
    \textit{FFF (mean-std)} & 35.13 & 34.85 & 34.66 & 38.75 & 22.62 & 19.05 & 26.25 & 34.98 \\
    \textit{FFF (no-data)} & 23.84 & 15.97 & 12.54 & 29.39 & 18.54 & 15.92 & 10.19 & 17.47 \\
    \textit{FFF (one-sample)} & 17.76 & 22.51 & 19.75 & 38.42 & 12.64 & 16.70 & 29.70 & 19.15 \\
    \textit{FG-UAP} & 2.69 & 0.82 & 1.78 & 1.04 & 2.18 & 0.92 & 1.27 & 2.90 \\
    \textit{GD-UAP (mean-std)} & 41.11 & 29.44 & 26.35 & 39.20 & 23.24 & 24.42 & 27.16 & 40.33 \\
    \textit{GD-UAP (no-data)} & 17.31 & 20.43 & 15.52 & 27.40 & 15.31 & 13.08 & 19.67 & 18.77 \\
    \textit{GD-UAP (one-sample)} & 5.92 & 17.22 & 4.99 & 12.35 & 3.51 & 5.88 & 16.04 & 6.47 \\
    \textit{L4A-base} & 33.94 & 28.37 & 34.16 & 37.86 & 9.80 & 8.62 & 6.39 & 33.17 \\
    \textit{L4A-fuse} & 33.29 & 28.72 & 33.13 & 38.13 & 8.72 & 8.68 & 6.72 & 33.03 \\
    \textit{L4A-ugs} & 56.10 & 38.76 & 47.90 & 55.12 & 29.59 & 21.00 & 34.34 & 55.49 \\
    \textit{PD-UAP} & 48.19 & 26.53 & 22.61 & 34.37 & 23.79 & 17.17 & 12.26 & 47.34 \\
    \textit{SSP} & 21.66 & 33.84 & 31.08 & 38.27 & 17.26 & 8.03 & 19.59 & 24.26 \\
    \textit{STD} & 37.64 & 31.02 & 34.57 & 39.36 & 16.42 & 24.72 & 28.23 & 36.97 \\
    \textit{UAP (DeepFool)} & 16.91 & 20.25 & 25.59 & 20.54 & 14.21 & 32.35 & 28.93 & 17.85 \\
    \textit{UAPEPGD} & 52.56 & 45.27 & 49.12 & 54.36 & 26.15 & 34.96 & 40.44 & 51.96 \\
    
    \hline
    Clean Accuracy & 64.20 & 57.62 & 65.62  & 63.61 & 43.81 & 47.10 & 59.78 & 64.12 \\
    IAA Avg. & 19.90 \textcolor{red}{\tiny{$\downarrow$69\%}}& 15.84\textcolor{red}{\tiny{$\downarrow$73\%}} & 17.54\textcolor{red}{\tiny{$\downarrow$73\%}} & 20.12\textcolor{red}{\tiny{$\downarrow$68\%}} & 11.14\textcolor{red}{\tiny{$\downarrow$75\%}} & 9.56\textcolor{red}{\tiny{$\downarrow$80\%}} & 14.95\textcolor{red}{\tiny{$\downarrow$75\%}} & 19.66\textcolor{red}{\tiny{$\downarrow$69\%}} \\

    UAP & 30.44\textcolor{red}{\tiny{$\downarrow$53\%}} & 26.86\textcolor{red}{\tiny{$\downarrow$54\%}} & 25.52\textcolor{red}{\tiny{$\downarrow$61\%}} & 34.77\textcolor{red}{\tiny{$\downarrow$45\%}} & 15.93\textcolor{red}{\tiny{$\downarrow$64\%}} & 17.44\textcolor{red}{\tiny{$\downarrow$63\%}} & 19.95\textcolor{red}{\tiny{$\downarrow$67\%}} & 28.17\textcolor{red}{\tiny{$\downarrow$56\%}} \\

    Adv Avg. & 24.86\textcolor{red}{\tiny{$\downarrow$61\%}} & 21.02\textcolor{red}{\tiny{$\downarrow$64\%}} & 22.13\textcolor{red}{\tiny{$\downarrow$66\%}} & 27.01\textcolor{red}{\tiny{$\downarrow$57\%}} & 13.90\textcolor{red}{\tiny{$\downarrow$68\%}} & 13.27\textcolor{red}{\tiny{$\downarrow$72\%}} & 17.93\textcolor{red}{\tiny{$\downarrow$70\%}} & 24.58\textcolor{red}{\tiny{$\downarrow$62\%}} \\

    \midrule
    \end{tabular}
    \end{adjustbox}
\end{table*}

\clearpage
\subsubsection{CIFAR 10}
\begin{table*}[h]
    \centering
    \scriptsize
    \caption{This table presents the results of various instance and universal adversarial perturbation (UAP) attacks on the CIFAR 10 dataset, with all UAP attack names in \textit{italics}. Different configurations of FGSM and PGD are denoted, such as $FGSM_1$ and $PGD_1$. Average results for universal adversarial perturbations (UAP Avg.), instance adversarial attacks (IAA Avg.), and overall adversarial performance (Adv Avg.) are reported at the bottom, including percentage drops relative to clean accuracy.}
    \label{tab:few_shot_small_domain_shift}

    \begin{adjustbox}{width=\textwidth}
    \begin{tabular}{lllllllll}
    \toprule
    {} & Barlow & BYOL & DINO & MoCoV3 & SimCLR & Supervised & SwAV & VICReg \\
    \midrule
    
    $FGSM_1$ & 32.95 & 31.04 & 27.57 & 33.04 & 37.86 & 42.84 & 19.38 & 33.04\\
    $FGSM_2$ & 53.83 & 50.24 & 52.58 & 52.51 & 59.88 & 29.71 & 47.54 & 53.94\\
    
    $PGD_1$ & 34.76 & 29.2 & 22.25 & 35.16 & 23.51 & 36.92 & 21.04 & 34.64\\
    $PGD_2$ & 0.02 & 0 & 0 & 0 & 0.03 & 0 & 0 & 0.01\\
    
    $PGD_3$ & 34.02 & 28.38 & 20.85 & 34.44 & 22.48 & 36.51 & 20.71 & 34.23\\
    $PGD_4$ & 0.02 & 0.02 & 0 & 0 & 0.03 & 0 & 0 & 0\\
    
    $PGD_5$ & 0 & 0 & 0 & 0 & 0.01 & 0 & 0 & 0\\
    
    DIFGSM  & 56.24 & 52.78 & 42.53 & 55.48 & 52.39 & 55.9 & 39.2 & 54.64\\
    
    CW & 0 & 0 & 0 & 0 & 0.06 & 0 & 0 & 0\\
    
    Jitter & 66.67 & 62.37 & 59.8 & 66.97 & 55.15 & 70.7 & 58.5 & 67.63\\
    
    TIFGSM & 52.32 & 48.88 & 41.23 & 50.64 & 56.11 & 56.88 & 42.38 & 54.51\\
    
    PIFGSM  & 0.39 & 0.22 & 0.04 & 0.28 & 0.45 & 5.18 & 0 & 0.41\\
    
    EADEN & 0 & 0 & 0 & 0 & 0 & 0 & 0 & 0\\
    OnePixel & 87.36 & 86.09 & 88.42 & 87.78 & 82.28 & 85.59 & 81.11 & 87.21\\
    Pixle & 5.55 & 2.15 & 4.44 & 3.02 & 1.82 & 2.22 & 1.93 & 5.41\\
    SPSA & 69.6 & 79.09 & 55.69 & 80.73 & 60.51 & 71.34 & 68.81 & 69.9\\
    Square & 0 & 0 & 0.05 & 0 & 0 & 0 & 0 & 0\\
    TAP & 88.51 & 91.06 & 89.82 & 91.4 & 87.56 & 77.66 & 92.14 & 88.59\\
    \textit{ASV} & 50.49 & 71.15 & 66.54 & 67.20 & 51.96 & 36.61 & 70.25 & 50.81 \\
    \textit{FFF (mean-std)} & 22.27 & 23.90 & 30.37 & 30.11 & 30.74 & 13.79 & 26.08 & 22.83 \\
    \textit{FFF (no-data)} & 14.30 & 10.41 & 12.37 & 19.61 & 31.12 & 13.33 & 10.83 & 12.70 \\
    \textit{FFF (one-sample)} & 10.00 & 10.00 & 9.99 & 10.00 & 10.00 & 11.15 & 10.03 & 10.05 \\
    \textit{FG-UAP} & 10.16 & 6.68 & 11.38 & 9.99 & 8.25 & 10.01 & 10.00 & 10.06 \\
    \textit{GD-UAP (mean-std)} & 33.92 & 18.14 & 22.02 & 17.21 & 28.83 & 19.01 & 17.66 & 26.04 \\
    \textit{GD-UAP (no-data)} & 13.37 & 10.56 & 12.09 & 10.58 & 10.51 & 12.33 & 10.28 & 11.49 \\
    \textit{GD-UAP (one-sample)} & 10.00 & 11.94 & 10.00 & 10.00 & 10.00 & 9.79 & 10.22 & 10.00 \\
    \textit{L4A-base} & 10.12 & 10.62 & 26.26 & 12.13 & 10.08 & 13.28 & 10.97 & 10.16 \\
    \textit{L4A-fuse} & 10.12 & 10.51 & 28.99 & 12.59 & 10.01 & 13.29 & 11.16 & 10.09 \\
    \textit{L4A-ugs} & 12.89 & 16.29 & 50.54 & 41.20 & 50.43 & 14.48 & 10.19 & 12.90 \\
    \textit{PD-UAP} & 67.51 & 16.31 & 14.67 & 12.94 & 49.02 & 14.89 & 40.65 & 66.96 \\
    \textit{SSP} & 12.66 & 10.13 & 10.01 & 41.45 & 15.03 & 10.42 & 9.92 & 11.01 \\
    \textit{STD} & 23.12 & 41.06 & 27.91 & 30.18 & 52.46 & 27.27 & 20.45 & 24.00 \\
    \textit{UAP (DeepFool)} & 8.93 & 13.05 & 16.73 & 9.92 & 12.51 & 18.11 & 11.23 & 9.14 \\
    \textit{UAPEPGD} & 64.61 & 73.37 & 63.41 & 74.96 & 55.63 & 63.25 & 63.91 & 63.76 \\

    \hline
    Clean Accuracy & 92.78 & 93.05 & 93.85  & 94.67 & 90.98 & 91.4 & 93.9 & 92.79 \\
    IAA Avg. & 32.34 \textcolor{red}{\tiny{$\downarrow$65\%}}& 31.19\textcolor{red}{\tiny{$\downarrow$66\%}} & 28.07\textcolor{red}{\tiny{$\downarrow$70\%}} & 32.85\textcolor{red}{\tiny{$\downarrow$65\%}} & 30.00\textcolor{red}{\tiny{$\downarrow$67\%}} & 31.74\textcolor{red}{\tiny{$\downarrow$65\%}} & 27.37\textcolor{red}{\tiny{$\downarrow$71\%}} & 32.45\textcolor{red}{\tiny{$\downarrow$65\%}} \\
    UAP Avg. & 23.40\textcolor{red}{\tiny{$\downarrow$74.77}} & 22.13\textcolor{red}{\tiny{$\downarrow$76.21}} & 25.83\textcolor{red}{\tiny{$\downarrow$72.48}} & 25.63\textcolor{red}{\tiny{$\downarrow$72.93}} & 27.29\textcolor{red}{\tiny{$\downarrow$70.01}} & 18.81\textcolor{red}{\tiny{$\downarrow$79.42}} & 21.49\textcolor{red}{\tiny{$\downarrow$77.11}} & 22.63\textcolor{red}{\tiny{$\downarrow$75.62}} \\

    Adv Avg. & 28.14\textcolor{red}{\tiny{$\downarrow$70\%}} & 26.93\textcolor{red}{\tiny{$\downarrow$71\%}} & 27.02\textcolor{red}{\tiny{$\downarrow$71\%}} & 29.46\textcolor{red}{\tiny{$\downarrow$69\%}} & 28.73\textcolor{red}{\tiny{$\downarrow$68\%}} & 25.66\textcolor{red}{\tiny{$\downarrow$72\%}} & 24.61\textcolor{red}{\tiny{$\downarrow$74\%}} & 27.83\textcolor{red}{\tiny{$\downarrow$70\%}} \\

    \midrule
    \end{tabular}
    \end{adjustbox}
\end{table*}
\clearpage
\subsubsection{CIFAR 100}

\begin{table*}[h]
    \centering
    \scriptsize
    \caption{This table presents the results of various instance and universal adversarial perturbation (UAP) attacks on the CIFAR 100 dataset, with all UAP attack names in \textit{italics}. Different configurations of FGSM and PGD are denoted, such as $FGSM_1$ and $PGD_1$. Average results for universal adversarial perturbations (UAP Avg.), instance adversarial attacks (IAA Avg.), and overall adversarial performance (Adv Avg.) are reported at the bottom, including percentage drops relative to clean accuracy.}
    \label{tab:new_few_shot_small_domain_shift}

    \begin{adjustbox}{width=\textwidth}
    \begin{tabular}{lllllllll}
    \toprule
    \
    {} & Barlow & BYOL & DINO & MoCoV3 & SimCLR & Supervised & SwAV & VICReg \\
    \midrule
    $FGSM_1$ & 20.52 & 19.01 & 16.03 & 19.29 & 19.49 & 24.51 & 11.07 & 22.34\\
    $FGSM_2$  & 34.07 & 31.02 & 34.08 & 28.84 & 30.06 & 18.20 & 29.16 & 35.71\\

    $PGD_1$ & 19.74 & 14.42 & 11.47 & 18.38 & 8.92 & 19.09 & 10.29 & 20.98\\
    $PGD_2$ & 0.04 & 0 & 0 & 0.02 & 0.12 & 0 & 0.01 & 0.06\\
    
    $PGD_3$) & 19.33 & 14.18 & 11.09 & 17.69 & 8.24 & 18.85 & 9.92 & 20.67\\
    $PGD_4$ & 0.06 & 0.01 & 0 & 0.01 & 0.08 & 0 & 0 & 0.02\\
    
    $PGD_5$ & 0 & 0 & 0 & 0 & 0.18 & 0.01 & 0 & 0\\
    
    DIFGSM  & 38.20 & 35.26 & 27.54 & 32.23 & 32.56 & 34.97 & 26.31 & 39.47\\
    
    CW & 0.01 & 0 & 0 & 0.06 & 0.02 & 0.02 & 0 & 0.04\\
    
    Jitter & 66.85 & 62.15 & 59.33 & 65.89 & 42.01 & 67.10 & 53.82 & 66.73\\
    
    TIFGSM  & 34.84 & 36.35 & 27.80 & 30.79 & 35.15 & 36.82 & 29.15 & 37.30\\
    
    PIFGSM  & 0.78 & 0.34 & 0.17 & 0.58 & 0.36 & 3.29 & 0.09 & 1.10\\
    
    EADEN & 0 & 0 & 0 & 0 & 0 & 0 & 0 & 0\\
    OnePixel & 67.73 & 66.25 & 69.87 & 67.64 & 58.73 & 64.76 & 61.41 & 68.19\\
    Pixle & 0.48 & 0.96 & 0.56 & 0.90 & 0.96 & 1.40 & 0.43 & 0.55\\
    SPSA & 47.25 & 54.96 & 38.62 & 53.70 & 28.82 & 48.30 & 44.86 & 49.46\\
    Square & 0.06 & 0.01 & 0.04 & 0 & 0 & 0 & 0.01 & 0.05\\
    TAP & 70.29 & 72.88 & 69.76 & 73.97 & 65.1 & 53.57 & 76.14 & 70.29\\
    \textit{ASV} & 34.24 & 49.85 & 45.83 & 39.93 & 26.05 & 29.08 & 42.43 & 36.42 \\
    \textit{FFF (mean-std)} & 10.69 & 15.61 & 9.80 & 9.78 & 14.54 & 3.56 & 15.74 & 11.36 \\
    \textit{FFF (no-data)} & 2.50 & 1.02 & 3.68 & 3.11 & 2.47 & 3.00 & 1.60 & 1.61 \\
    \textit{FFF (one-sample)} & 1.03 & 1.02 & 1.02 & 0.90 & 1.30 & 1.21 & 1.11 & 1.00 \\
    \textit{FG-UAP} & 1.00 & 1.06 & 0.76 & 0.96 & 1.01 & 1.05 & 1.00 & 1.00 \\
    \textit{GD-UAP (mean-std)} & 9.68 & 13.00 & 6.89 & 5.14 & 11.58 & 8.36 & 7.28 & 9.36 \\
    \textit{GD-UAP (no-data)} & 1.05 & 2.15 & 1.11 & 1.40 & 4.18 & 2.51 & 1.11 & 1.06 \\
    \textit{GD-UAP (one-sample)} & 1.97 & 1.18 & 1.00 & 1.29 & 1.28 & 1.03 & 1.94 & 1.41 \\
    \textit{L4A-base} & 1.79 & 3.18 & 17.08 & 5.35 & 1.09 & 4.42 & 0.75 & 1.68 \\
    \textit{L4A-fuse} & 1.64 & 3.00 & 17.30 & 5.23 & 1.20 & 4.16 & 0.81 & 1.66 \\
    \textit{L4A-ugs} & 4.31 & 11.44 & 27.63 & 23.33 & 28.02 & 11.62 & 2.36 & 4.77 \\
    \textit{PD-UAP} & 47.61 & 6.95 & 6.56 & 10.54 & 22.54 & 2.65 & 20.47 & 48.33 \\
    \textit{SSP} & 1.13 & 1.69 & 4.49 & 29.29 & 6.20 & 1.75 & 1.08 & 1.16 \\
    \textit{STD} & 10.57 & 24.42 & 11.34 & 14.05 & 19.14 & 13.12 & 8.17 & 10.82 \\
    \textit{UAP (DeepFool)} & 1.72 & 2.44 & 3.88 & 4.03 & 2.57 & 2.65 & 4.80 & 1.69 \\
    \textit{UAPEPGD} & 43.92 & 50.47 & 42.35 & 45.49 & 26.42 & 42.15 & 42.14 & 45.77 \\

    \hline
    Clean Accuracy & 77.86 & 78.18 & 76.67 & 80.19 & 72.97 & 73.86 & 79.41 & 77.79 \\
    IAA Avg. & 23.34 \textcolor{red}{\tiny{$\downarrow$70\%}}& 22.65\textcolor{red}{\tiny{$\downarrow$71\%}} & 20.45\textcolor{red}{\tiny{$\downarrow$74\%}} & 22.77\textcolor{red}{\tiny{$\downarrow$72\%}} & 18.36\textcolor{red}{\tiny{$\downarrow$75\%}} & 21.72\textcolor{red}{\tiny{$\downarrow$71\%}} & 19.59\textcolor{red}{\tiny{$\downarrow$75\%}} & 24.05\textcolor{red}{\tiny{$\downarrow$69\%}} \\
   UAP Avg. & 10.93 \textcolor{red}{\tiny{$\downarrow$86\%}} & 11.78 \textcolor{red}{\tiny{$\downarrow$85\%}} & 12.55 \textcolor{red}{\tiny{$\downarrow$84\%}} & 12.49 \textcolor{red}{\tiny{$\downarrow$84\%}} & 10.60 \textcolor{red}{\tiny{$\downarrow$85\%}} & 8.27 \textcolor{red}{\tiny{$\downarrow$89\%}} & 9.55 \textcolor{red}{\tiny{$\downarrow$88\%}} & 11.19 \textcolor{red}{\tiny{$\downarrow$86\%}} \\

    Adv Avg. & 17.50 \textcolor{red}{\tiny{$\downarrow$77\%}} & 17.54 \textcolor{red}{\tiny{$\downarrow$77\%}} & 16.68 \textcolor{red}{\tiny{$\downarrow$79\%}} & 17.94 \textcolor{red}{\tiny{$\downarrow$78\%}} & 14.71 \textcolor{red}{\tiny{$\downarrow$80\%}} & 15.39 \textcolor{red}{\tiny{$\downarrow$76\%}} & 14.87 \textcolor{red}{\tiny{$\downarrow$75\%}} & 18.00 \textcolor{red}{\tiny{$\downarrow$77\%}} \\

    \midrule
    \end{tabular}
    \end{adjustbox}
\end{table*}

\clearpage
\subsubsection{DTD}
\begin{table*}[h]
    \centering
    \scriptsize
    \caption{This table presents the results of various instance and universal adversarial perturbation (UAP) attacks on the DTD dataset, with all UAP attack names in \textit{italics}. Different configurations of FGSM and PGD are denoted, such as $FGSM_1$ and $PGD_1$. Average results for universal adversarial perturbations (UAP Avg.), instance adversarial attacks (IAA Avg.), and overall adversarial performance (Adv Avg.) are reported at the bottom, including percentage drops relative to clean accuracy.}
    \label{tab:few_shot_domain_shift}

    \begin{adjustbox}{width=\textwidth}
    \begin{tabular}{lllllllll}
    \toprule
    {} & Barlow & BYOL & DINO & MoCoV3 & SimCLR & Supervised & SwAV & VICReg \\
    \midrule
   
    $FGSM_1$ & 50.43 & 46.76 & 48.88 & 51.65 & 38.4 & 42.02 & 48.99 & 52.71\\
    $FGSM_2$ & 23.24 & 21.28 & 23.94 & 24.63 & 17.87 & 17.66 & 25.80 & 26.54\\
    
    $PGD_1$ & 50.05 & 46.01 & 47.93 & 51.17 & 39.31 & 40.05 & 48.99 & 51.65\\
    $PGD_2$ & 6.91 & 4.57 & 3.46 & 6.38 & 2.13 & 3.19 & 3.35 & 6.91\\
    
    $PGD_3$ & 50.11 & 46.54 & 48.14 & 51.17 & 39.04 & 40.27 & 48.62 & 51.65\\
    $PGD_4$ & 6.54 & 3.94 & 2.82 & 5.96 & 1.7 & 2.93 & 3.03 & 6.60\\
    
    $PGD_5$ & 14.89 & 12.23 & 11.81 & 16.76 & 3.99 & 10.37 & 10.53 & 16.22\\
    
    DIFGSM & 59.84 & 52.87 & 60.05 & 59.79 & 52.02 & 54.47 & 60.27 & 64.20\\
    
    CW & 0.32 & 0.32 & 0.74 & 0.69 & 0.43 & 0.64 & 0.90 & 0.90\\
    
    Jitter & 67.39 & 65.90 & 66.91 & 66.17 & 62.02 & 60.48 & 68.51 & 68.30\\
    
    TIFGSM  & 67.77 & 65.32 & 67.93 & 66.06 & 62.07 & 62.34 & 67.39 & 68.88\\
    
    PIFGSM & 42.77 & 38.83 & 40.16 & 45.53 & 26.76 & 35.43 & 38.40 & 43.94\\
    
    EADEN & 0 & 0 & 0 & 0 & 0 & 0 & 0 & 0\\
    OnePixel & 75.32 & 75.43 & 76.17 & 74.41 & 71.12 & 70.69 & 75.96 & 76.28\\
    Pixle & 49.89 & 46.28 & 46.97 & 49.57 & 40.48 & 37.62 & 41.38 & 50.90\\
    SPSA & 72.87 & 71.81 & 73.51 & 72.39 & 67.98 & 66.91 & 73.78 & 74.15\\
    Square & 8.09 & 5.96 & 6.7 & 7.77 & 5.74 & 1.49 & 8.46 & 8.67\\
    TAP & 74.10 & 73.78 & 73.72 & 72.50 & 72.07 & 62.98 & 76.97 & 75.05\\
   \textit{ASV} & 71.01 & 70.59 & 72.18 & 68.83 & 67.39 & 62.98 & 70.69 & 72.07 \\
    \textit{FFF (mean-std)} & 62.93 & 59.15 & 65.27 & 61.70 & 61.65 & 50.85 & 65.90 & 64.20 \\
    \textit{FFF (no-data)} & 58.62 & 54.15 & 55.32 & 56.33 & 62.61 & 43.99 & 49.73 & 60.59 \\
    \textit{FFF (one-sample)} & 55.32 & 53.88 & 49.36 & 57.23 & 40.69 & 37.02 & 38.62 & 57.02 \\
    \textit{FG-UAP} & 21.70 & 20.90 & 19.26 & 26.86 & 23.14 & 19.95 & 19.15 & 23.83 \\
    \textit{GD-UAP (mean-std)} & 62.13 & 58.46 & 59.41 & 57.39 & 61.97 & 54.89 & 61.38 & 63.94 \\
    \textit{GD-UAP (no-data)} & 53.62 & 55.27 & 52.18 & 57.18 & 54.04 & 44.89 & 53.62 & 55.80 \\
    \textit{GD-UAP (one-sample)} & 50.69 & 48.56 & 42.82 & 46.01 & 32.45 & 35.05 & 47.23 & 55.43 \\
    \textit{L4A-base} & 72.61 & 72.82 & 73.88 & 73.30 & 67.07 & 63.51 & 67.61 & 73.99 \\
    \textit{L4A-fuse} & 72.82 & 73.24 & 73.94 & 73.51 & 66.17 & 62.93 & 68.19 & 73.56 \\
    \textit{L4A-ugs} & 74.31 & 74.26 & 74.10 & 73.24 & 69.04 & 68.40 & 74.63 & 74.84 \\
    \textit{PD-UAP} & 72.82 & 60.74 & 59.84 & 62.82 & 65.74 & 49.15 & 60.16 & 73.14 \\
    \textit{SSP} & 72.50 & 71.86 & 73.09 & 72.98 & 66.17 & 59.89 & 71.38 & 73.83 \\
    \textit{STD} & 73.40 & 71.91 & 73.19 & 71.28 & 69.31 & 68.56 & 72.98 & 73.72 \\
    \textit{UAP (DeepFool)} & 69.26 & 65.69 & 67.02 & 68.67 & 68.19 & 61.01 & 62.29 & 69.89 \\
    \textit{UAPEPGD} & 73.46 & 73.19 & 73.40 & 72.82 & 68.40 & 68.14 & 74.10 & 73.78 \\

    \hline
     Clean Accuracy & 79.97 & 76.76 & 77.02 & 75.43 & 73.19 & 72.13 & 77.45 & 77.61 \\
    IAA Avg. & 40.02 \textcolor{red}{\tiny{$\downarrow$50\%}}& 37.65\textcolor{red}{\tiny{$\downarrow$51\%}} & 38.88\textcolor{red}{\tiny{$\downarrow$50\%}} & 40.14\textcolor{red}{\tiny{$\downarrow$50\%}} & 33.50\textcolor{red}{\tiny{$\downarrow$54\%}} & 33.86\textcolor{red}{\tiny{$\downarrow$53\%}} & 38.96\textcolor{red}{\tiny{$\downarrow$50\%}} & 41.30\textcolor{red}{\tiny{$\downarrow$47\%}} \\
    UAP Avg. & 63.57\textcolor{red}{\tiny{$\downarrow$17\%}}  & 61.54\textcolor{red}{\tiny{$\downarrow$20\%}} & 61.52\textcolor{red}{\tiny{$\downarrow$20\%}}  & 62.51\textcolor{red}{\tiny{$\downarrow$17\%}} & 59.00\textcolor{red}{\tiny{$\downarrow$19\%}} & 53.20\textcolor{red}{\tiny{$\downarrow$26\%}} & 59.85\textcolor{red}{\tiny{$\downarrow$23\%}}  & 64.98\textcolor{red}{\tiny{$\downarrow$16\%}} \\

    Adv Avg. & 51.11\textcolor{red}{\tiny{$\downarrow$34\%}}  & 48.90\textcolor{red}{\tiny{$\downarrow$36\%}} & 49.53\textcolor{red}{\tiny{$\downarrow$36\%}}  & 50.67\textcolor{red}{\tiny{$\downarrow$33\%}} & 45.51\textcolor{red}{\tiny{$\downarrow$38\%}} & 42.96\textcolor{red}{\tiny{$\downarrow$40\%}} & 48.79\textcolor{red}{\tiny{$\downarrow$37\%}}  & 52.45\textcolor{red}{\tiny{$\downarrow$32\%}} \\

    \midrule
    \end{tabular}
    \end{adjustbox}
\end{table*}

\clearpage
\subsubsection{Flowers}
\begin{table*}[h]
    \centering
    \scriptsize
    \caption{This table presents the results of various instance and universal adversarial perturbation (UAP) attacks on the Flowers dataset, with all UAP attack names in \textit{italics}. Different configurations of FGSM and PGD are denoted, such as $FGSM_1$ and $PGD_1$. Average results for universal adversarial perturbations (UAP Avg.), instance adversarial attacks (IAA Avg.), and overall adversarial performance (Adv Avg.) are reported at the bottom, including percentage drops relative to clean accuracy.}
    \label{tab:attack_results_models}

    \begin{adjustbox}{width=\textwidth}
    \begin{tabular}{lllllllll}
    \toprule
    {} & Barlow & BYOL & DINO & MocoV3 & SimCLR & Supervised & SwAV & VICReg \\
    \midrule
    $FGSM_1$ & 66.36 & 57.69 & 57.37 & 64.52 & 48.50 & 41.85 & 46.97 & 66.36\\
    $FGSM_2$ & 25.96 & 17.49 & 19.44 & 24.96 & 19.00 & 7.68 & 13.33 & 25.96\\
    
    $PGD_1$  & 66.03 & 55.99 & 55.60 & 63.31 & 50.45 & 36.97 & 46.65 & 65.81\\
    $PGD_2$  & 1.51 & 0.37 & 0.17 & 1.10 & 0.15 & 0.00 & 0.06 & 1.65\\
    
    $PGD_3$  & 66.19 & 56.37 & 55.95 & 63.50 & 51.00 & 37.31 & 46.72 & 66.44\\
    $PGD_4$  & 1.21 & 0.38 & 0.13 & 0.90 & 0.13 & 0.00 & 0.02 & 1.29\\
    
    $PGD_5$ & 8.03 & 4.90 & 2.81 & 7.17 & 0.92 & 0.72 & 0.89 & 8.05\\
    
    DI2FGSM & 74.42 & 72.08 & 69.73 & 75.75 & 62.56 & 56.94 & 67.56 & 78.12\\
    
    CW & 0.00 & 0.00 & 0.05 & 0.00 & 0.00 & 0.02 & 0.00 & 0.00\\
    
    Jitter & 84.93 & 80.12 & 81.87 & 82.53 & 79.85 & 73.62 & 79.24 & 84.33\\
    
    TIFGSM & 86.85 & 84.35 & 87.48 & 86.17 & 81.29 & 75.36 & 84.39 & 87.88\\
    
    PIFGSM & 53.81 & 43.06 & 39.04 & 51.65 & 29.16 & 27.46 & 28.63 & 53.85\\
    
    EADEN & 0.00 & 0.00 & 0.00 & 0.00 & 0.00 & 0.00 & 0.00 & 0.00\\
    
    OnePixel & 94.47 & 92.77 & 94.79 & 93.10 & 89.27 & 88.38 & 92.94 & 94.49\\
    
    Pixle & 35.32 & 38.34 & 31.21 & 45.08 & 32.07 & 20.88 & 24.05 & 35.09\\
    
    SPSA & 93.03 & 90.21 & 92.91 & 91.84 & 85.56 & 84.31 & 90.60 & 92.84\\
    
    Square & 6.70 & 4.17 & 4.40 & 5.60 & 4.90 & 0.06 & 3.32 & 6.70\\
    
    TAP & 94.01 & 92.77 & 94.76 & 93.31 & 89.76 & 75.93 & 93.14 & 94.01\\
    \textit{ASV} & 81.20 & 81.11 & 88.69 & 86.44 & 77.76 & 68.66 & 80.83 & 81.60 \\
    \textit{FFF (mean-std)} & 69.41 & 71.24 & 72.18 & 73.24 & 63.71 & 69.07 & 69.08 & 69.32 \\
    \textit{FFF (no-data)} & 52.74 & 49.93 & 43.70 & 63.13 & 58.11 & 49.88 & 32.65 & 52.55 \\
    \textit{FFF (one-sample)} & 23.79 & 50.12 & 37.17 & 77.48 & 31.42 & 16.90 & 20.18 & 22.73 \\
    \textit{FG-UAP} & 7.47 & 9.08 & 9.72 & 10.72 & 22.59 & 5.65 & 5.19 & 7.28 \\
    \textit{GD-UAP (mean-std)} & 78.75 & 67.02 & 64.74 & 76.07 & 69.99 & 70.11 & 65.18 & 78.08 \\
    \textit{GD-UAP (no-data)} & 48.38 & 51.64 & 39.25 & 57.48 & 57.36 & 42.94 & 35.00 & 47.88 \\
    \textit{GD-UAP (one-sample)} & 22.13 & 45.88 & 13.95 & 21.55 & 19.79 & 13.95 & 16.12 & 25.00 \\
    \textit{L4A-base} & 56.73 & 69.04 & 82.30 & 76.55 & 47.71 & 40.11 & 31.88 & 56.64 \\
    \textit{L4A-fuse} & 57.26 & 68.86 & 82.66 & 76.63 & 48.40 & 38.92 & 31.94 & 56.94 \\
    \textit{L4A-ugs} & 89.36 & 86.32 & 91.14 & 90.80 & 81.58 & 63.86 & 77.50 & 89.35 \\
    \textit{PD-UAP} & 85.11 & 59.04 & 48.22 & 68.65 & 75.03 & 53.21 & 52.31 & 84.59 \\
    \textit{SSP} & 53.40 & 47.61 & 69.40 & 87.01 & 70.49 & 22.10 & 42.08 & 54.17 \\
    \textit{STD} & 72.29 & 72.70 & 72.37 & 75.94 & 61.46 & 63.29 & 68.90 & 72.30 \\
    \textit{UAP (DeepFool)} & 31.95 & 34.80 & 28.73 & 34.62 & 25.15 & 44.85 & 22.05 & 28.10 \\
    \textit{UAPEPGD} & 90.23 & 88.85 & 91.75 & 91.00 & 81.15 & 81.37 & 88.44 & 90.18 \\

    \hline
    Clean Accuracy & 94.92 & 93.36 & 95.23 & 94.07 & 90.57 & 90.59 & 93.84 & 94.92 \\
    IAA Avg. & 47.71 \textcolor{red}{\tiny{$\downarrow$50\%}}& 43.94\textcolor{red}{\tiny{$\downarrow$53\%}} & 43.76\textcolor{red}{\tiny{$\downarrow$54\%}} & 47.25\textcolor{red}{\tiny{$\downarrow$50\%}} & 40.25\textcolor{red}{\tiny{$\downarrow$56\%}} & 34.86\textcolor{red}{\tiny{$\downarrow$62\%}} & 39.92\textcolor{red}{\tiny{$\downarrow$58\%}} & 47.94\textcolor{red}{\tiny{$\downarrow$50\%}} \\
    UAP Avg. & 57.51\textcolor{red}{\tiny{$\downarrow$39\%}} & 59.58\textcolor{red}{\tiny{$\downarrow$36\%}} & 58.50\textcolor{red}{\tiny{$\downarrow$39\%}} & 66.71\textcolor{red}{\tiny{$\downarrow$29\%}} & 55.73\textcolor{red}{\tiny{$\downarrow$38\%}} & 46.55\textcolor{red}{\tiny{$\downarrow$49\%}} & 46.21\textcolor{red}{\tiny{$\downarrow$51\%}} & 57.29\textcolor{red}{\tiny{$\downarrow$40\%}} \\

    Adv Avg. & 52.32\textcolor{red}{\tiny{$\downarrow$45\%}} & 51.30\textcolor{red}{\tiny{$\downarrow$45\%}} & 50.70\textcolor{red}{\tiny{$\downarrow$47\%}} & 56.41\textcolor{red}{\tiny{$\downarrow$40\%}} & 47.54\textcolor{red}{\tiny{$\downarrow$48\%}} & 40.36\textcolor{red}{\tiny{$\downarrow$55\%}} & 42.88\textcolor{red}{\tiny{$\downarrow$54\%}} & 52.34\textcolor{red}{\tiny{$\downarrow$45\%}} \\

    \midrule
    \end{tabular}
    \end{adjustbox}
\end{table*}

\clearpage
\subsubsection{Food}
\begin{table*}[h]
    \centering
    \scriptsize
    \caption{This table presents the results of various instance and universal adversarial perturbation (UAP) attacks on the Food dataset, with all UAP attack names in \textit{italics}. Different configurations of FGSM and PGD are denoted, such as $FGSM_1$ and $PGD_1$. Average results for universal adversarial perturbations (UAP Avg.), instance adversarial attacks (IAA Avg.), and overall adversarial performance (Adv Avg.) are reported at the bottom, including percentage drops relative to clean accuracy.}
    \label{tab:attack_results_models}

    \begin{adjustbox}{width=\textwidth}
    \begin{tabular}{lllllllll}
    \toprule

    {} & Barlow & BYOL & DINO & MocoV3 & SimCLR & Supervised & SwAV & VICReg \\
    \midrule
    
    $FGSM_1$ & 26.40 & 19.34 & 14.13 & 28.69 & 12.10 & 13.18 & 12.95 & 23.48 \\
    $FGSM_2$ & 3.24 & 1.50 & 1.39 & 4.02 & 1.41 & 1.29 & 0.95 & 2.52 \\
 
    $PGD_1$ & 26.60 & 19.03 & 13.87 & 28.54 & 13.69 & 11.30 & 13.15 & 23.91 \\
    $PGD_2$  & 0.04 & 0.01 & 0.01 & 0.05 & 0.00 & 0.02 & 0.00 & 0.04 \\

    $PGD_3$  & 26.72 & 19.21 & 14.13 & 28.76 & 13.92 & 11.42 & 13.48 & 24.12 \\
    $PGD_4$ & 0.04 & 0.01 & 0.00 & 0.04 & 0.00 & 0.01 & 0.00 & 0.03 \\
    
    $PGD_5$  & 0.59 & 0.19 & 0.10 & 0.82 & 0.04 & 0.13 & 0.01 & 0.47 \\
   
    DI2FGSM  & 44.15 & 37.23 & 37.35 & 44.94 & 33.02 & 32.45 & 37.32 & 40.14 \\
    
    CW & 0.00 & 0.00 & 0.00 & 0.00 & 0.00 & 0.00 & 0.00 & 0.00 \\
    
    Jitter & 60.70 & 56.00 & 61.34 & 58.14 & 55.13 & 53.14 & 61.79 & 59.79 \\
    
    TIFGSM & 57.43 & 51.93 & 53.38 & 56.41 & 48.65 & 45.76 & 54.04 & 56.51 \\
    
    PIFGSM & 17.53 & 11.71 & 6.67 & 19.93 & 5.17 & 6.80 & 5.46 & 14.85 \\
    
    EADEN & 0.00 & 0.00 & 0.00 & 0.00 & 0.00 & 0.00 & 0.00 & 0.00 \\
    
    OnePixel & 73.54 & 69.95 & 76.00 & 71.41 & 63.59 & 64.63 & 73.65 & 73.22 \\
    
    Pixle & 14.94 & 12.97 & 9.65 & 17.11 & 8.34 & 5.84 & 4.93 & 13.66 \\
    
    SPSA & 68.75 & 64.12 & 69.31 & 66.70 & 57.49 & 56.88 & 67.11 & 68.04 \\
    
    Square & 0.19 & 0.05 & 0.09 & 0.19 & 0.16 & 0.02 & 0.07 & 0.16 \\
    
    TAP & 74.21 & 71.43 & 76.25 & 72.68 & 65.96 & 53.74 & 76.18 & 73.75 \\
    \textit{ASV} & 62.46 & 59.82 & 66.04 & 61.50 & 52.91 & 47.60 & 62.86 & 62.02 \\
    \textit{FFF (mean-std)} & 51.98 & 49.45 & 55.35 & 49.68 & 44.52 & 41.29 & 53.77 & 49.99 \\
    \textit{FFF (no-data)} & 45.78 & 29.52 & 39.07 & 47.63 & 44.67 & 25.60 & 24.66 & 43.53 \\
    \textit{FFF (one-sample)} & 20.82 & 34.82 & 29.72 & 36.80 & 4.63 & 20.67 & 10.41 & 19.50 \\
    \textit{FG-UAP} & 3.98 & 2.50 & 2.22 & 4.28 & 3.90 & 1.84 & 1.67 & 3.88 \\
    \textit{GD-UAP (mean-std)} & 59.88 & 49.04 & 52.61 & 51.72 & 47.22 & 44.65 & 51.52 & 58.62 \\
    \textit{GD-UAP (no-data)} & 42.31 & 38.30 & 30.52 & 46.00 & 37.85 & 22.85 & 32.35 & 40.77 \\
    \textit{GD-UAP (one-sample)} & 13.89 & 29.22 & 8.69 & 14.46 & 5.33 & 7.26 & 15.75 & 12.92 \\
    \textit{L4A-base} & 26.08 & 35.83 & 48.51 & 43.13 & 5.36 & 15.57 & 7.30 & 25.89 \\
    \textit{L4A-fuse} & 25.76 & 35.50 & 47.07 & 43.72 & 5.11 & 15.43 & 7.70 & 25.86 \\
    \textit{L4A-ugs} & 58.75 & 46.16 & 53.52 & 61.22 & 41.61 & 18.25 & 39.10 & 57.71 \\
    \textit{PD-UAP} & 63.56 & 46.55 & 33.67 & 50.59 & 50.14 & 29.66 & 37.59 & 62.90 \\
    \textit{SSP} & 34.22 & 26.85 & 35.80 & 41.64 & 26.52 & 6.07 & 23.58 & 28.30 \\
    \textit{STD} & 45.39 & 37.86 & 42.17 & 40.49 & 34.37 & 34.74 & 43.51 & 44.38 \\
    \textit{UAP (DeepFool)} & 16.28 & 18.95 & 16.98 & 17.35 & 19.64 & 15.37 & 7.83 & 10.85 \\
    \textit{UAPEPGD} & 69.89 & 65.73 & 71.68 & 68.50 & 57.03 & 59.03 & 68.59 & 69.47 \\

    \hline
    Clean Accuracy & 76.09 & 73.07 & 78.42 & 73.83 & 67.24 & 69.05 & 76.51 & 75.81 \\
    IAA Avg. & 27.50 \textcolor{red}{\tiny{$\downarrow$64\%}}& 24.15\textcolor{red}{\tiny{$\downarrow$67\%}} & 24.09\textcolor{red}{\tiny{$\downarrow$69\%}} & 27.69\textcolor{red}{\tiny{$\downarrow$62\%}} & 21.03\textcolor{red}{\tiny{$\downarrow$69\%}} & 19.81\textcolor{red}{\tiny{$\downarrow$71\%}} & 23.39\textcolor{red}{\tiny{$\downarrow$69\%}} & 26.37\textcolor{red}{\tiny{$\downarrow$65\%}} \\
     UAP Avg. & 40.06\textcolor{red}{\tiny{$\downarrow$47\%}} & 37.88\textcolor{red}{\tiny{$\downarrow$48\%}} & 39.60\textcolor{red}{\tiny{$\downarrow$50\%}} & 42.42\textcolor{red}{\tiny{$\downarrow$43\%}} & 30.05\textcolor{red}{\tiny{$\downarrow$55\%}} & 25.37\textcolor{red}{\tiny{$\downarrow$63\%}} & 30.51\textcolor{red}{\tiny{$\downarrow$60\%}} & 38.54\textcolor{red}{\tiny{$\downarrow$49\%}} \\

    Adv Avg. & 33.41\textcolor{red}{\tiny{$\downarrow$56\%}} & 30.61\textcolor{red}{\tiny{$\downarrow$58\%}} & 31.39\textcolor{red}{\tiny{$\downarrow$60\%}} & 34.62\textcolor{red}{\tiny{$\downarrow$53\%}} & 25.28\textcolor{red}{\tiny{$\downarrow$62\%}} & 22.43\textcolor{red}{\tiny{$\downarrow$68\%}} & 26.74\textcolor{red}{\tiny{$\downarrow$65\%}} & 32.10\textcolor{red}{\tiny{$\downarrow$58\%}} \\

    \midrule
    \end{tabular}
    \end{adjustbox}
\end{table*}

\clearpage
\subsubsection{Pets}
\begin{table*}[h]
    \centering
    \scriptsize
    \caption{This table presents the results of various instance and universal adversarial perturbation (UAP) attacks on the Pets dataset, with all UAP attack names in \textit{italics}. Different configurations of FGSM and PGD are denoted, such as $FGSM_1$ and $PGD_1$. Average results for universal adversarial perturbations (UAP Avg.), instance adversarial attacks (IAA Avg.), and overall adversarial performance (Adv Avg.) are reported at the bottom, including percentage drops relative to clean accuracy.}
    \label{tab:attack_results_methods}

    \begin{adjustbox}{width=\textwidth}
    \begin{tabular}{lllllllll}
    \toprule
    Method & Barlow & BYOL & DINO & MocoV3 & SimCLR & Supervised & SwAV & VICReg \\
    \midrule
    $FGSM_1$  & 63.58 & 61.00 & 48.74 & 71.38 & 44.60 & 55.10 & 41.59 & 63.58 \\
    $FGSM_2$  & 25.08 & 21.62 & 11.81 & 34.65 & 17.20 & 14.17 & 8.74 & 25.08 \\
   
    $PGD_1$  & 64.38 & 60.82 & 48.07 & 71.07 & 46.76 & 52.20 & 43.00 & 64.30 \\
    $PGD_2$  & 0.82 & 0.41 & 0.08 & 2.96 & 0.16 & 0.00 & 0.03 & 0.79 \\
    
    $PGD_3$   & 64.52 & 61.21 & 48.10 & 71.29 & 47.25 & 52.21 & 43.42 & 64.52 \\
    $PGD_4$   & 0.63 & 0.27 & 0.03 & 2.39 & 0.11 & 0.00 & 0.03 & 0.57 \\
    
    $PGD_5$  & 6.54 & 5.69 & 0.89 & 14.03 & 0.98 & 1.38 & 0.43 & 6.51 \\
   
    DI2FGSM & 73.92 & 71.18 & 63.92 & 78.63 & 61.06 & 68.25 & 59.70 & 74.18 \\
   
    CW & 0.00 & 0.00 & 0.00 & 0.03 & 0.00 & 0.00 & 0.00 & 0.00 \\
    
    Jitter & 80.75 & 79.82 & 75.82 & 84.06 & 74.50 & 78.41 & 75.60 & 80.83 \\
    
    TIFGSM  & 81.43 & 80.30 & 78.13 & 84.89 & 75.31 & 80.60 & 76.11 & 82.31 \\
    
    PIFGSM & 54.24 & 51.35 & 34.23 & 64.67 & 31.70 & 41.02 & 26.11 & 54.24 \\
    
    EADEN & 0.00 & 0.00 & 0.00 & 0.00 & 0.00 & 0.00 & 0.00 & 0.00 \\
    OnePixel & 88.28 & 87.90 & 87.65 & 89.82 & 81.51 & 90.60 & 85.85 & 88.31 \\
    Pixle & 42.04 & 41.46 & 38.47 & 59.12 & 31.61 & 43.01 & 29.16 & 42.31 \\
    SPSA & 87.27 & 86.87 & 85.81 & 88.79 & 79.93 & 88.54 & 83.95 & 87.46 \\
    Square & 3.28 & 1.71 & 0.49 & 4.79 & 3.88 & 0.05 & 0.46 & 3.30 \\
    TAP & 88.97 & 89.04 & 88.41 & 90.66 & 83.01 & 86.83 & 87.09 & 88.97 \\
    \textit{ASV} & 84.04 & 86.07 & 86.52 & 89.11 & 79.46 & 86.89 & 83.59 & 84.07 \\
    \textit{FFF (mean-std)} & 79.82 & 82.97 & 82.04 & 85.08 & 73.86 & 83.03 & 81.53 & 79.69 \\
    \textit{FFF (no-data)} & 65.60 & 47.88 & 70.12 & 80.33 & 53.54 & 71.01 & 53.59 & 62.52 \\
    \textit{FFF (one-sample)} & 49.62 & 85.48 & 61.54 & 83.37 & 63.28 & 73.35 & 61.44 & 56.39 \\
    \textit{FG-UAP} & 12.32 & 6.27 & 8.53 & 14.09 & 12.36 & 8.46 & 5.18 & 12.34 \\
    \textit{GD-UAP (mean-std)} & 84.78 & 81.57 & 75.38 & 84.37 & 75.67 & 84.82 & 80.65 & 85.93 \\
    \textit{GD-UAP (no-data)} & 67.48 & 63.61 & 55.35 & 79.27 & 69.36 & 65.81 & 60.80 & 71.45 \\
    \textit{GD-UAP (one-sample)} & 31.59 & 67.83 & 23.04 & 39.50 & 19.63 & 34.28 & 37.79 & 26.46 \\
    \textit{L4A-base} & 84.68 & 84.02 & 82.28 & 88.44 & 74.45 & 79.91 & 61.65 & 84.55 \\
    \textit{L4A-fuse} & 84.54 & 83.83 & 81.98 & 88.16 & 73.96 & 79.23 & 61.69 & 83.96 \\
    \textit{L4A-ugs} & 88.28 & 88.31 & 87.47 & 90.22 & 80.84 & 87.68 & 86.17 & 88.39 \\
    \textit{PD-UAP} & 84.38 & 75.44 & 65.26 & 78.86 & 77.81 & 75.73 & 69.66 & 84.25 \\
    \textit{SSP} & 86.49 & 85.07 & 84.23 & 89.73 & 78.70 & 76.96 & 78.96 & 86.29 \\
    \textit{STD} & 85.41 & 85.83 & 84.30 & 88.53 & 78.35 & 88.91 & 85.46 & 85.54 \\
    \textit{UAP (DeepFool)} & 42.47 & 45.23 & 41.35 & 56.05 & 67.08 & 55.57 & 44.72 & 48.42 \\
    \textit{UAPEPGD} & 88.36 & 88.44 & 87.83 & 89.74 & 79.14 & 90.05 & 87.04 & 88.11 \\

    \hline
    Clean Accuracy & 89.13 & 89.08 & 89.15 & 90.77 & 83.23 & 92.06 & 87.47 & 89.13 \\
    IAA Avg. & 45.87 \textcolor{red}{\tiny{$\downarrow$49\%}}& 44.48\textcolor{red}{\tiny{$\downarrow$50\%}} & 39.48\textcolor{red}{\tiny{$\downarrow$56\%}} & 50.74\textcolor{red}{\tiny{$\downarrow$44\%}} & 37.75\textcolor{red}{\tiny{$\downarrow$55\%}} & 41.79\textcolor{red}{\tiny{$\downarrow$55\%}} & 36.73\textcolor{red}{\tiny{$\downarrow$58\%}} & 45.95\textcolor{red}{\tiny{$\downarrow$48.4\%}} \\
    \textit{UAP Avg.} & 69.99\textcolor{red}{\tiny{$\downarrow$21\%}} & 72.36\textcolor{red}{\tiny{$\downarrow$19\%}} & 67.33\textcolor{red}{\tiny{$\downarrow$24\%}} & 76.55\textcolor{red}{\tiny{$\downarrow$16\%}} & 66.09\textcolor{red}{\tiny{$\downarrow$21\%}} & 71.36\textcolor{red}{\tiny{$\downarrow$22\%}} & 64.99\textcolor{red}{\tiny{$\downarrow$26\%}} & 70.52\textcolor{red}{\tiny{$\downarrow$21\%}} \\

    Adv Avg. & 57.22 \textcolor{red}{\tiny{$\downarrow$36\%}}  & 57.60\textcolor{red}{\tiny{$\downarrow$35\%}} & 52.58\textcolor{red}{\tiny{$\downarrow$41\%}}  & 62.88\textcolor{red}{\tiny{$\downarrow$31\%}} & 51.09\textcolor{red}{\tiny{$\downarrow$39\%}} & 55.71\textcolor{red}{\tiny{$\downarrow$39\%}} & 50.03\textcolor{red}{\tiny{$\downarrow$43\%}}  & 57.52\textcolor{red}{\tiny{$\downarrow$35\%}} \\
    \midrule
    \end{tabular}
    \end{adjustbox}
\end{table*}

\clearpage
\subsection{Transfer Learning (Finetune)}\label{res:combined}
\begin{table*}[htbp]
\scriptsize
\centering
\caption{Combined results from transfer learning datasets showing Clean accuracy, UAP Avg. , IAA Avg. , and Adv Avg.  with percentage drops relative to Clean accuracy.}
\label{tab:results}
\begin{tabular}{@{}llllllllll@{}}
\toprule
{} & {} & Barlow & BYOL & DINO & MocoV3 & SimCLR & Supervised & SwAV & VICReg \\
\midrule
\multirow{4}{*}{\rotatebox{90}{Aircraft}} 
         &Clean  & 86.71 & 83.14 & 80.38  & 82.74 & 79.35 & 85.08 & 86.17 & 86.95 \\
    &IAA  & 39.62 \textcolor{red}{\tiny{$\downarrow$54\%}}  & 39.21\textcolor{red}{\tiny{$\downarrow$53\%}} & 32.36\textcolor{red}{\tiny{$\downarrow$60\%}}  & 40.18\textcolor{red}{\tiny{$\downarrow$51\%}} & 30.68\textcolor{red}{\tiny{$\downarrow$61\%}} & 45.4\textcolor{red}{\tiny{$\downarrow$47\%}} & 42.47\textcolor{red}{\tiny{$\downarrow$51\%}}  & 39.21\textcolor{red}{\tiny{$\downarrow$55\%}} \\
    &UAP  & 36.47\textcolor{red}{\tiny{$\downarrow$58\%}} & 30.12\textcolor{red}{\tiny{$\downarrow$64\%}} & 30.45\textcolor{red}{\tiny{$\downarrow$62\%}} & 38.95\textcolor{red}{\tiny{$\downarrow$53\%}} & 33.90\textcolor{red}{\tiny{$\downarrow$57\%}} & 42.25\textcolor{red}{\tiny{$\downarrow$50\%}} & 26.72\textcolor{red}{\tiny{$\downarrow$69\%}} & 35.36\textcolor{red}{\tiny{$\downarrow$59\%}} \\

    &Adv  & 38.14\textcolor{red}{\tiny{$\downarrow$56\%}} & 34.93\textcolor{red}{\tiny{$\downarrow$58\%}} & 31.46\textcolor{red}{\tiny{$\downarrow$61\%}} & 39.60\textcolor{red}{\tiny{$\downarrow$52\%}} & 32.20\textcolor{red}{\tiny{$\downarrow$59\%}} & 43.92\textcolor{red}{\tiny{$\downarrow$48\%}} & 35.06\textcolor{red}{\tiny{$\downarrow$59\%}} & 37.40\textcolor{red}{\tiny{$\downarrow$57\%}} \\
\midrule
\multirow{4}{*}{\rotatebox{90}{Caltech}} 
    &Clean  & 91.44 & 92.03 & 89.77  & 92.92 & 90.39 & 91.81 & 90.37 & 91.32 \\
    &IAA  & 53.63 \textcolor{red}{\tiny{$\downarrow$41\%}}  & 57.03\textcolor{red}{\tiny{$\downarrow$38\%}} & 49.71\textcolor{red}{\tiny{$\downarrow$45\%}}  & 59.32\textcolor{red}{\tiny{$\downarrow$36\%}} & 53.86\textcolor{red}{\tiny{$\downarrow$40\%}} & 54.18\textcolor{red}{\tiny{$\downarrow$41\%}} & 48.01\textcolor{red}{\tiny{$\downarrow$47\%}}  & 53.78\textcolor{red}{\tiny{$\downarrow$41\%}} \\
    &UAP  & 62.03\textcolor{red}{\tiny{$\downarrow$32\%}} & 63.09\textcolor{red}{\tiny{$\downarrow$31\%}} & 55.59\textcolor{red}{\tiny{$\downarrow$38\%}} & 64.37\textcolor{red}{\tiny{$\downarrow$31\%}} & 67.38\textcolor{red}{\tiny{$\downarrow$25\%}} & 58.60\textcolor{red}{\tiny{$\downarrow$36\%}} & 37.55\textcolor{red}{\tiny{$\downarrow$58\%}} & 62.47\textcolor{red}{\tiny{$\downarrow$32\%}} \\

    &Adv  & 57.58\textcolor{red}{\tiny{$\downarrow$37\%}} & 59.88\textcolor{red}{\tiny{$\downarrow$35\%}} & 52.48\textcolor{red}{\tiny{$\downarrow$42\%}} & 61.70\textcolor{red}{\tiny{$\downarrow$34\%}} & 60.23\textcolor{red}{\tiny{$\downarrow$33\%}} & 56.26\textcolor{red}{\tiny{$\downarrow$39\%}} & 43.09\textcolor{red}{\tiny{$\downarrow$52\%}} & 57.87\textcolor{red}{\tiny{$\downarrow$37\%}} \\
\midrule
\multirow{4}{*}{\rotatebox{90}{Cars}} 
    &Clean  & 90.54 & 89.91 & 89.80  & 90.11 & 89.62 & 88.99 & 89.73 & 90.41 \\
    &IAA  & 48.69 \textcolor{red}{\tiny{$\downarrow$46\%}}  & 54.91\textcolor{red}{\tiny{$\downarrow$39\%}} & 52.72\textcolor{red}{\tiny{$\downarrow$52\%}}  & 55.39\textcolor{red}{\tiny{$\downarrow$39\%}} & 47.63\textcolor{red}{\tiny{$\downarrow$47\%}} & 55.47\textcolor{red}{\tiny{$\downarrow$38\%}} & 52.04\textcolor{red}{\tiny{$\downarrow$42\%}}  & 48.70\textcolor{red}{\tiny{$\downarrow$48\%}} \\
    &UAP  & 45.75 \textcolor{red}{\tiny{$\downarrow$49\%}} & 60.09\textcolor{red}{\tiny{$\downarrow$33\%}} & 49.85\textcolor{red}{\tiny{$\downarrow$44\%}} & 65.77\textcolor{red}{\tiny{$\downarrow$27\%}} & 34.10\textcolor{red}{\tiny{$\downarrow$62\%}} & 55.54\textcolor{red}{\tiny{$\downarrow$38\%}} & 50.81\textcolor{red}{\tiny{$\downarrow$43\%}} & 44.46\textcolor{red}{\tiny{$\downarrow$51\%}} \\

    &Adv  & 47.31 \textcolor{red}{\tiny{$\downarrow$48\%}} & 57.35\textcolor{red}{\tiny{$\downarrow$36\%}} & 52.74\textcolor{red}{\tiny{$\downarrow$41\%}} & 60.28\textcolor{red}{\tiny{$\downarrow$33\%}} & 42.02\textcolor{red}{\tiny{$\downarrow$53\%}} & 55.50\textcolor{red}{\tiny{$\downarrow$38\%}} & 52.88\textcolor{red}{\tiny{$\downarrow$41\%}} & 47.90\textcolor{red}{\tiny{$\downarrow$47\%}} \\

\midrule
\multirow{4}{*}{\rotatebox{90}{CIFAR 10}} 
    &Clean  & 97.13 & 96.89 & 96.90  & 96.86 & 97.22 & 96.16 & 96.75 & 97.07 \\
    &IAA  & 48.74 \textcolor{red}{\tiny{$\downarrow$50\%}}  & 46.26\textcolor{red}{\tiny{$\downarrow$52\%}} & 54.18\textcolor{red}{\tiny{$\downarrow$44\%}}  & 49.66\textcolor{red}{\tiny{$\downarrow$49\%}} & 49.03\textcolor{red}{\tiny{$\downarrow$50\%}} & 43.37\textcolor{red}{\tiny{$\downarrow$55\%}} & 55.75\textcolor{red}{\tiny{$\downarrow$43\%}}  & 48.31\textcolor{red}{\tiny{$\downarrow$50\%}} \\
    &UAP  & 55.07 \textcolor{red}{\tiny{$\downarrow$43\%}} & 31.96\textcolor{red}{\tiny{$\downarrow$67\%}} & 58.16\textcolor{red}{\tiny{$\downarrow$40\%}} & 44.72\textcolor{red}{\tiny{$\downarrow$54\%}} & 55.76\textcolor{red}{\tiny{$\downarrow$43\%}} & 36.20\textcolor{red}{\tiny{$\downarrow$62\%}} & 53.76\textcolor{red}{\tiny{$\downarrow$44\%}} & 50.00\textcolor{red}{\tiny{$\downarrow$48\%}} \\

    &Adv  & 51.72 \textcolor{red}{\tiny{$\downarrow$47\%}}  & 39.53\textcolor{red}{\tiny{$\downarrow$59\%}} & 56.05\textcolor{red}{\tiny{$\downarrow$42\%}}  & 47.34\textcolor{red}{\tiny{$\downarrow$51\%}} & 52.20\textcolor{red}{\tiny{$\downarrow$46\%}} & 40.00\textcolor{red}{\tiny{$\downarrow$58\%}} & 54.81\textcolor{red}{\tiny{$\downarrow$43\%}}  & 49.10\textcolor{red}{\tiny{$\downarrow$49\%}} \\
\midrule
\multirow{4}{*}{\rotatebox{90}{CIFAR 100}} 
    &Clean  & 84.60 & 83.88 & 84.69  & 84.49 & 84.44 & 82.63 & 84.37 & 84.27 \\
    &IAA  & 35.85 \textcolor{red}{\tiny{$\downarrow$57\%}}  & 44.11\textcolor{red}{\tiny{$\downarrow$47\%}} & 39.68\textcolor{red}{\tiny{$\downarrow$53\%}}  & 44.95\textcolor{red}{\tiny{$\downarrow$47\%}} & 35.54\textcolor{red}{\tiny{$\downarrow$58\%}} & 42.33\textcolor{red}{\tiny{$\downarrow$49\%}} & 40.22\textcolor{red}{\tiny{$\downarrow$52\%}}  & 35.59\textcolor{red}{\tiny{$\downarrow$58\%}} \\
   &UAP  & 36.41 \textcolor{red}{\tiny{$\downarrow$57\%}} & 36.79\textcolor{red}{\tiny{$\downarrow$56\%}} & 42.02\textcolor{red}{\tiny{$\downarrow$50\%}} & 31.57\textcolor{red}{\tiny{$\downarrow$63\%}} & 28.24\textcolor{red}{\tiny{$\downarrow$67\%}} & 24.94\textcolor{red}{\tiny{$\downarrow$70\%}} & 41.03\textcolor{red}{\tiny{$\downarrow$51\%}} & 31.40\textcolor{red}{\tiny{$\downarrow$63\%}} \\

    &Adv  & 36.12 \textcolor{red}{\tiny{$\downarrow$57\%}} & 40.67\textcolor{red}{\tiny{$\downarrow$52\%}} & 40.78\textcolor{red}{\tiny{$\downarrow$52\%}} & 38.65\textcolor{red}{\tiny{$\downarrow$54\%}} & 32.10\textcolor{red}{\tiny{$\downarrow$62\%}} & 34.14\textcolor{red}{\tiny{$\downarrow$59\%}} & 40.60\textcolor{red}{\tiny{$\downarrow$52\%}} & 33.62\textcolor{red}{\tiny{$\downarrow$60\%}} \\
\midrule
\multirow{4}{*}{\rotatebox{90}{DTD}} 
    &Clean  & 76.12 & 76.28 & 78.09  & 75.43 & 75.90 & 73.30 & 74.47 & 77.29 \\
    &IAA  & 41.42 \textcolor{red}{\tiny{$\downarrow$46\%}}  & 41.89\textcolor{red}{\tiny{$\downarrow$45\%}} & 40.50\textcolor{red}{\tiny{$\downarrow$48\%}}  & 42.91\textcolor{red}{\tiny{$\downarrow$43\%}} & 42.42\textcolor{red}{\tiny{$\downarrow$44\%}} & 38.99\textcolor{red}{\tiny{$\downarrow$47\%}} & 39.65\textcolor{red}{\tiny{$\downarrow$47\%}}  & 41.90\textcolor{red}{\tiny{$\downarrow$46\%}} \\
    &UAP  & 62.74 \textcolor{red}{\tiny{$\downarrow$18\%}} & 62.74\textcolor{red}{\tiny{$\downarrow$18\%}} & 61.26\textcolor{red}{\tiny{$\downarrow$22\%}} & 63.58\textcolor{red}{\tiny{$\downarrow$16\%}} & 57.58\textcolor{red}{\tiny{$\downarrow$24\%}} & 57.78\textcolor{red}{\tiny{$\downarrow$21\%}} & 62.41\textcolor{red}{\tiny{$\downarrow$16\%}} & 62.54\textcolor{red}{\tiny{$\downarrow$19\%}} \\

    &Adv  & 51.46 \textcolor{red}{\tiny{$\downarrow$32\%}}  & 51.70\textcolor{red}{\tiny{$\downarrow$32\%}} & 50.27\textcolor{red}{\tiny{$\downarrow$36\%}}  & 52.64\textcolor{red}{\tiny{$\downarrow$30\%}} & 49.56\textcolor{red}{\tiny{$\downarrow$35\%}} & 47.83\textcolor{red}{\tiny{$\downarrow$35\%}} & 50.36\textcolor{red}{\tiny{$\downarrow$32\%}}  & 51.62\textcolor{red}{\tiny{$\downarrow$33\%}} \\

\midrule
\multirow{4}{*}{\rotatebox{90}{Flowers}} 
    &Clean  & 97.41 & 96.74 & 97.18  & 96.70 & 95.18 & 96.73 & 96.68 & 96.86 \\
    &IAA  & 60.24 \textcolor{red}{\tiny{$\downarrow$38\%}}  & 61.32\textcolor{red}{\tiny{$\downarrow$37\%}} & 54.30\textcolor{red}{\tiny{$\downarrow$44\%}}  & 60.98\textcolor{red}{\tiny{$\downarrow$37\%}} & 51.34\textcolor{red}{\tiny{$\downarrow$46\%}} & 60.79\textcolor{red}{\tiny{$\downarrow$37\%}} & 55.66\textcolor{red}{\tiny{$\downarrow$42\%}}  & 59.97\textcolor{red}{\tiny{$\downarrow$38\%}} \\
    &UAP  & 60.69 \textcolor{red}{\tiny{$\downarrow$38\%}} & 67.75\textcolor{red}{\tiny{$\downarrow$30\%}} & 54.18\textcolor{red}{\tiny{$\downarrow$44\%}} & 68.26\textcolor{red}{\tiny{$\downarrow$29\%}} & 60.09\textcolor{red}{\tiny{$\downarrow$37\%}} & 56.33\textcolor{red}{\tiny{$\downarrow$42\%}} & 54.51\textcolor{red}{\tiny{$\downarrow$44\%}} & 63.09\textcolor{red}{\tiny{$\downarrow$35\%}} \\

    &Adv  & 60.45 \textcolor{red}{\tiny{$\downarrow$38\%}}  & 64.34\textcolor{red}{\tiny{$\downarrow$33\%}} & 54.25\textcolor{red}{\tiny{$\downarrow$44\%}}  & 64.41\textcolor{red}{\tiny{$\downarrow$33\%}} & 55.46\textcolor{red}{\tiny{$\downarrow$42\%}} & 58.69\textcolor{red}{\tiny{$\downarrow$39\%}} & 55.12\textcolor{red}{\tiny{$\downarrow$43\%}}  & 61.44\textcolor{red}{\tiny{$\downarrow$37\%}} \\

\midrule
\multirow{4}{*}{\rotatebox{90}{Food}} 
    &Clean  & 83.93 & 85.63 & 87.65  & 85.82 & 82.30 & 84.35 & 87.16 & 83.71 \\
    &IAA  & 27.80 \textcolor{red}{\tiny{$\downarrow$67\%}}  & 36.30\textcolor{red}{\tiny{$\downarrow$58\%}} & 34.12\textcolor{red}{\tiny{$\downarrow$61.06\%}} & 36.09\textcolor{red}{\tiny{$\downarrow$58\%}} & 28.89\textcolor{red}{\tiny{$\downarrow$65\%}} & 34.74\textcolor{red}{\tiny{$\downarrow$59\%}} & 34.26\textcolor{red}{\tiny{$\downarrow$59\%}}  & 31.71\textcolor{red}{\tiny{$\downarrow$62\%}} \\
    &UAP  & 35.93 \textcolor{red}{\tiny{$\downarrow$57\%}} & 43.98\textcolor{red}{\tiny{$\downarrow$49\%}} & 46.95\textcolor{red}{\tiny{$\downarrow$46\%}} & 44.88\textcolor{red}{\tiny{$\downarrow$48\%}} & 32.33\textcolor{red}{\tiny{$\downarrow$61\%}} & 29.07\textcolor{red}{\tiny{$\downarrow$66\%}} & 39.09\textcolor{red}{\tiny{$\downarrow$55\%}} & 41.18\textcolor{red}{\tiny{$\downarrow$51\%}} \\

    &Adv  & 35.94 \textcolor{red}{\tiny{$\downarrow$57\%}}  & 39.91\textcolor{red}{\tiny{$\downarrow$53\%}} & 40.16\textcolor{red}{\tiny{$\downarrow$54\%}}  & 40.23\textcolor{red}{\tiny{$\downarrow$53\%}} & 30.51\textcolor{red}{\tiny{$\downarrow$63\%}} & 32.07\textcolor{red}{\tiny{$\downarrow$62\%}} & 36.53\textcolor{red}{\tiny{$\downarrow$58\%}}  & 36.17\textcolor{red}{\tiny{$\downarrow$57\%}} \\
\midrule
\multirow{4}{*}{\rotatebox{90}{Pets}} 
    &Clean  & 90.89 & 91.34 & 90.24  & 92.19 & 88.51 & 93.94 & 90.50 & 90.92 \\
    &IAA  & 43.21 \textcolor{red}{\tiny{$\downarrow$52\%}}  & 44.69\textcolor{red}{\tiny{$\downarrow$51\%}} & 40.01\textcolor{red}{\tiny{$\downarrow$56\%}}  & 48.33\textcolor{red}{\tiny{$\downarrow$48\%}} & 42.03\textcolor{red}{\tiny{$\downarrow$53\%}} & 46.53\textcolor{red}{\tiny{$\downarrow$50\%}} & 39.53\textcolor{red}{\tiny{$\downarrow$56\%}}  & 43.76\textcolor{red}{\tiny{$\downarrow$52\%}} \\
    &UAP  & 67.27 \textcolor{red}{\tiny{$\downarrow$26\%}} & 71.96\textcolor{red}{\tiny{$\downarrow$21\%}} & 63.83\textcolor{red}{\tiny{$\downarrow$29\%}} & 73.20\textcolor{red}{\tiny{$\downarrow$21\%}} & 67.90\textcolor{red}{\tiny{$\downarrow$23\%}} & 75.66\textcolor{red}{\tiny{$\downarrow$19\%}} & 64.06\textcolor{red}{\tiny{$\downarrow$29\%}} & 67.63\textcolor{red}{\tiny{$\downarrow$26\%}} \\

    &Adv  & 54.53 \textcolor{red}{\tiny{$\downarrow$40\%}}  & 57.52\textcolor{red}{\tiny{$\downarrow$37\%}} & 51.23\textcolor{red}{\tiny{$\downarrow$43\%}}  & 60.03\textcolor{red}{\tiny{$\downarrow$35\%}} & 54.20\textcolor{red}{\tiny{$\downarrow$39\%}} & 60.24\textcolor{red}{\tiny{$\downarrow$36\%}} & 51.07\textcolor{red}{\tiny{$\downarrow$44\%}}  & 54.99\textcolor{red}{\tiny{$\downarrow$40\%}} \\
\midrule
\multirow{4}{*}{\rotatebox{90}{\textbf{All}}} 
     & \textbf{Clean} & 88.75 & 88.43 & 88.30 & 88.58 & 86.99 & 88.11 & 88.47 & 88.76 \\
     & \textbf{IAA} & 44.36\textcolor{red}{\tiny{$\downarrow$50\%}} & 47.30\textcolor{red}{\tiny{$\downarrow$47\%}} & 44.18\textcolor{red}{\tiny{$\downarrow$50\%}} & 48.65\textcolor{red}{\tiny{$\downarrow$45\%}} & 42.38\textcolor{red}{\tiny{$\downarrow$51\%}} & 46.87\textcolor{red}{\tiny{$\downarrow$47\%}} & 45.29\textcolor{red}{\tiny{$\downarrow$49\%}} & 44.77\textcolor{red}{\tiny{$\downarrow$50\%}} \\
     & \textbf{UAP} & 51.37\textcolor{red}{\tiny{$\downarrow$42\%}} & 52.05\textcolor{red}{\tiny{$\downarrow$41\%}} & 51.37\textcolor{red}{\tiny{$\downarrow$42\%}} & 55.03\textcolor{red}{\tiny{$\downarrow$38\%}} & 48.59\textcolor{red}{\tiny{$\downarrow$44\%}} & 48.49\textcolor{red}{\tiny{$\downarrow$45\%}} & 47.77\textcolor{red}{\tiny{$\downarrow$46\%}} & 50.90\textcolor{red}{\tiny{$\downarrow$43\%}} \\
     & \textbf{Adv} & 48.14\textcolor{red}{\tiny{$\downarrow$46\%}} & 49.54\textcolor{red}{\tiny{$\downarrow$44\%}} & 47.71\textcolor{red}{\tiny{$\downarrow$46\%}} & 51.65\textcolor{red}{\tiny{$\downarrow$42\%}} & 45.39\textcolor{red}{\tiny{$\downarrow$48\%}} & 47.63\textcolor{red}{\tiny{$\downarrow$46\%}} & 46.61\textcolor{red}{\tiny{$\downarrow$47\%}} & 47.79\textcolor{red}{\tiny{$\downarrow$46\%}} \\
\bottomrule
\end{tabular}
\end{table*}
\clearpage
\subsubsection{AirCraft}
\begin{table*}[h]
    \centering
    \scriptsize
    \caption{This table presents the results of various instance and universal adversarial perturbation (UAP) attacks on the AirCraft dataset, with all UAP attack names in \textit{italics}. Different configurations of FGSM and PGD are denoted, such as $FGSM_1$ and $PGD_1$. Average results for universal adversarial perturbations (UAP Avg.), instance adversarial attacks (IAA Avg.), and overall adversarial performance (Adv Avg.) are reported at the bottom, including percentage drops relative to clean accuracy.}
    \label{tab:few_shot_small_domain_shift}

    \begin{adjustbox}{width=\textwidth}
    \begin{tabular}{lllllllll}
    \toprule
    
    {} & Barlow & BYOL & DINO & MoCoV3 & SimCLR & Supervised & SwAV & VICReg \\
    \midrule
    
    \textbf{$FGSM_1$}& 8.92 & 5.94 & 4.84 & 11.41 & 2.7 & 2.58 & 3.64 & 8.86\\
    $FGSM_2$ & 1.52 & 0.69 & 0.45 & 1.95 & 0.78 & 0.81 & 2.57 & 1.8\\
    
    $PGD_1$ & 10.03 & 5.72 & 4.54 & 10.96 & 3.44 & 1.61 & 4 & 10.18\\
    $PGD_2$ & 0.06 & 0 & 0 & 0.12 & 0.24 & 0.18 & 0.64 & 0.06\\
    
    $PGD_3$  & 10.27 & 6.02 & 4.63 & 11.09 & 3.27 & 1.61 & 3.83 & 10.06\\
    $PGD_4$ & 0.06 & 0 & 0 & 0.12 & 0.18 & 0.12 & 0.61 & 0.06\\
    
    $PGD_5$ & 0.12 & 0.03 & 0 & 0.24 & 0.18 & 0.24 & 0.79 & 0.12\\
    
    DIFGSM & 24.56 & 24.16 & 20.83 & 28.01 & 19.39 & 19.43 & 16.74 & 27.41\\
    
    CW & 0 & 0 & 0 & 0 & 0 & 0 & 0 & 0\\
    
    Jitter & 45.87 & 44.28 & 48.39 & 45.42 & 37.43 & 31.98 & 43.75 & 44.73\\
    
    TIFGSM  & 32.78 & 31.08 & 29.68 & 35.76 & 28.31 & 18.99 & 29.83 & 33.04\\
    
    PIFGSM  & 3.62 & 2.1 & 1.62 & 4.46 & 0.9 & 0.6 & 1.71 & 3.44\\
    
    EADEN & 0 & 0 & 0 & 0 & 0 & 0 & 0 & 0\\ 
    OnePixel & 51.75 & 49.39 & 54.93 & 53.41 & 41.4 & 36.01 & 47.55 & 51.54\\
    Pixle & 3.67 & 1.9 & 2.17 & 6.16 & 2.8 & 1.48 & 2.26 & 3.8\\
    SPSA & 44.36 & 42.91 & 44.2 & 46.6 & 30.76 & 28.51 & 38.42 & 44.31\\
    Square  & 0.03 & 0 & 0 & 0.03 & 0.03 & 0 & 0 & 0.03\\
    TAP & 55.53 & 53.4 & 58.55 & 57.72 & 42.93 & 32.54 & 52.48 & 55.35\\
    \textit{ASV} & 53.03 & 55.42 & 49.24 & 59.86 & 51.83 & 64.20 & 62.10 & 52.77 \\
    \textit{FFF (mean-std)} & 48.07 & 41.76 & 42.26 & 39.16 & 43.73 & 66.04 & 48.39 & 54.66 \\
    \textit{FFF (no-data)} & 6.41 & 17.70 & 8.80 & 35.53 & 53.12 & 7.79 & 1.09 & 2.77 \\
    \textit{FFF (one-sample)} & 3.71 & 3.05 & 1.51 & 18.25 & 4.41 & 1.21 & 2.30 & 2.01 \\
    \textit{FG-UAP} & 1.03 & 1.34 & 1.36 & 1.27 & 1.00 & 1.47 & 1.03 & 1.12 \\
    \textit{GD-UAP (mean-std)} & 46.54 & 19.30 & 38.41 & 38.56 & 17.72 & 56.61 & 35.26 & 47.09 \\
    \textit{GD-UAP (no-data)} & 7.43 & 9.67 & 7.29 & 7.59 & 28.10 & 9.88 & 1.18 & 1.33 \\
    \textit{GD-UAP (one-sample)} & 2.30 & 2.52 & 1.18 & 2.92 & 3.94 & 1.69 & 1.06 & 1.03 \\
    \textit{L4A-base} & 63.04 & 52.01 & 58.56 & 56.52 & 42.81 & 69.48 & 10.52 & 56.20 \\
    \textit{L4A-fuse} & 64.09 & 49.22 & 58.50 & 55.87 & 40.42 & 68.70 & 10.54 & 57.69 \\
    \textit{L4A-ugs} & 74.26 & 64.69 & 60.52 & 73.28 & 55.91 & 73.75 & 67.25 & 74.52 \\
    \textit{PD-UAP} & 5.03 & 11.22 & 7.77 & 18.77 & 37.70 & 13.39 & 1.66 & 1.54 \\
    \textit{SSP} & 60.82 & 28.61 & 38.99 & 69.89 & 42.61 & 75.01 & 20.45 & 55.00 \\
    \textit{STD} & 63.15 & 54.63 & 48.31 & 65.25 & 49.60 & 75.97 & 76.34 & 66.58 \\
    \textit{UAP (DeepFool)} & 13.69 & 4.30 & 6.35 & 8.80 & 12.43 & 14.44 & 12.04 & 20.83 \\
    \textit{UAPEPGD} & 70.90 & 66.42 & 58.16 & 71.67 & 57.00 & 76.42 & 76.22 & 70.58 \\

    \midrule
    Clean Accuracy & 86.71 & 83.14 & 80.38  & 82.74 & 79.35 & 85.08 & 86.17 & 86.95 \\
    IAA Avg. & 39.62 \textcolor{red}{\scriptsize{$\downarrow$54\%}}  & 39.21\textcolor{red}{\scriptsize{$\downarrow$53\%}} & 32.36\textcolor{red}{\scriptsize{$\downarrow$60\%}}  & 40.18\textcolor{red}{\scriptsize{$\downarrow$51\%}} & 30.68\textcolor{red}{\scriptsize{$\downarrow$61\%}} & 45.4\textcolor{red}{\scriptsize{$\downarrow$47\%}} & 42.47\textcolor{red}{\scriptsize{$\downarrow$51\%}}  & 39.21\textcolor{red}{\scriptsize{$\downarrow$55\%}} \\
    UAP Avg. & 36.47\textcolor{red}{\scriptsize{$\downarrow$58\%}} & 30.12\textcolor{red}{\scriptsize{$\downarrow$64\%}} & 30.45\textcolor{red}{\scriptsize{$\downarrow$62\%}} & 38.95\textcolor{red}{\scriptsize{$\downarrow$53\%}} & 33.90\textcolor{red}{\scriptsize{$\downarrow$57\%}} & 42.25\textcolor{red}{\scriptsize{$\downarrow$50\%}} & 26.72\textcolor{red}{\scriptsize{$\downarrow$69\%}} & 35.36\textcolor{red}{\scriptsize{$\downarrow$59\%}} \\

    Adv Avg. & 38.14\textcolor{red}{\scriptsize{$\downarrow$56\%}} & 34.93\textcolor{red}{\scriptsize{$\downarrow$58\%}} & 31.46\textcolor{red}{\scriptsize{$\downarrow$61\%}} & 39.60\textcolor{red}{\scriptsize{$\downarrow$52\%}} & 32.20\textcolor{red}{\scriptsize{$\downarrow$59\%}} & 43.92\textcolor{red}{\scriptsize{$\downarrow$48\%}} & 35.06\textcolor{red}{\scriptsize{$\downarrow$59\%}} & 37.40\textcolor{red}{\scriptsize{$\downarrow$57\%}} \\

    \midrule
    \end{tabular}
    \end{adjustbox}
\end{table*}

\clearpage
\subsubsection{Caltech 101}
\begin{table*}[h]
    \centering
    \scriptsize
    \caption{This table presents the results of various instance and universal adversarial perturbation (UAP) attacks on the AirCraft dataset, with all UAP attack names in \textit{italics}. Different configurations of FGSM and PGD are denoted, such as $FGSM_1$ and $PGD_1$. Average results for universal adversarial perturbations (UAP Avg.), instance adversarial attacks (IAA Avg.), and overall adversarial performance (Adv Avg.) are reported at the bottom, including percentage drops relative to clean accuracy.}
    \label{tab:few_shot_small_domain_shift}

    \begin{adjustbox}{width=\textwidth}
    \begin{tabular}{lllllllll}
    \toprule
    
    {} & Barlow & BYOL & DINO & MoCoV3 & SimCLR & Supervised & SwAV & VICReg \\
    \midrule
    
    \textbf{$FGSM_1$}& 8.92 & 5.94 & 4.84 & 11.41 & 2.7 & 2.58 & 3.64 & 8.86\\
    $FGSM_2$ & 1.52 & 0.69 & 0.45 & 1.95 & 0.78 & 0.81 & 2.57 & 1.8\\
    
    $PGD_1$ & 10.03 & 5.72 & 4.54 & 10.96 & 3.44 & 1.61 & 4 & 10.18\\
    $PGD_2$ & 0.06 & 0 & 0 & 0.12 & 0.24 & 0.18 & 0.64 & 0.06\\
    
    $PGD_3$  & 10.27 & 6.02 & 4.63 & 11.09 & 3.27 & 1.61 & 3.83 & 10.06\\
    $PGD_4$ & 0.06 & 0 & 0 & 0.12 & 0.18 & 0.12 & 0.61 & 0.06\\
    
    $PGD_5$ & 0.12 & 0.03 & 0 & 0.24 & 0.18 & 0.24 & 0.79 & 0.12\\
    
    DIFGSM & 24.56 & 24.16 & 20.83 & 28.01 & 19.39 & 19.43 & 16.74 & 27.41\\
    
    CW & 0 & 0 & 0 & 0 & 0 & 0 & 0 & 0\\
    
    Jitter & 45.87 & 44.28 & 48.39 & 45.42 & 37.43 & 31.98 & 43.75 & 44.73\\
    
    TIFGSM  & 32.78 & 31.08 & 29.68 & 35.76 & 28.31 & 18.99 & 29.83 & 33.04\\
    
    PIFGSM  & 3.62 & 2.1 & 1.62 & 4.46 & 0.9 & 0.6 & 1.71 & 3.44\\
    
    EADEN & 0 & 0 & 0 & 0 & 0 & 0 & 0 & 0\\ 
    OnePixel & 51.75 & 49.39 & 54.93 & 53.41 & 41.4 & 36.01 & 47.55 & 51.54\\
    Pixle & 3.67 & 1.9 & 2.17 & 6.16 & 2.8 & 1.48 & 2.26 & 3.8\\
    SPSA & 44.36 & 42.91 & 44.2 & 46.6 & 30.76 & 28.51 & 38.42 & 44.31\\
    Square  & 0.03 & 0 & 0 & 0.03 & 0.03 & 0 & 0 & 0.03\\
    TAP & 55.53 & 53.4 & 58.55 & 57.72 & 42.93 & 32.54 & 52.48 & 55.35\\
    \textit{ASV} & 87.68 & 84.73 & 81.98 & 90.28 & 81.40 & 88.77 & 63.56 & 87.29 \\
    \textit{FFF (mean-std)} & 86.13 & 80.89 & 71.22 & 89.66 & 74.99 & 78.03 & 47.85 & 83.06 \\
    \textit{FFF (no-data)} & 77.25 & 74.04 & 42.23 & 48.19 & 86.39 & 41.23 & 64.82 & 70.62 \\
    \textit{FFF (one-sample)} & 44.59 & 70.90 & 24.11 & 46.35 & 29.45 & 32.53 & 19.60 & 42.75 \\
    \textit{FG-UAP} & 9.42 & 9.03 & 2.68 & 11.08 & 9.40 & 6.14 & 1.76 & 9.35 \\
    \textit{GD-UAP (mean-std)} & 84.10 & 78.40 & 59.99 & 86.99 & 71.77 & 83.20 & 54.39 & 84.00 \\
    \textit{GD-UAP (no-data)} & 61.24 & 66.08 & 51.16 & 63.14 & 76.93 & 61.03 & 36.66 & 65.29 \\
    \textit{GD-UAP (one-sample)} & 53.10 & 67.34 & 44.78 & 51.65 & 56.50 & 62.41 & 24.78 & 62.74 \\
    \textit{L4A-base} & 40.18 & 38.60 & 63.97 & 59.38 & 79.84 & 46.59 & 6.71 & 40.71 \\
    \textit{L4A-fuse} & 39.33 & 39.98 & 65.70 & 58.64 & 79.27 & 45.31 & 6.28 & 39.46 \\
    \textit{L4A-ugs} & 84.46 & 78.05 & 69.87 & 85.20 & 82.89 & 82.04 & 38.86 & 84.11 \\
    \textit{PD-UAP} & 89.44 & 71.47 & 62.44 & 65.06 & 76.64 & 58.79 & 56.72 & 88.92 \\
    \textit{SSP} & 38.38 & 48.11 & 64.43 & 50.72 & 81.07 & 39.81 & 5.19 & 43.45 \\
    \textit{STD} & 90.32 & 91.14 & 87.86 & 91.82 & 88.33 & 90.47 & 87.83 & 90.52 \\
    \textit{UAP (DeepFool)} & 17.40 & 22.13 & 14.11 & 40.02 & 19.29 & 31.40 & 9.62 & 17.39 \\
    \textit{UAPEPGD} & 89.50 & 88.60 & 82.84 & 91.80 & 83.98 & 89.76 & 76.12 & 89.86 \\

    \midrule
    Clean Accuracy & 91.44 & 92.03 & 89.77  & 92.92 & 90.39 & 91.81 & 90.37 & 91.32 \\
    IAA Avg. & 53.63 \textcolor{red}{\tiny{$\downarrow$41\%}}  & 57.03\textcolor{red}{\tiny{$\downarrow$38\%}} & 49.71\textcolor{red}{\tiny{$\downarrow$45\%}}  & 59.32\textcolor{red}{\tiny{$\downarrow$36\%}} & 53.86\textcolor{red}{\tiny{$\downarrow$40\%}} & 54.18\textcolor{red}{\tiny{$\downarrow$41\%}} & 48.01\textcolor{red}{\tiny{$\downarrow$47\%}}  & 53.78\textcolor{red}{\tiny{$\downarrow$41\%}} \\
    UAP Avg. & 62.03\textcolor{red}{\tiny{$\downarrow$32\%}} & 63.09\textcolor{red}{\tiny{$\downarrow$31\%}} & 55.59\textcolor{red}{\tiny{$\downarrow$38\%}} & 64.37\textcolor{red}{\tiny{$\downarrow$31\%}} & 67.38\textcolor{red}{\tiny{$\downarrow$25\%}} & 58.60\textcolor{red}{\tiny{$\downarrow$36\%}} & 37.55\textcolor{red}{\tiny{$\downarrow$58\%}} & 62.47\textcolor{red}{\tiny{$\downarrow$32\%}} \\

    Adv Avg. & 57.58\textcolor{red}{\tiny{$\downarrow$37\%}} & 59.88\textcolor{red}{\tiny{$\downarrow$35\%}} & 52.48\textcolor{red}{\tiny{$\downarrow$42\%}} & 61.70\textcolor{red}{\tiny{$\downarrow$34\%}} & 60.23\textcolor{red}{\tiny{$\downarrow$33\%}} & 56.26\textcolor{red}{\tiny{$\downarrow$39\%}} & 43.09\textcolor{red}{\tiny{$\downarrow$52\%}} & 57.87\textcolor{red}{\tiny{$\downarrow$37\%}} \\

    \midrule
    \end{tabular}
    \end{adjustbox}
\end{table*}

\clearpage
\subsubsection{Cars}
\begin{table*}[h]
    \centering
    \scriptsize
    \caption{This table presents the results of various instance and universal adversarial perturbation (UAP) attacks on the AirCraft dataset, with all UAP attack names in \textit{italics}. Different configurations of FGSM and PGD are denoted, such as $FGSM_1$ and $PGD_1$. Average results for universal adversarial perturbations (UAP Avg.), instance adversarial attacks (IAA Avg.), and overall adversarial performance (Adv Avg.) are reported at the bottom, including percentage drops relative to clean accuracy.}
    \label{tab:few_shot_small_domain_shift}

    \begin{adjustbox}{width=\textwidth}
    \begin{tabular}{lllllllll}
    \toprule
    
    {} & Barlow & BYOL & DINO & MoCoV3 & SimCLR & Supervised & SwAV & VICReg \\
    \midrule
    
    \textbf{$FGSM_1$}& 8.92 & 5.94 & 4.84 & 11.41 & 2.7 & 2.58 & 3.64 & 8.86\\
    $FGSM_2$ & 1.52 & 0.69 & 0.45 & 1.95 & 0.78 & 0.81 & 2.57 & 1.8\\
    
    $PGD_1$ & 10.03 & 5.72 & 4.54 & 10.96 & 3.44 & 1.61 & 4 & 10.18\\
    $PGD_2$ & 0.06 & 0 & 0 & 0.12 & 0.24 & 0.18 & 0.64 & 0.06\\
    
    $PGD_3$  & 10.27 & 6.02 & 4.63 & 11.09 & 3.27 & 1.61 & 3.83 & 10.06\\
    $PGD_4$ & 0.06 & 0 & 0 & 0.12 & 0.18 & 0.12 & 0.61 & 0.06\\
    
    $PGD_5$ & 0.12 & 0.03 & 0 & 0.24 & 0.18 & 0.24 & 0.79 & 0.12\\
    
    DIFGSM & 24.56 & 24.16 & 20.83 & 28.01 & 19.39 & 19.43 & 16.74 & 27.41\\
    
    CW & 0 & 0 & 0 & 0 & 0 & 0 & 0 & 0\\
    
    Jitter & 45.87 & 44.28 & 48.39 & 45.42 & 37.43 & 31.98 & 43.75 & 44.73\\
    
    TIFGSM  & 32.78 & 31.08 & 29.68 & 35.76 & 28.31 & 18.99 & 29.83 & 33.04\\
    
    PIFGSM  & 3.62 & 2.1 & 1.62 & 4.46 & 0.9 & 0.6 & 1.71 & 3.44\\
    
    EADEN & 0 & 0 & 0 & 0 & 0 & 0 & 0 & 0\\ 
    OnePixel & 51.75 & 49.39 & 54.93 & 53.41 & 41.4 & 36.01 & 47.55 & 51.54\\
    Pixle & 3.67 & 1.9 & 2.17 & 6.16 & 2.8 & 1.48 & 2.26 & 3.8\\
    SPSA & 44.36 & 42.91 & 44.2 & 46.6 & 30.76 & 28.51 & 38.42 & 44.31\\
    Square  & 0.03 & 0 & 0 & 0.03 & 0.03 & 0 & 0 & 0.03\\
    TAP & 55.53 & 53.4 & 58.55 & 57.72 & 42.93 & 32.54 & 52.48 & 55.35\\
    \textit{ASV} & 72.91 & 82.15 & 78.42 & 82.19 & 66.07 & 80.09 & 78.46 & 76.31 \\
    \textit{FFF (mean-std)} & 70.69 & 75.79 & 69.49 & 72.04 & 46.51 & 72.89 & 69.07 & 65.95 \\
    \textit{FFF (no-data)} & 31.91 & 37.45 & 46.05 & 60.81 & 35.42 & 20.62 & 49.68 & 47.43 \\
    \textit{FFF (one-sample)} & 47.34 & 39.57 & 44.55 & 53.91 & 2.18 & 55.48 & 43.60 & 65.58 \\
    \textit{FG-UAP} & 0.82 & 1.87 & 1.79 & 3.93 & 0.60 & 2.11 & 1.36 & 0.63 \\
    \textit{GD-UAP (mean-std)} & 60.14 & 66.63 & 69.87 & 71.37 & 38.22 & 56.11 & 70.99 & 53.70 \\
    \textit{GD-UAP (no-data)} & 32.32 & 40.18 & 19.64 & 68.37 & 33.95 & 40.78 & 31.25 & 18.19 \\
    \textit{GD-UAP (one-sample)} & 8.66 & 27.60 & 16.48 & 34.76 & 7.62 & 21.95 & 16.20 & 7.20 \\
    \textit{L4A-base} & 43.38 & 87.19 & 63.86 & 76.66 & 41.94 & 71.56 & 63.09 & 56.05 \\
    \textit{L4A-fuse} & 40.58 & 86.36 & 65.68 & 77.37 & 42.44 & 75.14 & 63.01 & 54.18 \\
    \textit{L4A-ugs} & 68.64 & 85.40 & 82.13 & 86.22 & 35.49 & 79.16 & 84.80 & 75.12 \\
    \textit{PD-UAP} & 55.27 & 64.43 & 36.02 & 77.24 & 57.46 & 46.35 & 57.79 & 31.04 \\
    \textit{SSP} & 28.33 & 56.45 & 52.54 & 68.45 & 13.66 & 60.51 & 39.88 & 26.18 \\
    \textit{STD} & 75.91 & 79.33 & 82.30 & 80.89 & 68.19 & 76.01 & 80.03 & 78.03 \\
    \textit{UAP (DeepFool)} & 11.27 & 42.93 & 28.39 & 49.53 & 5.38 & 43.58 & 24.75 & 13.23 \\
    \textit{UAPEPGD} & 83.88 & 88.09 & 87.02 & 88.56 & 76.12 & 86.32 & 87.23 & 83.14 \\

    \midrule
    Clean Accuracy & 90.54 & 89.91 & 89.80  & 90.11 & 89.62 & 88.99 & 89.73 & 90.41 \\
    IAA Avg. & 48.69 \textcolor{red}{\tiny{$\downarrow$46\%}}  & 54.91\textcolor{red}{\tiny{$\downarrow$39\%}} & 52.72\textcolor{red}{\tiny{$\downarrow$52\%}}  & 55.39\textcolor{red}{\tiny{$\downarrow$39\%}} & 47.63\textcolor{red}{\tiny{$\downarrow$47\%}} & 55.47\textcolor{red}{\tiny{$\downarrow$38\%}} & 52.04\textcolor{red}{\tiny{$\downarrow$42\%}}  & 48.70\textcolor{red}{\tiny{$\downarrow$48\%}} \\
    UAP Avg. & 45.75 \textcolor{red}{\tiny{$\downarrow$49\%}} & 60.09\textcolor{red}{\tiny{$\downarrow$33\%}} & 49.85\textcolor{red}{\tiny{$\downarrow$44\%}} & 65.77\textcolor{red}{\tiny{$\downarrow$27\%}} & 34.10\textcolor{red}{\tiny{$\downarrow$62\%}} & 55.54\textcolor{red}{\tiny{$\downarrow$38\%}} & 50.81\textcolor{red}{\tiny{$\downarrow$43\%}} & 44.46\textcolor{red}{\tiny{$\downarrow$51\%}} \\

    Adv Avg. & 47.31 \textcolor{red}{\tiny{$\downarrow$48\%}} & 57.35\textcolor{red}{\tiny{$\downarrow$36\%}} & 52.74\textcolor{red}{\tiny{$\downarrow$41\%}} & 60.28\textcolor{red}{\tiny{$\downarrow$33\%}} & 42.02\textcolor{red}{\tiny{$\downarrow$53\%}} & 55.50\textcolor{red}{\tiny{$\downarrow$38\%}} & 52.88\textcolor{red}{\tiny{$\downarrow$41\%}} & 47.90\textcolor{red}{\tiny{$\downarrow$47\%}} \\

    \midrule
    \end{tabular}
    \end{adjustbox}
\end{table*}

\clearpage
\subsubsection{CIFAR 10}
\begin{table*}[h]
    \centering
    \scriptsize
    \caption{This table presents the results of various instance and universal adversarial perturbation (UAP) attacks on the AirCraft dataset, with all UAP attack names in \textit{italics}. Different configurations of FGSM and PGD are denoted, such as $FGSM_1$ and $PGD_1$. Average results for universal adversarial perturbations (UAP Avg.), instance adversarial attacks (IAA Avg.), and overall adversarial performance (Adv Avg.) are reported at the bottom, including percentage drops relative to clean accuracy.}
    \label{tab:few_shot_small_domain_shift}

    \begin{adjustbox}{width=\textwidth}
    \begin{tabular}{lllllllll}
    \toprule
    
    {} & Barlow & BYOL & DINO & MoCoV3 & SimCLR & Supervised & SwAV & VICReg \\
    \midrule
    
    \textbf{$FGSM_1$}& 8.92 & 5.94 & 4.84 & 11.41 & 2.7 & 2.58 & 3.64 & 8.86\\
    $FGSM_2$ & 1.52 & 0.69 & 0.45 & 1.95 & 0.78 & 0.81 & 2.57 & 1.8\\
    
    $PGD_1$ & 10.03 & 5.72 & 4.54 & 10.96 & 3.44 & 1.61 & 4 & 10.18\\
    $PGD_2$ & 0.06 & 0 & 0 & 0.12 & 0.24 & 0.18 & 0.64 & 0.06\\
    
    $PGD_3$  & 10.27 & 6.02 & 4.63 & 11.09 & 3.27 & 1.61 & 3.83 & 10.06\\
    $PGD_4$ & 0.06 & 0 & 0 & 0.12 & 0.18 & 0.12 & 0.61 & 0.06\\
    
    $PGD_5$ & 0.12 & 0.03 & 0 & 0.24 & 0.18 & 0.24 & 0.79 & 0.12\\
    
    DIFGSM & 24.56 & 24.16 & 20.83 & 28.01 & 19.39 & 19.43 & 16.74 & 27.41\\
    
    CW & 0 & 0 & 0 & 0 & 0 & 0 & 0 & 0\\
    
    Jitter & 45.87 & 44.28 & 48.39 & 45.42 & 37.43 & 31.98 & 43.75 & 44.73\\
    
    TIFGSM  & 32.78 & 31.08 & 29.68 & 35.76 & 28.31 & 18.99 & 29.83 & 33.04\\
    
    PIFGSM  & 3.62 & 2.1 & 1.62 & 4.46 & 0.9 & 0.6 & 1.71 & 3.44\\
    
    EADEN & 0 & 0 & 0 & 0 & 0 & 0 & 0 & 0\\ 
    OnePixel & 51.75 & 49.39 & 54.93 & 53.41 & 41.4 & 36.01 & 47.55 & 51.54\\
    Pixle & 3.67 & 1.9 & 2.17 & 6.16 & 2.8 & 1.48 & 2.26 & 3.8\\
    SPSA & 44.36 & 42.91 & 44.2 & 46.6 & 30.76 & 28.51 & 38.42 & 44.31\\
    Square  & 0.03 & 0 & 0 & 0.03 & 0.03 & 0 & 0 & 0.03\\
    TAP & 55.53 & 53.4 & 58.55 & 57.72 & 42.93 & 32.54 & 52.48 & 55.35\\
    \textit{ASV} & 93.41 & 71.89 & 94.31 & 86.68 & 88.88 & 76.26 & 93.55 & 91.31 \\
    \textit{FFF (mean-std)} & 85.96 & 63.01 & 44.25 & 72.36 & 86.67 & 55.1 & 29.72 & 77.9 \\
    \textit{FFF (no-data)} & 68.68 & 12.87 & 52.85 & 52.54 & 71.1 & 25.55 & 35.1 & 36.08 \\
    \textit{FFF (one-sample)} & 16.72 & 13.51 & 47.16 & 12.91 & 12.22 & 10.25 & 23.85 & 14.84 \\
    \textit{FG-UAP} & 10.03 & 10.27 & 11.52 & 11.15 & 11.46 & 10.76 & 14.94 & 10.63 \\
    \textit{GD-UAP (mean-std)} & 71.66 & 20.62 & 34.9 & 61.37 & 85.5 & 33.53 & 24.17 & 72.08 \\
    \textit{GD-UAP (no-data)} & 71.78 & 10.31 & 39.89 & 33.17 & 15.38 & 17.34 & 38.98 & 54.72 \\
    \textit{GD-UAP (one-sample)} & 12.43 & 10.12 & 26.46 & 10.35 & 10.26 & 10.05 & 22.34 & 12.13 \\
    \textit{L4A-base} & 33.41 & 14.15 & 71.00 & 25.06 & 58.13 & 14.48 & 73.48 & 29.14 \\
    \textit{L4A-fuse} & 32.70 & 13.32 & 71.38 & 27.32 & 57.63 & 14.83 & 74.09 & 29.74 \\
    \textit{L4A-ugs} & 80.68 & 52.12 & 91.30 & 78.32 & 88.76 & 61.03 & 79.24 & 68.42 \\
    \textit{PD-UAP} & 86.15 & 18.99 & 87.32 & 27.60 & 92.59 & 27.82 & 82.89 & 81.42 \\
    \textit{SSP} & 17.16 & 10.66 & 43.62 & 23.58 & 16.46 & 21.83 & 61.30 & 20.30 \\
    \textit{STD} & 88.90 & 86.48 & 92.52 & 81.73 & 89.29 & 88.23 & 88.49 & 90.13 \\
    \textit{UAP (DeepFool)} & 15.98 & 12.61 & 26.41 & 17.99 & 14.46 & 20.88 & 22.31 & 17.42 \\
    \textit{UAPEPGD} & 95.39 & 90.42 & 95.70 & 93.44 & 93.36 & 91.23 & 95.63 & 93.73 \\

    \midrule
    Clean Accuracy & 97.13 & 96.89 & 96.90  & 96.86 & 97.22 & 96.16 & 96.75 & 97.07 \\
    IAA Avg. & 48.74 \textcolor{red}{\tiny{$\downarrow$50\%}}  & 46.26\textcolor{red}{\tiny{$\downarrow$52\%}} & 54.18\textcolor{red}{\tiny{$\downarrow$44\%}}  & 49.66\textcolor{red}{\tiny{$\downarrow$49\%}} & 49.03\textcolor{red}{\tiny{$\downarrow$50\%}} & 43.37\textcolor{red}{\tiny{$\downarrow$55\%}} & 55.75\textcolor{red}{\tiny{$\downarrow$43\%}}  & 48.31\textcolor{red}{\tiny{$\downarrow$50\%}} \\
    UAP Avg. & 55.07 \textcolor{red}{\tiny{$\downarrow$43\%}} & 31.96\textcolor{red}{\tiny{$\downarrow$67\%}} & 58.16\textcolor{red}{\tiny{$\downarrow$40\%}} & 44.72\textcolor{red}{\tiny{$\downarrow$54\%}} & 55.76\textcolor{red}{\tiny{$\downarrow$43\%}} & 36.20\textcolor{red}{\tiny{$\downarrow$62\%}} & 53.76\textcolor{red}{\tiny{$\downarrow$44\%}} & 50.00\textcolor{red}{\tiny{$\downarrow$48\%}} \\

    Adv Avg. & 51.72 \textcolor{red}{\tiny{$\downarrow$47\%}}  & 39.53\textcolor{red}{\tiny{$\downarrow$59\%}} & 56.05\textcolor{red}{\tiny{$\downarrow$42\%}}  & 47.34\textcolor{red}{\tiny{$\downarrow$51\%}} & 52.20\textcolor{red}{\tiny{$\downarrow$46\%}} & 40.00\textcolor{red}{\tiny{$\downarrow$58\%}} & 54.81\textcolor{red}{\tiny{$\downarrow$43\%}}  & 49.10\textcolor{red}{\tiny{$\downarrow$49\%}} \\

    \midrule
    \end{tabular}
    \end{adjustbox}
\end{table*}

\clearpage
\subsubsection{CIFAR 100}
\begin{table*}[h]
    \centering
    \scriptsize
    \caption{This table presents the results of various instance and universal adversarial perturbation (UAP) attacks on the AirCraft dataset, with all UAP attack names in \textit{italics}. Different configurations of FGSM and PGD are denoted, such as $FGSM_1$ and $PGD_1$. Average results for universal adversarial perturbations (UAP Avg.), instance adversarial attacks (IAA Avg.), and overall adversarial performance (Adv Avg.) are reported at the bottom, including percentage drops relative to clean accuracy.}
    \label{tab:few_shot_small_domain_shift}

    \begin{adjustbox}{width=\textwidth}
    \begin{tabular}{lllllllll}
    \toprule
    
    {} & Barlow & BYOL & DINO & MoCoV3 & SimCLR & Supervised & SwAV & VICReg \\
    \midrule
    
    \textbf{$FGSM_1$}& 8.92 & 5.94 & 4.84 & 11.41 & 2.7 & 2.58 & 3.64 & 8.86\\
    $FGSM_2$ & 1.52 & 0.69 & 0.45 & 1.95 & 0.78 & 0.81 & 2.57 & 1.8\\
    
    $PGD_1$ & 10.03 & 5.72 & 4.54 & 10.96 & 3.44 & 1.61 & 4 & 10.18\\
    $PGD_2$ & 0.06 & 0 & 0 & 0.12 & 0.24 & 0.18 & 0.64 & 0.06\\
    
    $PGD_3$  & 10.27 & 6.02 & 4.63 & 11.09 & 3.27 & 1.61 & 3.83 & 10.06\\
    $PGD_4$ & 0.06 & 0 & 0 & 0.12 & 0.18 & 0.12 & 0.61 & 0.06\\
    
    $PGD_5$ & 0.12 & 0.03 & 0 & 0.24 & 0.18 & 0.24 & 0.79 & 0.12\\
    
    DIFGSM & 24.56 & 24.16 & 20.83 & 28.01 & 19.39 & 19.43 & 16.74 & 27.41\\
    
    CW & 0 & 0 & 0 & 0 & 0 & 0 & 0 & 0\\
    
    Jitter & 45.87 & 44.28 & 48.39 & 45.42 & 37.43 & 31.98 & 43.75 & 44.73\\
    
    TIFGSM  & 32.78 & 31.08 & 29.68 & 35.76 & 28.31 & 18.99 & 29.83 & 33.04\\
    
    PIFGSM  & 3.62 & 2.1 & 1.62 & 4.46 & 0.9 & 0.6 & 1.71 & 3.44\\
    
    EADEN & 0 & 0 & 0 & 0 & 0 & 0 & 0 & 0\\ 
    OnePixel & 51.75 & 49.39 & 54.93 & 53.41 & 41.4 & 36.01 & 47.55 & 51.54\\
    Pixle & 3.67 & 1.9 & 2.17 & 6.16 & 2.8 & 1.48 & 2.26 & 3.8\\
    SPSA & 44.36 & 42.91 & 44.2 & 46.6 & 30.76 & 28.51 & 38.42 & 44.31\\
    Square  & 0.03 & 0 & 0 & 0.03 & 0.03 & 0 & 0 & 0.03\\
    TAP & 55.53 & 53.4 & 58.55 & 57.72 & 42.93 & 32.54 & 52.48 & 55.35\\
    \textit{ASV} & 76.76 & 64.38 & 74.69 & 54.69 & 49.74 & 61.44 & 72.13 & 75.5 \\
    \textit{FFF (mean-std)} & 61.98 & 35.03 & 38.93 & 19.71 & 18.57 & 5.46 & 48.25 & 36.28 \\
    \textit{FFF (no-data)} & 27.47 & 34.56 & 37.48 & 21.76 & 34.83 & 12.85 & 28.24 & 21.22 \\
    \textit{FFF (one-sampl)} & 1.83 & 14.42 & 27.3 & 15.76 & 1.53 & 5.2 & 22.05 & 5.92 \\
    \textit{FG-UAP} & 1.65 & 3.13 & 6.22 & 2.38 & 1.49 & 3.99 & 4.3 & 1.02 \\
    \textit{GD-UAP (mean-std)} & 48.83 & 22.66 & 17.57 & 14.95 & 43.36 & 7.59 & 51.07 & 36.65 \\
    \textit{GD-UAP (no-sample)} & 52.7 & 31.68 & 48.65 & 36.84 & 4.43 & 18.53 & 42.65 & 25.12 \\
    \textit{GD-UAP (one-sample)} & 4.45 & 11.91 & 9.57 & 11.5 & 4.57 & 7.88 & 18.26 & 2.17 \\
    \textit{L4A-base} & 8.49 & 24.86 & 41.34 & 16.66 & 17.31 & 19.07 & 27.93 & 8.92 \\
    \textit{L4A-fuse} & 8.92 & 24.55 & 40.87 & 17.12 & 18.88 & 20.3 & 26.87 & 9.18 \\
    \textit{L4A-ugs} & 57.08 & 61.96 & 73.13 & 46.77 & 49.34 & 45.16 & 70.88 & 59.27 \\
    \textit{PD-UAP} & 66.13 & 64.98 & 65.27 & 60.42 & 53.51 & 17.84 & 53.14 & 66.23 \\
    \textit{SSP} & 24.29 & 32.31 & 29.39 & 31.57 & 7.48 & 13.24 & 36.67 & 8.23 \\
    \textit{STD} & 57.19 & 66.71 & 64.13 & 57.24 & 65.29 & 68.65 & 55.1 & 61.38 \\
    \textit{UAP (DeepFool)} & 5.2 & 14.25 & 16.71 & 16.07 & 7.15 & 11.73 & 18.48 & 6.07 \\
    \textit{UAPEPGD} & 79.63 & 81.17 & 81.13 & 81.71 & 74.3 & 80.04 & 80.4 & 79.19 \\

    \midrule
    Clean Accuracy & 84.60 & 83.88 & 84.69  & 84.49 & 84.44 & 82.63 & 84.37 & 84.27 \\
    IAA Avg. & 35.85 \textcolor{red}{\tiny{$\downarrow$57\%}}  & 44.11\textcolor{red}{\tiny{$\downarrow$47\%}} & 39.68\textcolor{red}{\tiny{$\downarrow$53\%}}  & 44.95\textcolor{red}{\tiny{$\downarrow$47\%}} & 35.54\textcolor{red}{\tiny{$\downarrow$58\%}} & 42.33\textcolor{red}{\tiny{$\downarrow$49\%}} & 40.22\textcolor{red}{\tiny{$\downarrow$52\%}}  & 35.59\textcolor{red}{\tiny{$\downarrow$58\%}} \\
   UAP Avg. & 36.41 \textcolor{red}{\tiny{$\downarrow$57\%}} & 36.79\textcolor{red}{\tiny{$\downarrow$56\%}} & 42.02\textcolor{red}{\tiny{$\downarrow$50\%}} & 31.57\textcolor{red}{\tiny{$\downarrow$63\%}} & 28.24\textcolor{red}{\tiny{$\downarrow$67\%}} & 24.94\textcolor{red}{\tiny{$\downarrow$70\%}} & 41.03\textcolor{red}{\tiny{$\downarrow$51\%}} & 31.40\textcolor{red}{\tiny{$\downarrow$63\%}} \\

    Adv Avg. & 36.12 \textcolor{red}{\tiny{$\downarrow$57\%}} & 40.67\textcolor{red}{\tiny{$\downarrow$52\%}} & 40.78\textcolor{red}{\tiny{$\downarrow$52\%}} & 38.65\textcolor{red}{\tiny{$\downarrow$54\%}} & 32.10\textcolor{red}{\tiny{$\downarrow$62\%}} & 34.14\textcolor{red}{\tiny{$\downarrow$59\%}} & 40.60\textcolor{red}{\tiny{$\downarrow$52\%}} & 33.62\textcolor{red}{\tiny{$\downarrow$60\%}} \\

    \midrule
    \end{tabular}
    \end{adjustbox}
\end{table*}

\clearpage
\subsubsection{DTD}
\begin{table*}[h]
    \centering
    \scriptsize
    \caption{This table presents the results of various instance and universal adversarial perturbation (UAP) attacks on the AirCraft dataset, with all UAP attack names in \textit{italics}. Different configurations of FGSM and PGD are denoted, such as $FGSM_1$ and $PGD_1$. Average results for universal adversarial perturbations (UAP Avg.), instance adversarial attacks (IAA Avg.), and overall adversarial performance (Adv Avg.) are reported at the bottom, including percentage drops relative to clean accuracy.}
    \label{tab:few_shot_small_domain_shift}

    \begin{adjustbox}{width=\textwidth}
    \begin{tabular}{lllllllll}
    \toprule
    
    {} & Barlow & BYOL & DINO & MoCoV3 & SimCLR & Supervised & SwAV & VICReg \\
    \midrule
    
    \textbf{$FGSM_1$}& 8.92 & 5.94 & 4.84 & 11.41 & 2.7 & 2.58 & 3.64 & 8.86\\
    $FGSM_2$ & 1.52 & 0.69 & 0.45 & 1.95 & 0.78 & 0.81 & 2.57 & 1.8\\
    
    $PGD_1$ & 10.03 & 5.72 & 4.54 & 10.96 & 3.44 & 1.61 & 4 & 10.18\\
    $PGD_2$ & 0.06 & 0 & 0 & 0.12 & 0.24 & 0.18 & 0.64 & 0.06\\
    
    $PGD_3$  & 10.27 & 6.02 & 4.63 & 11.09 & 3.27 & 1.61 & 3.83 & 10.06\\
    $PGD_4$ & 0.06 & 0 & 0 & 0.12 & 0.18 & 0.12 & 0.61 & 0.06\\
    
    $PGD_5$ & 0.12 & 0.03 & 0 & 0.24 & 0.18 & 0.24 & 0.79 & 0.12\\
    
    DIFGSM & 24.56 & 24.16 & 20.83 & 28.01 & 19.39 & 19.43 & 16.74 & 27.41\\
    
    CW & 0 & 0 & 0 & 0 & 0 & 0 & 0 & 0\\
    
    Jitter & 45.87 & 44.28 & 48.39 & 45.42 & 37.43 & 31.98 & 43.75 & 44.73\\
    
    TIFGSM  & 32.78 & 31.08 & 29.68 & 35.76 & 28.31 & 18.99 & 29.83 & 33.04\\
    
    PIFGSM  & 3.62 & 2.1 & 1.62 & 4.46 & 0.9 & 0.6 & 1.71 & 3.44\\
    
    EADEN & 0 & 0 & 0 & 0 & 0 & 0 & 0 & 0\\ 
    OnePixel & 51.75 & 49.39 & 54.93 & 53.41 & 41.4 & 36.01 & 47.55 & 51.54\\
    Pixle & 3.67 & 1.9 & 2.17 & 6.16 & 2.8 & 1.48 & 2.26 & 3.8\\
    SPSA & 44.36 & 42.91 & 44.2 & 46.6 & 30.76 & 28.51 & 38.42 & 44.31\\
    Square  & 0.03 & 0 & 0 & 0.03 & 0.03 & 0 & 0 & 0.03\\
    TAP & 55.53 & 53.4 & 58.55 & 57.72 & 42.93 & 32.54 & 52.48 & 55.35\\
     \textit{ASV} & 68.51 & 67.77 & 69.20 & 65.11 & 61.76 & 65.74 & 69.15 & 68.56 \\
    \textit{FFF (mean-std)} & 67.87 & 64.95 & 69.52 & 66.54 & 56.97 & 52.34 & 67.50 & 68.72 \\
    \textit{FFF (no-data)} & 57.18 & 57.13 & 46.91 & 59.52 & 57.39 & 46.22 & 58.56 & 51.97 \\
    \textit{FFF (one-sample)} & 40.16 & 54.63 & 51.86 & 57.29 & 49.95 & 47.29 & 60.16 & 38.56 \\
    \textit{FG-UAP} & 19.10 & 21.54 & 13.30 & 21.28 & 24.84 & 18.94 & 13.46 & 17.23 \\
    \textit{GD-UAP (mean-std)} & 64.47 & 61.91 & 69.68 & 65.27 & 52.82 & 57.18 & 63.67 & 66.33 \\
    \textit{GD-UAP (no-data)} & 57.13 & 59.47 & 46.54 & 58.62 & 57.23 & 50.48 & 56.86 & 56.22 \\
    \textit{GD-UAP (one-sample)} & 57.66 & 52.29 & 42.55 & 48.94 & 45.64 & 47.02 & 47.71 & 55.64 \\
    \textit{L4A-base} & 72.61 & 73.35 & 72.13 & 73.51 & 64.26 & 68.14 & 71.01 & 72.66 \\
    \textit{L4A-fuse} & 72.55 & 72.71 & 72.29 & 73.40 & 64.20 & 68.30 & 71.38 & 72.93 \\
    \textit{L4A-ugs} & 71.60 & 70.69 & 72.39 & 73.30 & 66.65 & 71.12 & 72.93 & 72.87 \\
    \textit{PD-UAP} & 72.13 & 69.79 & 65.74 & 70.90 & 60.11 & 60.64 & 63.35 & 73.24 \\
    \textit{SSP} & 72.45 & 70.21 & 73.03 & 71.70 & 59.73 & 66.17 & 69.63 & 72.18 \\
    \textit{STD} & 71.44 & 71.49 & 73.30 & 72.39 & 69.73 & 71.54 & 72.93 & 72.18 \\
    \textit{UAP (DeepFool)} & 68.03 & 66.49 & 69.47 & 67.71 & 63.09 & 62.77 & 67.93 & 68.56 \\
    \textit{UAPEPGD} & 70.96 & 69.41 & 72.23 & 71.76 & 66.91 & 70.59 & 72.39 & 72.77 \\

    \midrule
    Clean Accuracy & 76.12 & 76.28 & 78.09  & 75.43 & 75.90 & 73.30 & 74.47 & 77.29 \\
    IAA Avg. & 41.42 \textcolor{red}{\tiny{$\downarrow$46\%}}  & 41.89\textcolor{red}{\tiny{$\downarrow$45\%}} & 40.50\textcolor{red}{\tiny{$\downarrow$48\%}}  & 42.91\textcolor{red}{\tiny{$\downarrow$43\%}} & 42.42\textcolor{red}{\tiny{$\downarrow$44\%}} & 38.99\textcolor{red}{\tiny{$\downarrow$47\%}} & 39.65\textcolor{red}{\tiny{$\downarrow$47\%}}  & 41.90\textcolor{red}{\tiny{$\downarrow$46\%}} \\
    UAP Avg. & 62.74 \textcolor{red}{\tiny{$\downarrow$18\%}} & 62.74\textcolor{red}{\tiny{$\downarrow$18\%}} & 61.26\textcolor{red}{\tiny{$\downarrow$22\%}} & 63.58\textcolor{red}{\tiny{$\downarrow$16\%}} & 57.58\textcolor{red}{\tiny{$\downarrow$24\%}} & 57.78\textcolor{red}{\tiny{$\downarrow$21\%}} & 62.41\textcolor{red}{\tiny{$\downarrow$16\%}} & 62.54\textcolor{red}{\tiny{$\downarrow$19\%}} \\

    Adv Avg. & 51.46 \textcolor{red}{\tiny{$\downarrow$32\%}}  & 51.70\textcolor{red}{\tiny{$\downarrow$32\%}} & 50.27\textcolor{red}{\tiny{$\downarrow$36\%}}  & 52.64\textcolor{red}{\tiny{$\downarrow$30\%}} & 49.56\textcolor{red}{\tiny{$\downarrow$35\%}} & 47.83\textcolor{red}{\tiny{$\downarrow$35\%}} & 50.36\textcolor{red}{\tiny{$\downarrow$32\%}}  & 51.62\textcolor{red}{\tiny{$\downarrow$33\%}} \\

    \midrule
    \end{tabular}
    \end{adjustbox}
\end{table*}

\clearpage
\subsubsection{Flowers}
\begin{table*}[h]
    \centering
    \scriptsize
    \caption{This table presents the results of various instance and universal adversarial perturbation (UAP) attacks on the AirCraft dataset, with all UAP attack names in \textit{italics}. Different configurations of FGSM and PGD are denoted, such as $FGSM_1$ and $PGD_1$. Average results for universal adversarial perturbations (UAP Avg.), instance adversarial attacks (IAA Avg.), and overall adversarial performance (Adv Avg.) are reported at the bottom, including percentage drops relative to clean accuracy.}
    \label{tab:few_shot_small_domain_shift}

    \begin{adjustbox}{width=\textwidth}
    \begin{tabular}{lllllllll}
    \toprule
    
    {} & Barlow & BYOL & DINO & MoCoV3 & SimCLR & Supervised & SwAV & VICReg \\
    \midrule
    
    \textbf{$FGSM_1$}& 8.92 & 5.94 & 4.84 & 11.41 & 2.7 & 2.58 & 3.64 & 8.86\\
    $FGSM_2$ & 1.52 & 0.69 & 0.45 & 1.95 & 0.78 & 0.81 & 2.57 & 1.8\\
    
    $PGD_1$ & 10.03 & 5.72 & 4.54 & 10.96 & 3.44 & 1.61 & 4 & 10.18\\
    $PGD_2$ & 0.06 & 0 & 0 & 0.12 & 0.24 & 0.18 & 0.64 & 0.06\\
    
    $PGD_3$  & 10.27 & 6.02 & 4.63 & 11.09 & 3.27 & 1.61 & 3.83 & 10.06\\
    $PGD_4$ & 0.06 & 0 & 0 & 0.12 & 0.18 & 0.12 & 0.61 & 0.06\\
    
    $PGD_5$ & 0.12 & 0.03 & 0 & 0.24 & 0.18 & 0.24 & 0.79 & 0.12\\
    
    DIFGSM & 24.56 & 24.16 & 20.83 & 28.01 & 19.39 & 19.43 & 16.74 & 27.41\\
    
    CW & 0 & 0 & 0 & 0 & 0 & 0 & 0 & 0\\
    
    Jitter & 45.87 & 44.28 & 48.39 & 45.42 & 37.43 & 31.98 & 43.75 & 44.73\\
    
    TIFGSM  & 32.78 & 31.08 & 29.68 & 35.76 & 28.31 & 18.99 & 29.83 & 33.04\\
    
    PIFGSM  & 3.62 & 2.1 & 1.62 & 4.46 & 0.9 & 0.6 & 1.71 & 3.44\\
    
    EADEN & 0 & 0 & 0 & 0 & 0 & 0 & 0 & 0\\ 
    OnePixel & 51.75 & 49.39 & 54.93 & 53.41 & 41.4 & 36.01 & 47.55 & 51.54\\
    Pixle & 3.67 & 1.9 & 2.17 & 6.16 & 2.8 & 1.48 & 2.26 & 3.8\\
    SPSA & 44.36 & 42.91 & 44.2 & 46.6 & 30.76 & 28.51 & 38.42 & 44.31\\
    Square  & 0.03 & 0 & 0 & 0.03 & 0.03 & 0 & 0 & 0.03\\
    TAP & 55.53 & 53.4 & 58.55 & 57.72 & 42.93 & 32.54 & 52.48 & 55.35\\
    \textit{ASV} & 69.45 & 77.27 & 69.56 & 82.40 & 75.33 & 77.53 & 63.57 & 73.27 \\
    \textit{FFF (mean-std)} & 67.79 & 90.99 & 54.47 & 71.41 & 61.67 & 77.85 & 56.39 & 57.47 \\
    \textit{FFF (no-data)} & 57.49 & 74.01 & 53.72 & 61.22 & 58.68 & 39.71 & 52.00 & 72.22 \\
    \textit{FFF (one-sample)} & 57.62 & 68.44 & 36.91 & 62.99 & 62.40 & 25.02 & 66.46 & 51.71 \\
    \textit{FG-UAP} & 14.02 & 17.87 & 2.56 & 13.45 & 6.25 & 10.20 & 4.10 & 11.32 \\
    \textit{GD-UAP (mean-std)} & 63.39 & 70.67 & 58.55 & 75.25 & 61.16 & 70.12 & 56.21 & 63.13 \\
    \textit{GD-UAP (no-data)} & 66.62 & 95.02 & 46.14 & 62.45 & 60.30 & 47.44 & 55.74 & 74.80 \\
    \textit{GD-UAP (one-sample)} & 30.40 & 14.74 & 26.38 & 41.49 & 59.61 & 11.85 & 22.44 & 40.30 \\
    \textit{L4A-base} & 67.70 & 81.02 & 81.03 & 75.93 & 56.36 & 70.48 & 64.16 & 72.19 \\
    \textit{L4A-fuse} & 68.34 & 83.02 & 81.59 & 76.85 & 56.86 & 71.46 & 67.54 & 72.42 \\
    \textit{L4A-ugs} & 84.09 & 84.89 & 63.96 & 88.48 & 75.71 & 85.65 & 81.73 & 84.32 \\
    \textit{PD-UAP} & 81.61 & 87.36 & 47.50 & 77.88 & 70.19 & 71.09 & 52.30 & 78.08 \\
    \textit{SSP} & 55.77 & 43.54 & 62.99 & 77.63 & 63.71 & 43.64 & 64.75 & 65.42 \\
    \textit{STD} & 69.43 & 74.60 & 66.85 & 71.59 & 67.20 & 69.42 & 67.86 & 67.67 \\
    \textit{UAP (DeepFool)} & 30.68 & 30.28 & 31.87 & 61.57 & 40.14 & 38.88 & 13.47 & 37.46 \\
    \textit{UAPEPGD} & 86.74 & 90.21 & 82.86 & 91.61 & 85.85 & 90.96 & 83.43 & 87.68 \\

    \midrule
    Clean Accuracy & 97.41 & 96.74 & 97.18  & 96.70 & 95.18 & 96.73 & 96.68 & 96.86 \\
    IAA Avg. & 60.24 \textcolor{red}{\tiny{$\downarrow$38\%}}  & 61.32\textcolor{red}{\tiny{$\downarrow$37\%}} & 54.30\textcolor{red}{\tiny{$\downarrow$44\%}}  & 60.98\textcolor{red}{\tiny{$\downarrow$37\%}} & 51.34\textcolor{red}{\tiny{$\downarrow$46\%}} & 60.79\textcolor{red}{\tiny{$\downarrow$37\%}} & 55.66\textcolor{red}{\tiny{$\downarrow$42\%}}  & 59.97\textcolor{red}{\tiny{$\downarrow$38\%}} \\
    UAP Avg. & 60.69 \textcolor{red}{\tiny{$\downarrow$38\%}} & 67.75\textcolor{red}{\tiny{$\downarrow$30\%}} & 54.18\textcolor{red}{\tiny{$\downarrow$44\%}} & 68.26\textcolor{red}{\tiny{$\downarrow$29\%}} & 60.09\textcolor{red}{\tiny{$\downarrow$37\%}} & 56.33\textcolor{red}{\tiny{$\downarrow$42\%}} & 54.51\textcolor{red}{\tiny{$\downarrow$44\%}} & 63.09\textcolor{red}{\tiny{$\downarrow$35\%}} \\

    Adv Avg. & 60.45 \textcolor{red}{\tiny{$\downarrow$38\%}}  & 64.34\textcolor{red}{\tiny{$\downarrow$33\%}} & 54.25\textcolor{red}{\tiny{$\downarrow$44\%}}  & 64.41\textcolor{red}{\tiny{$\downarrow$33\%}} & 55.46\textcolor{red}{\tiny{$\downarrow$42\%}} & 58.69\textcolor{red}{\tiny{$\downarrow$39\%}} & 55.12\textcolor{red}{\tiny{$\downarrow$43\%}}  & 61.44\textcolor{red}{\tiny{$\downarrow$37\%}} \\

    \midrule
    \end{tabular}
    \end{adjustbox}
\end{table*}

\clearpage
\subsubsection{Food}
\begin{table*}[h]
    \centering
    \scriptsize
    \caption{This table presents the results of various instance and universal adversarial perturbation (UAP) attacks on the AirCraft dataset, with all UAP attack names in \textit{italics}. Different configurations of FGSM and PGD are denoted, such as $FGSM_1$ and $PGD_1$. Average results for universal adversarial perturbations (UAP Avg.), instance adversarial attacks (IAA Avg.), and overall adversarial performance (Adv Avg.) are reported at the bottom, including percentage drops relative to clean accuracy.}
    \label{tab:few_shot_small_domain_shift}

    \begin{adjustbox}{width=\textwidth}
    \begin{tabular}{lllllllll}
    \toprule
    
    {} & Barlow & BYOL & DINO & MoCoV3 & SimCLR & Supervised & SwAV & VICReg \\
    \midrule
    
    \textbf{$FGSM_1$}& 8.92 & 5.94 & 4.84 & 11.41 & 2.7 & 2.58 & 3.64 & 8.86\\
    $FGSM_2$ & 1.52 & 0.69 & 0.45 & 1.95 & 0.78 & 0.81 & 2.57 & 1.8\\
    
    $PGD_1$ & 10.03 & 5.72 & 4.54 & 10.96 & 3.44 & 1.61 & 4 & 10.18\\
    $PGD_2$ & 0.06 & 0 & 0 & 0.12 & 0.24 & 0.18 & 0.64 & 0.06\\
    
    $PGD_3$  & 10.27 & 6.02 & 4.63 & 11.09 & 3.27 & 1.61 & 3.83 & 10.06\\
    $PGD_4$ & 0.06 & 0 & 0 & 0.12 & 0.18 & 0.12 & 0.61 & 0.06\\
    
    $PGD_5$ & 0.12 & 0.03 & 0 & 0.24 & 0.18 & 0.24 & 0.79 & 0.12\\
    
    DIFGSM & 24.56 & 24.16 & 20.83 & 28.01 & 19.39 & 19.43 & 16.74 & 27.41\\
    
    CW & 0 & 0 & 0 & 0 & 0 & 0 & 0 & 0\\
    
    Jitter & 45.87 & 44.28 & 48.39 & 45.42 & 37.43 & 31.98 & 43.75 & 44.73\\
    
    TIFGSM  & 32.78 & 31.08 & 29.68 & 35.76 & 28.31 & 18.99 & 29.83 & 33.04\\
    
    PIFGSM  & 3.62 & 2.1 & 1.62 & 4.46 & 0.9 & 0.6 & 1.71 & 3.44\\
    
    EADEN & 0 & 0 & 0 & 0 & 0 & 0 & 0 & 0\\ 
    OnePixel & 51.75 & 49.39 & 54.93 & 53.41 & 41.4 & 36.01 & 47.55 & 51.54\\
    Pixle & 3.67 & 1.9 & 2.17 & 6.16 & 2.8 & 1.48 & 2.26 & 3.8\\
    SPSA & 44.36 & 42.91 & 44.2 & 46.6 & 30.76 & 28.51 & 38.42 & 44.31\\
    Square  & 0.03 & 0 & 0 & 0.03 & 0.03 & 0 & 0 & 0.03\\
    TAP & 55.53 & 53.4 & 58.55 & 57.72 & 42.93 & 32.54 & 52.48 & 55.35\\
    \textit{ASV} & 66.85 & 72.56 & 76.44 & 67.11 & 63.10 & 68.71 & 73.28 & 68.88 \\
    \textit{FFF (mean-std)} & 56.35 & 64.24 & 72.43 & 62.39 & 46.44 & 21.16 & 70.21 & 63.32 \\
    \textit{FFF (no-data)} & 27.32 & 42.11 & 34.44 & 43.99 & 30.61 & 24.10 & 35.49 & 36.72 \\
    \textit{FFF (one-sample)} & 16.11 & 20.92 & 49.03 & 22.00 & 11.71 & 6.63 & 23.65 & 18.35 \\
    \textit{FG-UAP} & 3.60 & 3.51 & 5.25 & 3.18 & 2.35 & 5.27 & 5.69 & 3.48 \\
    \textit{GD-UAP (mean-std)} & 51.73 & 47.56 & 64.19 & 69.35 & 51.11 & 34.44 & 68.26 & 66.41 \\
    \textit{GD-UAP (no-data)} & 20.73 & 49.44 & 30.21 & 45.98 & 25.59 & 15.42 & 25.43 & 39.66 \\
    \textit{GD-UAP (one-sample)} & 14.78 & 18.09 & 24.88 & 41.16 & 22.17 & 7.45 & 7.39 & 31.94 \\
    \textit{L4A-base} & 34.63 & 40.48 & 42.81 & 32.70 & 26.26 & 34.83 & 18.31 & 31.63 \\
    \textit{L4A-fuse} & 33.93 & 40.76 & 42.48 & 33.41 & 24.73 & 34.40 & 17.74 & 24.70 \\
    \textit{L4A-ugs} & 59.73 & 67.54 & 72.25 & 60.77 & 10.41 & 35.42 & 58.80 & 34.90 \\
    \textit{PD-UAP} & 28.06 & 67.88 & 57.45 & 62.27 & 65.10 & 34.61 & 36.81 & 75.30 \\
    \textit{SSP} & 18.81 & 25.50 & 25.47 & 32.13 & 8.99 & 6.28 & 32.83 & 21.57 \\
    \textit{STD} & 57.58 & 55.80 & 60.02 & 50.13 & 42.12 & 52.72 & 61.46 & 55.82 \\
    \textit{UAP (DeepFool)} & 7.20 & 7.26 & 12.26 & 10.01 & 13.10 & 5.28 & 8.71 & 8.13 \\
    \textit{UAPEPGD} & 77.46 & 80.01 & 81.65 & 81.49 & 73.45 & 78.37 & 81.32 & 78.15 \\

    \midrule
    Clean Accuracy & 83.93 & 85.63 & 87.65  & 85.82 & 82.30 & 84.35 & 87.16 & 83.71 \\
    IAA Avg. & 27.80 \textcolor{red}{\tiny{$\downarrow$67\%}}  & 36.30\textcolor{red}{\tiny{$\downarrow$58\%}} & 34.12\textcolor{red}{\tiny{$\downarrow$61.06\%}} & 36.09\textcolor{red}{\tiny{$\downarrow$58\%}} & 28.89\textcolor{red}{\tiny{$\downarrow$65\%}} & 34.74\textcolor{red}{\tiny{$\downarrow$59\%}} & 34.26\textcolor{red}{\tiny{$\downarrow$59\%}}  & 31.71\textcolor{red}{\tiny{$\downarrow$62\%}} \\
    UAP Avg. & 35.93 \textcolor{red}{\tiny{$\downarrow$57\%}} & 43.98\textcolor{red}{\tiny{$\downarrow$49\%}} & 46.95\textcolor{red}{\tiny{$\downarrow$46\%}} & 44.88\textcolor{red}{\tiny{$\downarrow$48\%}} & 32.33\textcolor{red}{\tiny{$\downarrow$61\%}} & 29.07\textcolor{red}{\tiny{$\downarrow$66\%}} & 39.09\textcolor{red}{\tiny{$\downarrow$55\%}} & 41.18\textcolor{red}{\tiny{$\downarrow$51\%}} \\

    Adv Avg. & 35.94 \textcolor{red}{\tiny{$\downarrow$57\%}}  & 39.91\textcolor{red}{\tiny{$\downarrow$53\%}} & 40.16\textcolor{red}{\tiny{$\downarrow$54\%}}  & 40.23\textcolor{red}{\tiny{$\downarrow$53\%}} & 30.51\textcolor{red}{\tiny{$\downarrow$63\%}} & 32.07\textcolor{red}{\tiny{$\downarrow$62\%}} & 36.53\textcolor{red}{\tiny{$\downarrow$58\%}}  & 36.17\textcolor{red}{\tiny{$\downarrow$57\%}} \\

    \midrule
    \end{tabular}
    \end{adjustbox}
\end{table*}

\clearpage
\subsubsection{Pets}
\begin{table*}[h]
    \centering
    \scriptsize
    \caption{This table presents the results of various instance and universal adversarial perturbation (UAP) attacks on the AirCraft dataset, with all UAP attack names in \textit{italics}. Different configurations of FGSM and PGD are denoted, such as $FGSM_1$ and $PGD_1$. Average results for universal adversarial perturbations (UAP Avg.), instance adversarial attacks (IAA Avg.), and overall adversarial performance (Adv Avg.) are reported at the bottom, including percentage drops relative to clean accuracy.}
    \label{tab:few_shot_small_domain_shift}

    \begin{adjustbox}{width=\textwidth}
    \begin{tabular}{lllllllll}
    \toprule
    
    {} & Barlow & BYOL & DINO & MoCoV3 & SimCLR & Supervised & SwAV & VICReg \\
    \midrule
    
    \textbf{$FGSM_1$}& 8.92 & 5.94 & 4.84 & 11.41 & 2.7 & 2.58 & 3.64 & 8.86\\
    $FGSM_2$ & 1.52 & 0.69 & 0.45 & 1.95 & 0.78 & 0.81 & 2.57 & 1.8\\
    
    $PGD_1$ & 10.03 & 5.72 & 4.54 & 10.96 & 3.44 & 1.61 & 4 & 10.18\\
    $PGD_2$ & 0.06 & 0 & 0 & 0.12 & 0.24 & 0.18 & 0.64 & 0.06\\
    
    $PGD_3$  & 10.27 & 6.02 & 4.63 & 11.09 & 3.27 & 1.61 & 3.83 & 10.06\\
    $PGD_4$ & 0.06 & 0 & 0 & 0.12 & 0.18 & 0.12 & 0.61 & 0.06\\
    
    $PGD_5$ & 0.12 & 0.03 & 0 & 0.24 & 0.18 & 0.24 & 0.79 & 0.12\\
    
    DIFGSM & 24.56 & 24.16 & 20.83 & 28.01 & 19.39 & 19.43 & 16.74 & 27.41\\
    
    CW & 0 & 0 & 0 & 0 & 0 & 0 & 0 & 0\\
    
    Jitter & 45.87 & 44.28 & 48.39 & 45.42 & 37.43 & 31.98 & 43.75 & 44.73\\
    
    TIFGSM  & 32.78 & 31.08 & 29.68 & 35.76 & 28.31 & 18.99 & 29.83 & 33.04\\
    
    PIFGSM  & 3.62 & 2.1 & 1.62 & 4.46 & 0.9 & 0.6 & 1.71 & 3.44\\
    
    EADEN & 0 & 0 & 0 & 0 & 0 & 0 & 0 & 0\\ 
    OnePixel & 51.75 & 49.39 & 54.93 & 53.41 & 41.4 & 36.01 & 47.55 & 51.54\\
    Pixle & 3.67 & 1.9 & 2.17 & 6.16 & 2.8 & 1.48 & 2.26 & 3.8\\
    SPSA & 44.36 & 42.91 & 44.2 & 46.6 & 30.76 & 28.51 & 38.42 & 44.31\\
    Square  & 0.03 & 0 & 0 & 0.03 & 0.03 & 0 & 0 & 0.03\\
    TAP & 55.53 & 53.4 & 58.55 & 57.72 & 42.93 & 32.54 & 52.48 & 55.35\\
    \textit{ASV} & 86.65 & 87.60 & 86.78 & 85.55 & 78.08 & 87.31 & 86.05 & 86.12 \\
    \textit{FFF (mean-std)} & 84.69 & 84.25 & 79.83 & 85.91 & 73.65 & 85.76 & 68.35 & 83.45 \\
    \textit{FFF (no-data)} & 48.95 & 75.83 & 47.83 & 74.82 & 66.53 & 70.05 & 73.33 & 64.11 \\
    \textit{FFF (one-sample)} & 26.50 & 79.70 & 45.52 & 68.01 & 50.15 & 74.97 & 54.46 & 29.36 \\
    \textit{FG-UAP} & 8.10 & 6.85 & 7.12 & 13.08 & 25.21 & 6.23 & 5.79 & 5.42 \\
    \textit{GD-UAP (mean-std)} & 82.46 & 85.03 & 76.63 & 85.23 & 66.49 & 83.99 & 81.65 & 84.72 \\
    \textit{GD-UAP (no-data)} & 65.76 & 68.77 & 53.73 & 66.01 & 72.91 & 75.24 & 76.37 & 62.12 \\
    \textit{GD-UAP (one-sample)} & 34.98 & 28.91 & 16.56 & 50.63 & 52.54 & 55.38 & 25.65 & 31.27 \\
    \textit{L4A-base} & 83.76 & 81.41 & 83.84 & 82.30 & 72.93 & 88.88 & 58.86 & 83.19 \\
    \textit{L4A-fuse} & 83.03 & 81.08 & 83.04 & 82.39 & 72.59 & 88.36 & 57.81 & 82.03 \\
    \textit{L4A-ugs} & 88.82 & 88.76 & 86.91 & 90.43 & 77.88 & 90.76 & 87.71 & 88.77 \\
    \textit{PD-UAP} & 86.63 & 81.05 & 63.30 & 79.72 & 74.99 & 76.75 & 74.91 & 86.63 \\
    \textit{SSP} & 79.54 & 81.60 & 83.18 & 82.57 & 72.05 & 84.12 & 70.15 & 76.15 \\
    \textit{STD} & 86.94 & 86.94 & 81.23 & 89.15 & 79.74 & 90.81 & 84.33 & 86.63 \\
    \textit{UAP (DeepFool)} & 40.81 & 44.95 & 38.77 & 44.69 & 69.07 & 60.11 & 31.96 & 43.56 \\
    \textit{UAPEPGD} & 88.68 & 88.69 & 87.07 & 90.78 & 81.50 & 91.88 & 87.52 & 88.64 \\

    \midrule
    Clean Accuracy & 90.89 & 91.34 & 90.24  & 92.19 & 88.51 & 93.94 & 90.50 & 90.92 \\
    IAA Avg. & 43.21 \textcolor{red}{\tiny{$\downarrow$52\%}}  & 44.69\textcolor{red}{\tiny{$\downarrow$51\%}} & 40.01\textcolor{red}{\tiny{$\downarrow$56\%}}  & 48.33\textcolor{red}{\tiny{$\downarrow$48\%}} & 42.03\textcolor{red}{\tiny{$\downarrow$53\%}} & 46.53\textcolor{red}{\tiny{$\downarrow$50\%}} & 39.53\textcolor{red}{\tiny{$\downarrow$56\%}}  & 43.76\textcolor{red}{\tiny{$\downarrow$52\%}} \\
    UAP Avg. & 67.27 \textcolor{red}{\tiny{$\downarrow$26\%}} & 71.96\textcolor{red}{\tiny{$\downarrow$21\%}} & 63.83\textcolor{red}{\tiny{$\downarrow$29\%}} & 73.20\textcolor{red}{\tiny{$\downarrow$21\%}} & 67.90\textcolor{red}{\tiny{$\downarrow$23\%}} & 75.66\textcolor{red}{\tiny{$\downarrow$19\%}} & 64.06\textcolor{red}{\tiny{$\downarrow$29\%}} & 67.63\textcolor{red}{\tiny{$\downarrow$26\%}} \\

    Adv Avg. & 54.53 \textcolor{red}{\tiny{$\downarrow$40\%}}  & 57.52\textcolor{red}{\tiny{$\downarrow$37\%}} & 51.23\textcolor{red}{\tiny{$\downarrow$43\%}}  & 60.03\textcolor{red}{\tiny{$\downarrow$35\%}} & 54.20\textcolor{red}{\tiny{$\downarrow$39\%}} & 60.24\textcolor{red}{\tiny{$\downarrow$36\%}} & 51.07\textcolor{red}{\tiny{$\downarrow$44\%}}  & 54.99\textcolor{red}{\tiny{$\downarrow$40\%}} \\

    \midrule
    \end{tabular}
    \end{adjustbox}
\end{table*}

\end{document}